\documentclass[10pt,twocolumn,letterpaper]{article}

\usepackage{iccv}
\usepackage{times}
\usepackage{epsfig}
\usepackage{graphicx}
\usepackage{relsize}
\usepackage[normalem]{ulem}
\usepackage{amsmath}

\DeclareMathOperator*{\argmin}{arg\,min}
\usepackage{amssymb}
\usepackage{xcolor}
\usepackage{multirow}
\usepackage{tabularx}
\usepackage[export]{adjustbox}
\usepackage{booktabs} 
\usepackage[accsupp]{axessibility}
\newcommand{\graph}{\mathcal{G}}

\usepackage[pagebackref=true,breaklinks=true,letterpaper=true,colorlinks,bookmarks=false]{hyperref}

\iccvfinalcopy

\def\iccvPaperID{11550}

\ificcvfinal\fi

\begin{document}

\title{Box-based Refinement for Weakly Supervised\\ and Unsupervised Localization Tasks}

\author{Eyal Gomel\\
Tel Aviv University\\
{\tt\small eyalgomel12@gmail.com}
\and
Tal Shaharabany\\
Tel Aviv University\\
{\tt\small shaharabany@mail.tau.ac.il}
\and
Lior Wolf\\
Tel Aviv University\\
{\tt\small wolf@cs.tau.ac.il}
}

\maketitle
\ificcvfinal\thispagestyle{empty}\fi

\begin{abstract}
{\color{black}It has been established that training a box-based detector network can enhance the localization performance of weakly supervised and unsupervised methods. Moreover, we extend this understanding by demonstrating that these detectors can be utilized to improve the original network, paving the way for further advancements. To accomplish this, we train the detectors on top of the network output instead of the image data and apply suitable loss backpropagation. Our findings reveal a significant improvement in phrase grounding for the ``what is where by looking'' task, as well as various methods of unsupervised object discovery. Our code is available at \href{https://github.com/eyalgomel/box-based-refinement}{https://github.com/eyalgomel/box-based-refinement}.}
\end{abstract}

\section{Introduction}

In the task of unsupervised object discovery, one uses clustering methods to find a subset of the image in which the patches are highly similar, while being different from patches in other image locations. The similarity is computed using the embedding provided, e.g., by a transformer $f$ that was trained using a self-supervised loss. The grouping in the embedding space does not guarantee that a single continuous image region will be selected, and often one region out of many is selected, based on some heuristic. 

It has been repeatedly shown~\cite{simeoni2021lost, wang2022tokencut, bielski2022move} that by training a detection network, such as faster R-CNN\cite{renNIPS15fasterrcnn}, one can improve the object discovery metrics. This subsequent detector has two favorable properties over the primary discovery method: it is bounded to a box shape and shares knowledge across the various samples. 

\begin{figure}[t]
    \centering
    \begin{tabular}{@{}c@{~}c@{~}c@{}}
     \includegraphics[valign=B,width=0.324\linewidth]{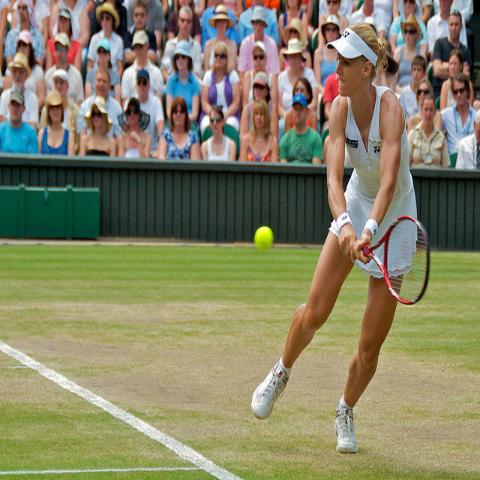} &
    \includegraphics[valign=B,width=0.324\linewidth]{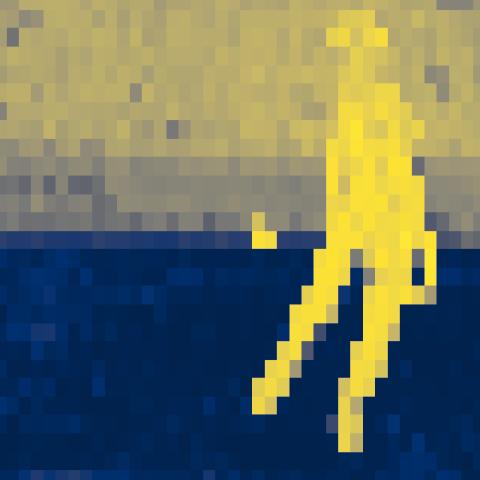} &
    \includegraphics[valign=B,width=0.324\linewidth]{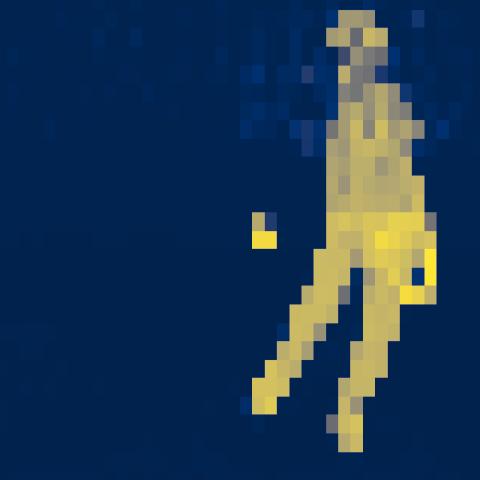} \\
(a) & (b) & (c)\\

     \includegraphics[valign=B,width=0.324\linewidth]{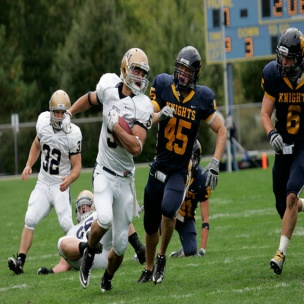} &
    \includegraphics[valign=B,width=0.324\linewidth]{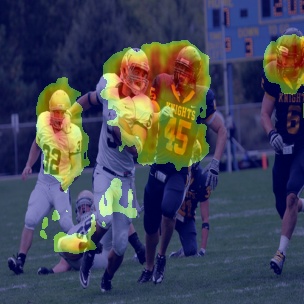} &
    \includegraphics[valign=B,width=0.324\linewidth]{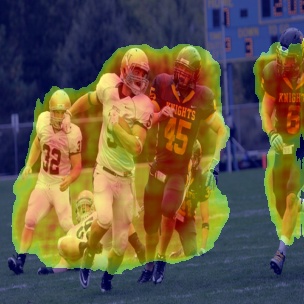} \\
(d) & (e) & (f)\\
      ~\\
    \end{tabular}
    \caption{Examples of refining localization networks. The top row depicts an example of unsupervised object discovery. (a) the input image (b) the normalized cut eigenvector using the original DINO~\cite{caron2021emerging} network $f$, as extracted with the TokenCut\cite{wang2022tokencut} method. (c) the same eigenvector using the refined DINO network $f^h$ our method produces. The bottom row contains phrase grounding results (d) the original input corresponding to the phrase ``two football teams'', (e) the localization map using the image-text network $g$ of \cite{shaharabany2022looking}, and (f) the localization map using the refined $g^h$. }
    \label{fig:teaser}
\end{figure}

In this work, we show that such a detector can also be used to improve the underlying self-supervised similarity. This is done by training a detector network $h$ not on top of the image features, as was done previously, but on the output map of network $f$. Once the detector network $h$ is trained, we freeze it and use the same loss that was used to train the detector network to refine the underlying representation of $f$. 

At this point, the detector network serves as a way to link a recovered set of detection boxes to an underlying feature map of $f$. Without it, deriving a loss would be extremely challenging, since the process used for extracting the detection box from $f$ is typically non-differentiable.  

The outcome of this process is a refined network $f^h$, obtained by fine-tuning $f$ using network $h$. The finetuned network produces a representation that leads to a spatially coherent grouping of regions, as demonstrated in Fig.~\ref{fig:teaser}(a-c).

A similar process is used for the phrase grounding problem. In this case, given a textual phrase, a network $g$ is trained to mark a matching image region. Supervision is performed at the image level, without localization information, a process known as weakly supervised training. In this case, the same loss is used to train a network $h$ on a set of extracted regions, and then to refine $g$. 

Our method exhibits remarkable versatility, as demonstrated through extensive testing on multiple benchmarks, two phrase grounding tasks, and various unsupervised object discovery methods. In all cases, our method consistently achieves significant improvements across all metrics, surpassing the performance of state-of-the-art methods. The move approach introduced trains a detector on the network output rather than the image data. This strategy, distinct from previous work, allows us to refine the primary network independently and further enhance its performance. 

\begin{figure*}
	\centering
	\includegraphics[width=0.95\linewidth]{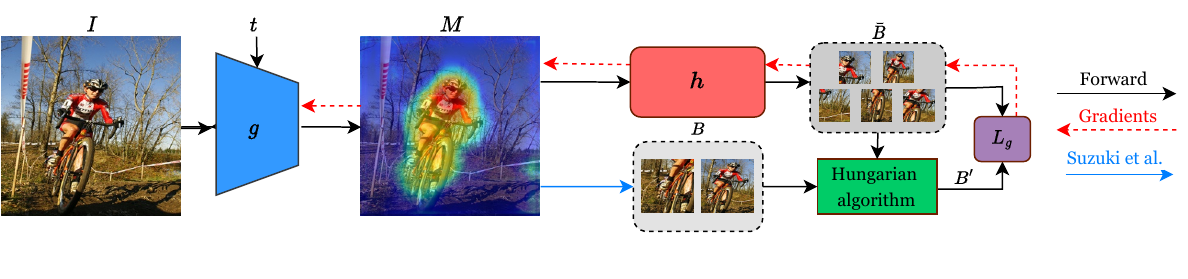}
	\caption{An illustration of our method. The phrased grounding network $f$ is given the input image $I$ and a text phrase $t$ and produces a heatmap $M$. A heuristic (blue line) then produces a set of bounding boxes $B$ from this map that are used to train a detection network $h$, which outputs a set of boxes $\bar B$. The loss that is used is applied after applying the optimal permutation.}
	\label{fig:arch}
\end{figure*}

\section{Related work}

Our method is tested on two localization tasks that are not fully supervised: unsupervised object discovery (detection) and phrase grounding. {\color{black}Numerous studies have been introduced in the realm of unsupervised object discovery, alongside akin tasks involving detection and segmentation}, using different techniques and methods to discover and localize objects in images (and videos) without requiring explicit object annotations. In particular, deep learning-based approaches have been combined with clustering-based methods~\cite{zhang2021refining, siméoni2023unsupervised, shin2022selfmask, wang2023cut}, generative models~\cite{voynov2021object, bielski2019emergence,liu2019generative}, and object-level grouping~\cite{shin2022namedmask, bai2022point}.
{\color{black}Two of the methods} we build upon in our experiments, LOST~\cite{simeoni2021lost} and TokenCUT~\cite{wang2022tokencut}, employ clustering methods on top of the DINO network~\cite{caron2021emerging}, {\color{black} while MOVE~\cite{bielski2022move} uses a segmentation head on top of DINO representation}.

In the phrase grounding task, text phrases are associated with specific image locations~\cite{zhang2018top,javed2018learning}.  When relying on weakly supervised learning, the locations are not given during training, only during test time~\cite{akbari2019multi}. A common way to link the phrase to the image is to embed both the text and image patches in a shared embedding space~\cite{chiu2016named, sarzynska2021detecting, kamath2021mdetr}. Recent contributions employ CLIP~\cite{radford2019language} for linking text with image locations since it has powerful text and image encoders and relies on weakly supervised training~\cite{li2021grounded, shaharabany2022looking}. It can, therefore, be used both to represent the text and to obtain a training signal for the phrase grounding network. 

We are not aware of other work in which one network $f$ trains another network $h$, which in turn is used to refine the first network. There are contributions in which two networks are trained symbiotically at the same time. For example, for the task of semi-supervised semantic segmentation, two differently initialized networks were trained jointly, with each network creating pseudo-labels for the other~\cite{chen2021semi}. The DINO unsupervised representation learning method~\cite{caron2021emerging} employs a self-distillation process in which the teacher is a combination of frozen student networks.

The role of $h$ in propagating a detection-based loss back to $f$ is reminiscent of other cases in which a network is used for the purpose of supervising another, e.g., GANs~\cite{goodfellow2020generative}. In other cases, an auxiliary network can be trained in a supervised way to provide a differentiable approximation of an indifferentiable black box~\cite{polyak2018unsupervised}.

\section{The Phrase Grounding Method}

While we apply the same method for multiple applications, each application relies on a different configuration of baseline networks. Therefore, to minimize confusion, we first focus on phrase grounding. Applying our method to unsupervised object discovery is explored in Sec.~\ref{sec:object_discovery}.

In phrase grounding, we refine a pre-trained localization model ($g$) using a detection model ($h$) that we add. $h$ is trained based on $g$ and then the predictions of $h$, now serving as a teacher, are used to finetune network $g$, which becomes the student. This cyclic process is illustrated in Fig.~\ref{fig:arch} and serves to make $g$ more spatially coherent, see Fig.~\ref{fig:teaser}(d-f).

The phrase grounding network $g$ is based on an encoder-decoder architecture adapted to support text-based conditioning~\cite{shaharabany2022looking}. The input signals are (i) a text $t$ and (ii) an RGB image $I \in  R^{3 \times W \times H}$. It outputs a localization heatmap $M$ that identifies image regions in $I$ that correspond to the part of the scene described by $t$.
\begin{equation}
M = g(I, Z_{t}(t))\,,
\end{equation}
where $M \in  R^{W \times H}$ contains values between 0 and 1, and $Z_{t}(t)$ is a text embedding of the input text $t$, given by the text encoder of CLIP \cite{radford21}. Our refinement algorithm uses $g$ with the pre-trained weights published by  \cite{shaharabany22}.

Our method trains a model $h$ to generate a set of bounding boxes $\bar{B}$ that match the localization map $M$. 
\begin{equation}
\bar B = h(M)
\end{equation}

Thus $h$ provides a feedforward way to generate bounding boxes from $M$. The alternative provided, for example, by \cite{shaharabany22} is a multi-step process in which $M$ is first converted to a binary mask  by zeroing out any pixel value lower than half the mask's max value \cite{qin19,choe2020evaluating,choe2021region}. Next, contours are extracted from the binary mask using the method of \cite{suzuki1985topological}. For each detected contour, a bounding box is extracted, whose score is given by taking the mean value of $M$ for that bounding box. Finally, a non-maximal suppression is applied over the boxes with an overlap of at least 0.05 IOU, filtering out low-score boxes (0.5 of the maximal score). 

$h$ replaces this process with a single feed-forward pass. However, its main goal is to provide a training signal for refining $g$. This is done by considering the output of $h$ as foreground masks and considering the values of $g$'s output inside and outside these masks.

\subsection{Training $h$} \label{ssec.h_train_pg}

The network $h$ is trained to predict a fixed number $k$ of bounding boxes $\bar{B}$.  Each box is represented as a vector $b_i\in \mathbb{R}^6$ that contains the center coordinates of the box, its width, and its height. In addition, the network $h$ contains a logit value, which denotes whether there is an expected object within each box. 

Training is performed maintaining the semi-supervised nature of the phrase grounding method. The bounding boxes used for training $h$ are extracted using network $g$ and the method of Suzuki et~al\cite{suzuki1985topological}, as explained above. We call the set of resulting bounding boxes $B$. 

Following Carion et~al. \cite{carion20}, we train $h$ using a loss $L_{h}$ that has three terms: (1) a classification loss $L_\text{cls}$, (2) an $l1$ loss $L_\text{box}$, and (3) the GIoU\cite{Rezatofighi19} loss $L_\text{giou}$. 

If the number of objects $k$ returned by $h$ is smaller than the number of target boxes $|B|$, the $k$ boxes with the highest confidence are used. In the opposite case, $B$ is padded with zero-coordinate vectors with a ``no object'' label.

For computing the loss, one assumes a one-to-one correspondence between the ground truth objects and the detected boxes. This matching is obtained by minimizing $L_h$ over all possible permutations, using the Hungarian algorithm \cite{Kuhn55} for minimal cost bipartite matching. 
Denote as $B'=[b_0',b_1',...,b_{k-1}']$ the matrix that holds the set of boxes $B$ ordered optimally.

The classification loss $L_{cls}$ is a Negative log-likelihood loss 
\begin{equation}
L_\text{cls} =  \sum_{\substack{i=0}}^{k-1} {-\log {\bar p_i}}
\end{equation} 
where $\bar p_i$ is the predicted box logit, representing the probability of the existence of an object. 

$L_{box}$ is applied directly to the coordinates of the centers of the bounding boxes, their height and width:
\begin{equation}
L_\text{box} =  \sum_{\substack{i=0}}^{k-1} {\| b_i' - \bar{b_i} \|_1}
\end{equation}

While the loss $L_{box}$ is affected by the size of the box, the 2nd loss, $L_{giou}$, is a scale-invariant loss given by 
\begin{equation}
L_\text{giou}(B', \bar{B}) = \mathlarger{\mathlarger{\sum}}_{\substack{i=0}}^{k-1}{ 1 - \left(
 \frac{\bigl| \bar{b_i} \cap b_i' \bigr| }{ \bigl| \bar{b_i} \cup b_i' \bigr| } - \frac{\bigl| c_i \setminus (\bar{b_i} \cup b_i')\bigr|}{\bigl| C_i \bigr|} 
 \right)}
\end{equation}
where $c_i$ is the smallest box containing $b'_i$ and $\bar{b_i}$.
All losses are normalized by the number of boxes. 

The final loss is a weighted sum of all three losses:
\begin{equation}
\begin{split}
 L_{h}(B',\bar{B}) = \lambda_1 *L_\text{cls}(B',\bar{B}) + \lambda_2 *L_\text{box}(B',\bar{B}) + \\ 
 \lambda_3 *L_\text{giou}(B',\bar{B})
 \end{split}
\end{equation}
where $\lambda_1=2, \lambda_2=5, \lambda_3=2$. These weights are similar to those used in previous work, with an extra emphasis on $\lambda_1$ (using a value of 2 instead of 1), but there was no attempt to optimize them beyond inspecting a few training images.

\subsection{Refining $g$}\label{ssec.g_refine_pg}

For finetuning $g$, we use the multiple loss terms, including the same loss terms that are used for training $h$, with a modification. Here, instead of just calculating the loss between two sets of boxes, we also compute the union box of ground truth boxes: $BU = Union(B)$. With probability $0.5$ we use $BU$ instead of $B$ for calculating the loss (in this case, the matching is done with a single box only)

\begin{equation}
L_{h_{BU}} = \begin{cases}
         L_{h}(BU, \bar{B}), & \text{if } p \geq 0.5 \\
         L_{h}(B, \bar{B}), & \text{otherwise}
       \end{cases}, p \sim \text{Uniform}[0,1]
\end{equation}

In addition to the bounding box loss, we use losses for the localization maps used by~\cite{shaharabany22} to train $g$. This prevents the fine-tuned model from following $h$ ``blindly'', without considering the underlying data. 

The relevancy map loss, uses a CLIP-based relevancy 
~\cite{chefer2021transformer} to provide rough estimation for the localization map
\begin{equation}
L_\text{rmap}(I,H) = \| H - g^h(I,Z^T) \|^2,
\end{equation}
where $H$ is the relevancy map and $g^h$ is the refined network $g$. The foreground loss $L_{fore}(I,T)$ is given by 
\begin{equation}
  L_\text{fore}(I,t) = - CLIP(g^h(I,Z^T)\odot I, t),
\end{equation}
where $\odot$ is the Hadamard product. The loss maximizes the similarity given by CLIP between the mask's foreground region and the input text $t$. On the other hand, the background loss $L_{back}(I,t)$ minimizes the similarity CLIP distance between the background and text $t$
\begin{equation}
   L_{back}(I,t) = CLIP((1-g^h(I,Z^T))\odot I, t),
\end{equation}

\noindent The overall loss is given by:
\begin{equation}
\notag
\begin{split}
 L_{g} = L_{h_{BU}} + \lambda_4 * L_{reg}(I,g^h) + 
 \lambda_5 *L_\text{rmap}(I,H) + \\
 \lambda_6 *L_\text{back}(I,T) + 
 \lambda_7 *L_\text{fore}(I,T) 
\end{split}
\end{equation} 
where $\lambda_4=1, \lambda_5=64, \lambda_6=2 , \lambda_7=1$. These hyperparameters reflect the values assigned by previous work, multiplied by 4 in order to approximately balance the loss that arises from $h$ with the other loss terms. 

\paragraph{Architecture}\label{ssec.h_arch_pg}
$h$ is a VGG16 \cite{simonyan2014very}, pre-trained on the ImageNet\cite{imagenet} dataset. In order to apply it to the single channel heatmap $M \in  R^{\times W \times H}$, this input is repeated three times across the channel dimension. The last layer of the classifier is replaced by a linear layer of dimensions $4096 \times (6k)$, $k$ being the number of boxes predicted by $h$. 

\section{Unsupervised object discovery}\label{sec:object_discovery}

For the task of unsupervised object discovery, a vision transformer $f$ is pretrained in a self-supervised manner, using DINO~\cite{caron2021emerging}. It is then used to extract features $F$ from an input image $I \in  R^{3 \times W \times H}$

\begin{equation}
F = \bar f(I) 
\end{equation}
where $\bar f$ denotes the latent variables from the transformer $f$. $F \in  R^{d \times N}$, where $d$ is the features dimension and $N$ denotes the number of patches for $f$.  For each patch $p$, we denoted by $f_{p} \in  R^{d}$ the associated feature vector. Bounding boxes based on these features are extracted using unsupervised techniques, such as LOST~\cite{simeoni2021lost}, TokenCut~\cite{wang2022tokencut} {\color{black} or MOVE~\cite{bielski2022move}}.  

\smallskip
\label{lost_algorithm}
\noindent{\bf LOST} builds a patch similarities graph $\graph$, with a binary symmetric adjacency matrix $A \,{=}\, (a_{pq})_{1 \leq p, q\leq N} \in \{0, 1\}^{N \times N}$
where
\begin{align}
a_{pq} = 
\left\{
    \begin{array}{ll}
    1 & \text{if } f_p^\top{f_q} \geq 0, \\
    0 & \text{otherwise}.
    \end{array}
\right.
\end{align}

An initial seed $p*$ is selected as the patch with the smallest number of connections to other patches. 
\begin{align}
    p^* = \argmin_{p \in \{1, \ldots, N\}} d_p \text{~~~where~~~} d_p = \sum_{q=1}^{N} a_{pq}.
\end{align}

This is based on the assumptions that connectivity implies belonging to the same object, since patch embeddings are similar for the same object, and that each object occupies less area than the background.

Denote the list of $a$ patches with the lowest degree $d_p$  as $\mathcal{D}_a$. LOST then considers the subset of $\mathcal{D}_a$ that is positively correlated, in the embedding space, with $p^*$
\begin{equation}
\mathcal{S} = \{q \in \mathcal{D}_a |  f_q^\top{f_{p^*}} \geq 0\} 
\end{equation}
This set is then expanded obtaining
\begin{equation}
\mathcal{S}^+ = \{q | \sum_{p\in \mathcal S} f_q^\top{f_{p}} \geq 0\}    
\end{equation}

We note that in the image itself, the patches of $\mathcal{S}^+$ can be part of multiple separate regions. The method selects the connected component (4-connectivity in the image space) in $\mathcal{S}^+$ that contains the seed $p^*$ as its single discovered object. 

\smallskip
\textbf{TokenCut}\cite{wang2022tokencut} employs a slightly different adjacency matrix, $A$, which employs the cosine similarity score between pairs of feature vectors. 
\begin{align}
	A{p,q} =
	\begin{cases} 
		1,  & \mbox{if }\frac{f_p^\top f_q}{\lVert f_p \rVert_2 \lVert f_q \rVert_2} \ge \tau \\
		\epsilon, & \mbox{else}
	\end{cases}\,,
\end{align}
where $\tau = 0.2$ and $\epsilon = 1e-5$.

The normalized cut method~\cite{shi2000normalized}  is applied to the graph to achieve object discovery. This method clusters all patches into two groups, based on the 2nd smallest eigenvector of the normalized adjacency matrix, and selects the group with the maximal absolute value in this eigenvector. The bounding box of the patches in this group is returned.

\smallskip
{\color{black}\label{move_algorithm}
\textbf{MOVE}\cite{bielski2022move}, in contradistinction to the preceding two methodologies, employs a segmentation network that is trained atop the latent transformer features denoted as $F$. The resulting output of this network takes the form of a segmentation map denoted as $M \in  R^{W \times H}$. Subsequently, this segmentation map undergoes binarization with a threshold set at 0.5, followed by the detection of connected components ~\cite{bolelli2019connectedcomponents}. The most sizable bounding box is then selected to correspond to the most extensive connected component.}

\subsection{Training $h$ and refining $f$} \label{ssec.h_train_odis}
The training process of detector $h$ follows the details described in Sec.~\ref{ssec.h_train_pg}, with a few minor changes. There is a single ground-truth bounding box $B$, extracted from an image $I$ by model $f$ using the unsupervised techniques described above. Using the same loss term $L_{h}$, $h$ is optimized to minimize $L_{h}(B,\bar B)$, where $\bar B$ are the $k$ predicted boxes. 

To maintain the unsupervised nature of the task, $h$ is initialized with weights from the self-supervised method DINO\cite{caron2021emerging}, using a ResNet-50\cite{resnetcvpr2016} backbone. 
In the phrase grounding case {\color{black}and MOVE~\cite{bielski2022move}}, the input of $h$ is the map $M$, and the analogue for {\color{black}non-trainable} unsupervised object discovery is the map $F$ {\color{black}where such map $M$ is missing}.

For refining the DINO-trained transformer model $f$, we use the same loss term $L_{h}$ as is used in phrase grounding and add loss terms to prevent it from diverging too far. While in phrase grounding we used the loss terms that were used to train the phrase grounding network, here, for runtime considerations, we explicitly keep the transformer $f$ in the vicinity of the DINO-pretrained network. 

The loss term is defined as the distance between the output of $f$ and that of the refined model $f^h$
\begin{align}
	L_{f}(I) = \| f(I) - f^h(I) \|^2, 
\end{align}

Both methods \cite{simeoni2021lost,wang2022tokencut} are  improved by training a Class Agnostic Detector (CAD) on the extracted bounding boxes.
Faster R-CNN~\cite{renNIPS15fasterrcnn} is used for CAD, with the \textit{R50-C4} model of Detectron2~\cite{wu2019detectron2} based on a ResNet-50\cite{resnetcvpr2016} backbone. This backbone is pre-trained with DINO self-supervision. Following this process, we train an identical CAD using the refined model $f^{h}$. Note that CAD and our method are complementary. While both train with the same pseudo-labels, CAD is trained on the original image and cannot backpropagate a loss to the underlying network $f$.

\begin{figure}[t]
    \setlength{\tabcolsep}{3pt} 
    \renewcommand{\arraystretch}{1} 
    \centering
    \begin{tabularx}{\textwidth}{@{}p{.1cm}@{~}c@{}c@{}c@{~}c@{}}

&\rotatebox{90}{\small a man}&
     \includegraphics[valign=B,width=0.324\linewidth]{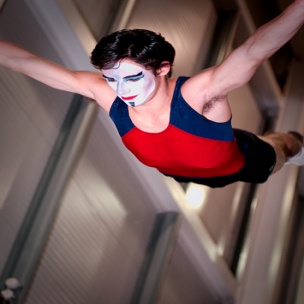} &
    \includegraphics[valign=B,width=0.324\linewidth]{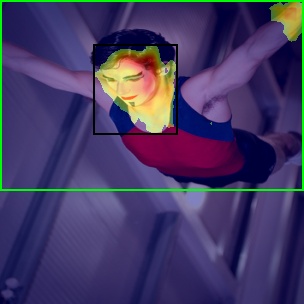} &
    \includegraphics[valign=B,width=0.324\linewidth]{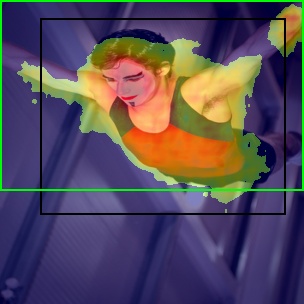} \\

&\rotatebox{90}{\small a mountain biker}&
   \includegraphics[valign=B,width=0.324\linewidth]{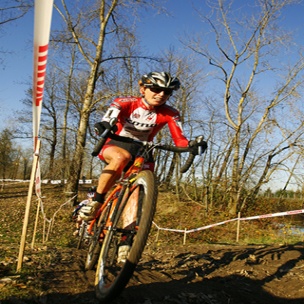} &
  \includegraphics[valign=B,width=0.324\linewidth]{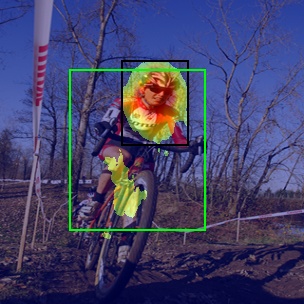} &
  \includegraphics[valign=B,width=0.324\linewidth]{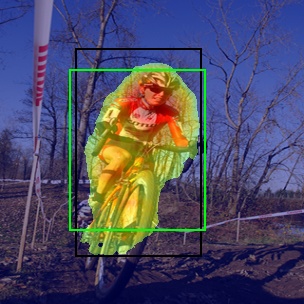} \\

&\rotatebox{90}{\small several individuals}&
     \includegraphics[valign=B,width=0.324\linewidth]{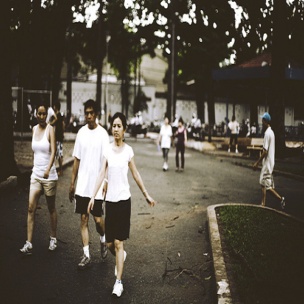} &
    \includegraphics[valign=B,width=0.324\linewidth]{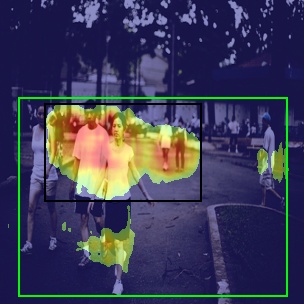} &
    \includegraphics[valign=B,width=0.324\linewidth]{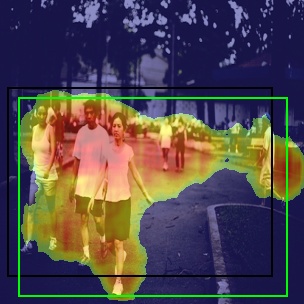} \\

&\rotatebox{90}{\small a boy}&
     \includegraphics[valign=B,width=0.324\linewidth]{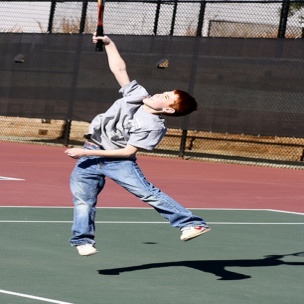} &
    \includegraphics[valign=B,width=0.324\linewidth]{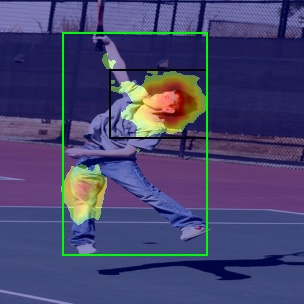} &
    \includegraphics[valign=B,width=0.324\linewidth]{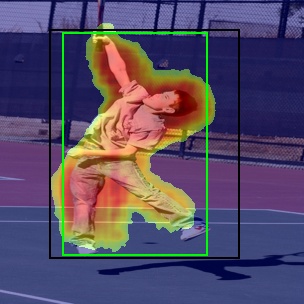} \\

&\rotatebox{90}{\small muzzles}&
     \includegraphics[valign=B,width=0.324\linewidth]{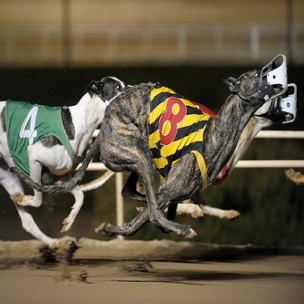} &
    \includegraphics[valign=B,width=0.324\linewidth]{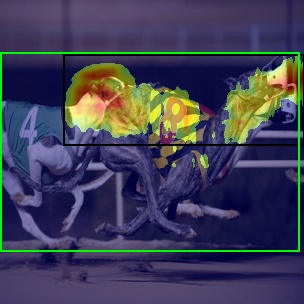} &
    \includegraphics[valign=B,width=0.324\linewidth]{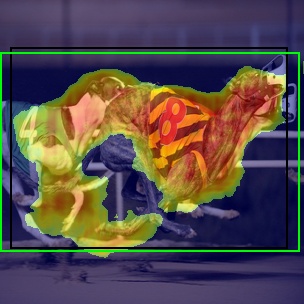} \\

&\rotatebox{90}{\small a very young girl}&
     \includegraphics[valign=B,width=0.324\linewidth]{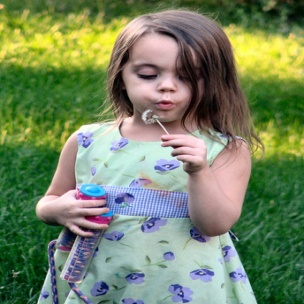} &
    \includegraphics[valign=B,width=0.324\linewidth]{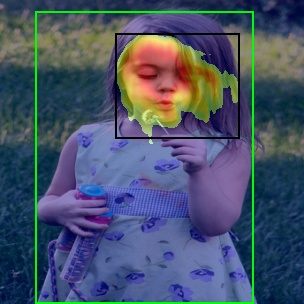} &
    \includegraphics[valign=B,width=0.324\linewidth]{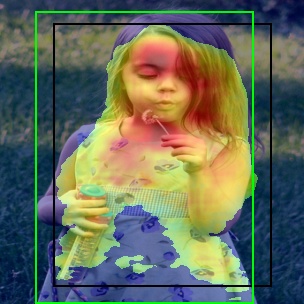} \\

      & (a) & (b) & (c) &(d)\\
    \end{tabularx}
    \caption{Sample phrase-grounding results. where (a) the phrase (b) the input image (c) results (black) for network $g$ \cite{shaharabany22} compared to ground-truth box (green) (d) same for refined network $g^{h}$.}
    \label{fig:compare_hm}
\vspace{-3mm}
\end{figure}

\begin{figure*}[h!]
    \setlength{\tabcolsep}{3pt} 
    \renewcommand{\arraystretch}{1}
     \centering
\begin{tabular}{ccccc}
    \includegraphics[valign=B,width=.170860\linewidth]{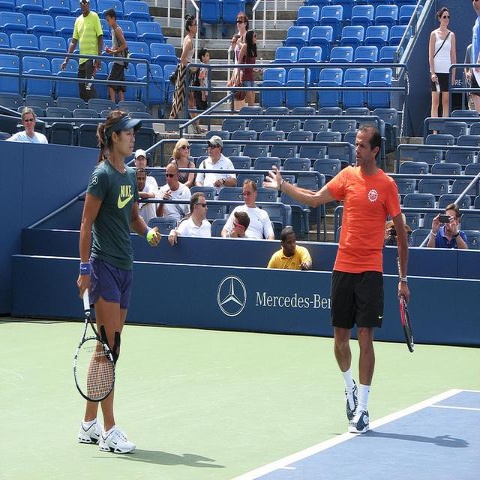} &
    \includegraphics[valign=B,width=.170860\linewidth]{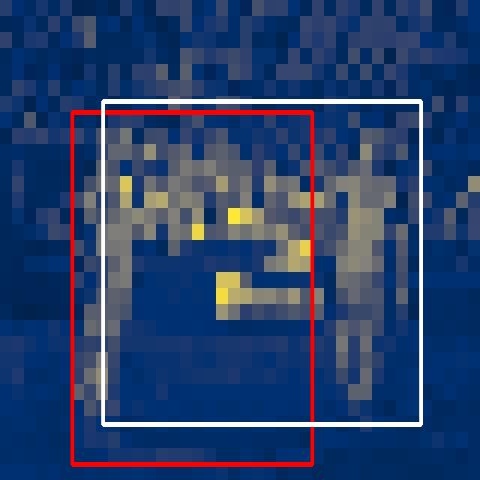} &
    \includegraphics[valign=B,width=.170860\linewidth]{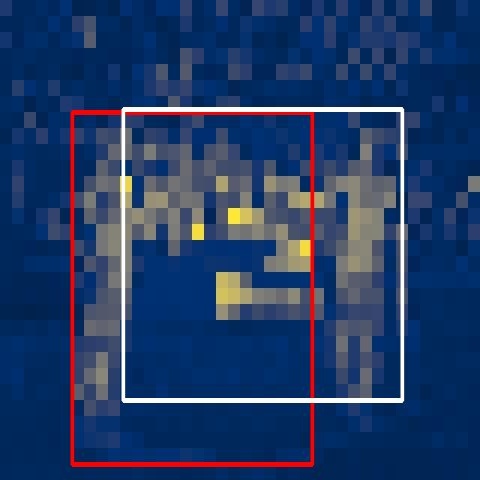} &
 \includegraphics[valign=B,width=.170860\linewidth]{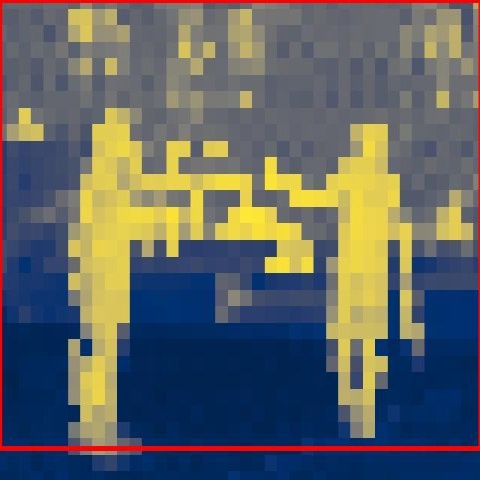} &
    \includegraphics[valign=B,width=.170860\linewidth]{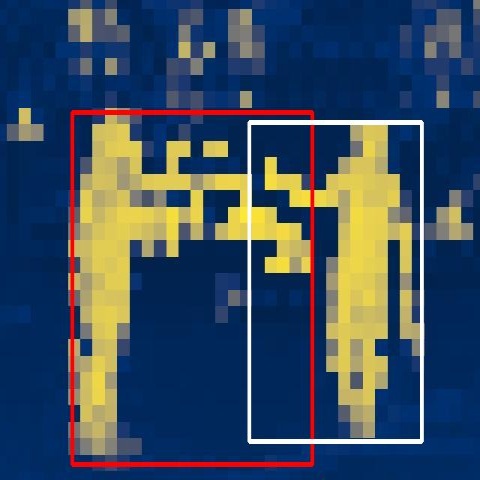} \\
         \includegraphics[valign=B,width=.170860\linewidth]{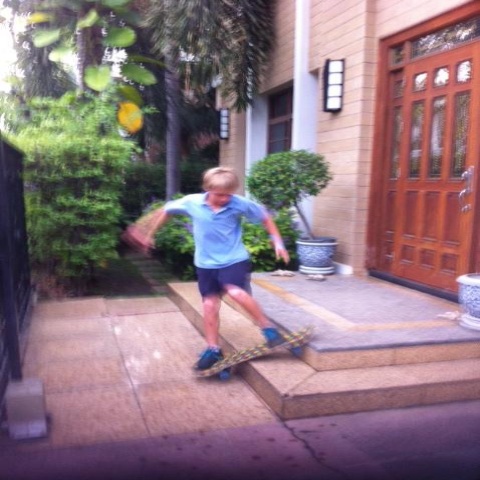} &
    \includegraphics[valign=B,width=.170860\linewidth]{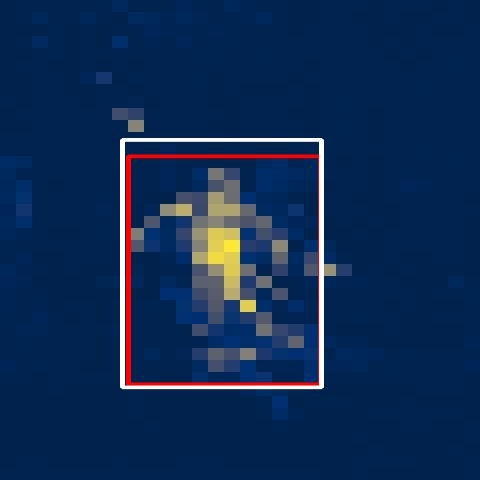} &
    \includegraphics[valign=B,width=.170860\linewidth]{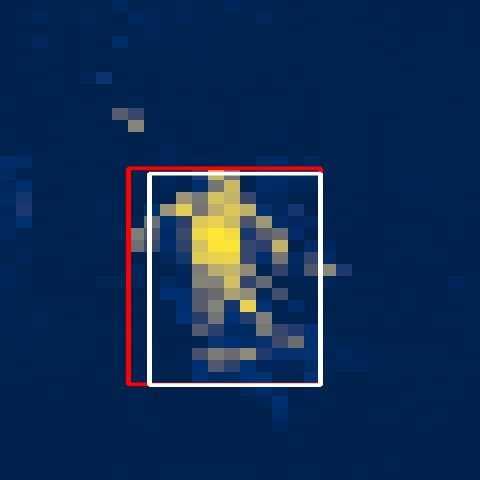} &
 \includegraphics[valign=B,width=.170860\linewidth]{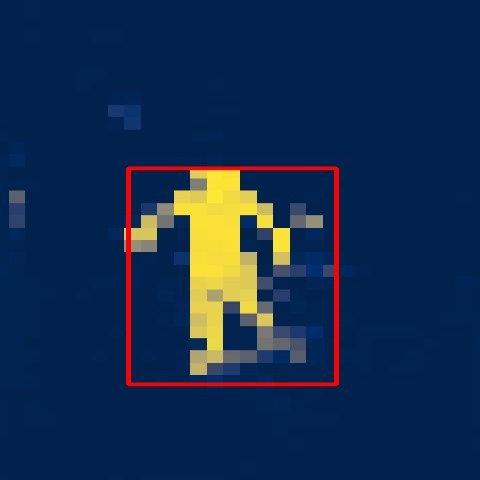} &
    \includegraphics[valign=B,width=.170860\linewidth]{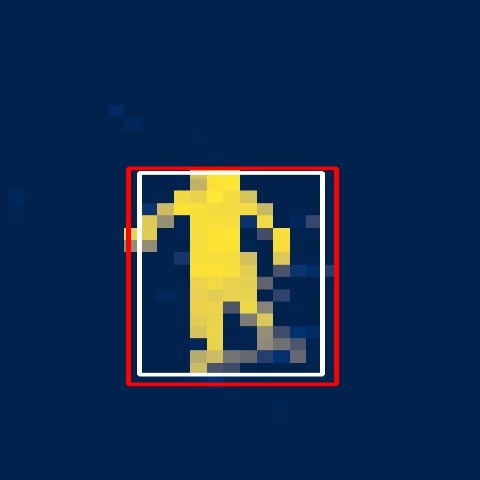} \\
 \includegraphics[valign=B,width=.170860\linewidth]{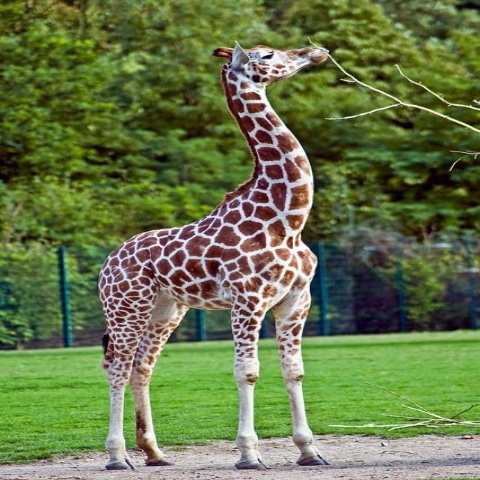} &
    \includegraphics[valign=B,width=.170860\linewidth]{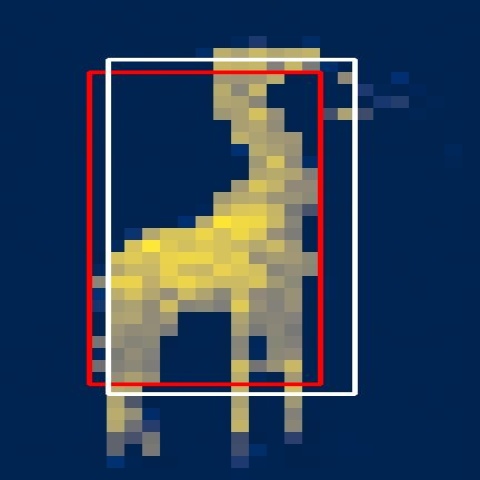} &
    \includegraphics[valign=B,width=.170860\linewidth]{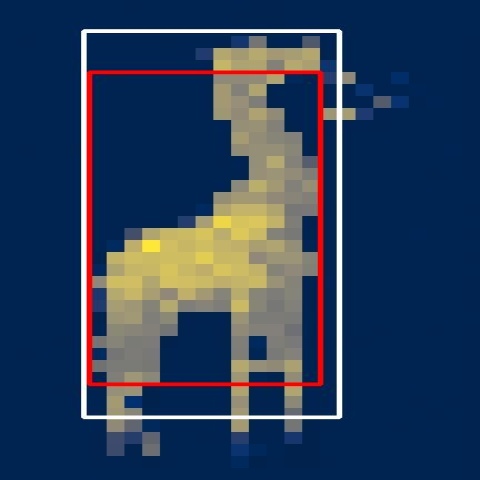} &
 \includegraphics[valign=B,width=.170860\linewidth]{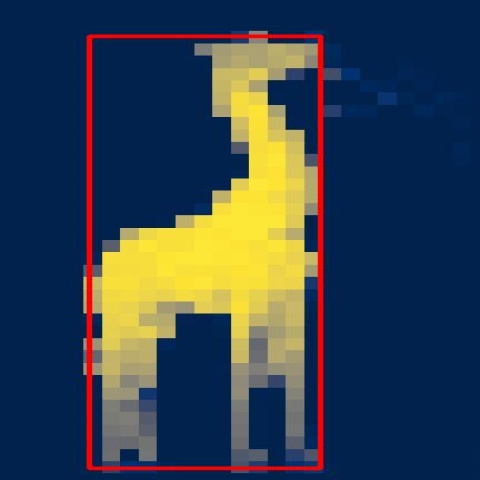} &
    \includegraphics[valign=B,width=.170860\linewidth]{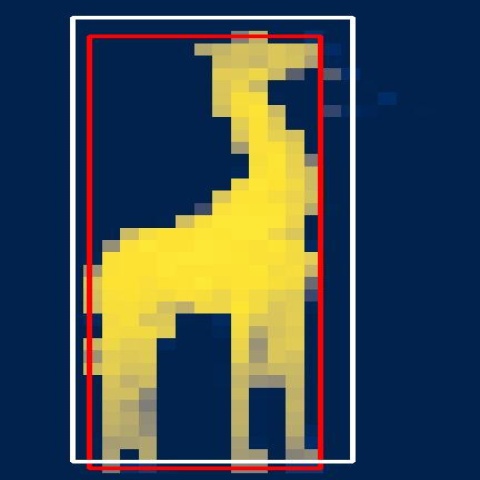} \\
    \includegraphics[valign=B,width=.170860\linewidth]{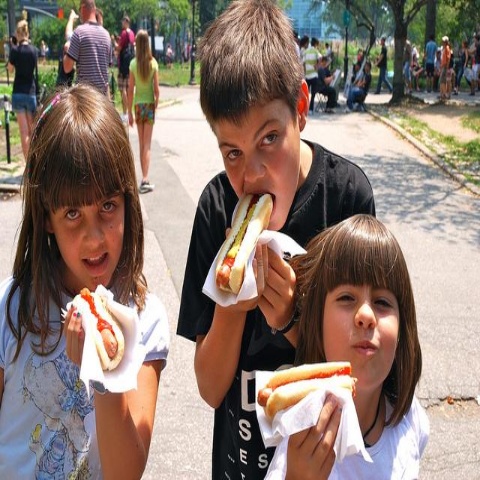} &
    \includegraphics[valign=B,width=.170860\linewidth]{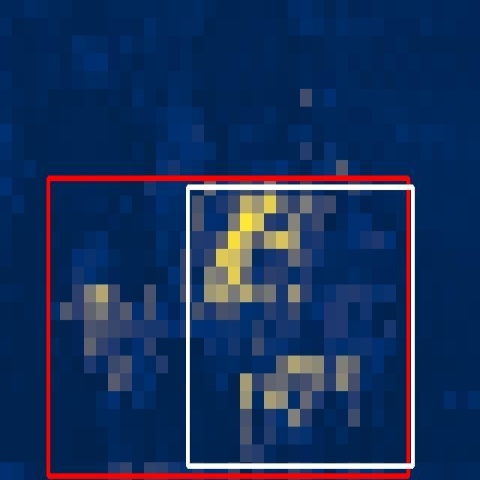} &
    \includegraphics[valign=B,width=.170860\linewidth]{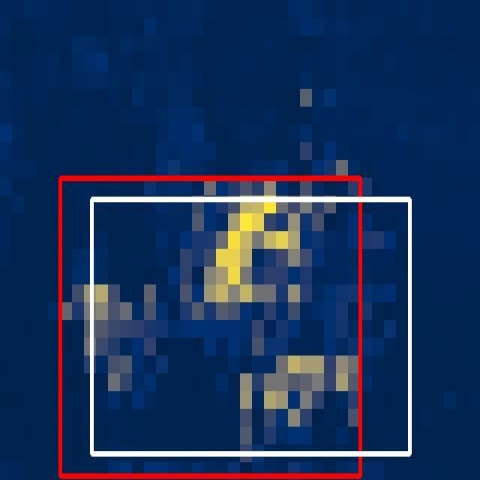} &
 \includegraphics[valign=B,width=.170860\linewidth]{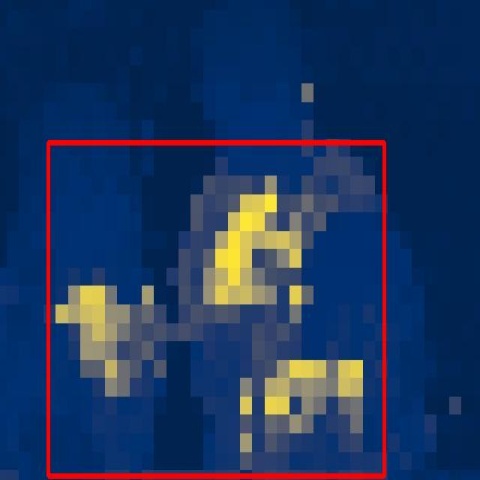} &
    \includegraphics[valign=B,width=.170860\linewidth]{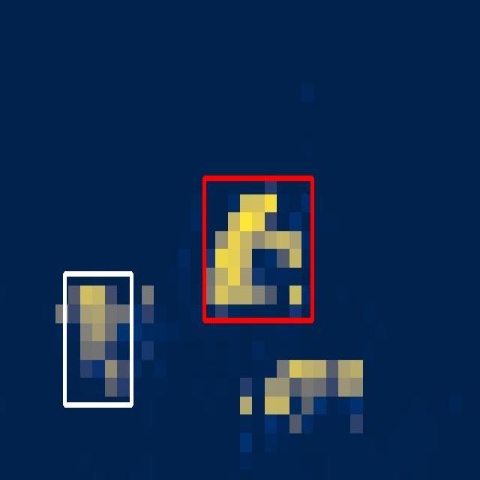} \\

      (a) & (b) & (c) & (d) & (e)\\
    \end{tabular}
    \caption{Single object discovery results. (a) the input image, (b) the inverse degree of the LOST~\cite{simeoni2021lost} graph obtained over $f$  (published model); the red bounding box is directly from LOST, the white is the prediction of CAD trained on top of it (c) same with our refined model $f^h$ and LOST (d) same as b, but using $f$ together with TokenCut\cite{wang2022tokencut}, (using the published weights; the CAD model was not released and is not shown) (e) the results of $f^h$ and TokenCut.
    }
    \label{fig:object_discovery_f_fh_cad}
    \vspace{-2mm}
\end{figure*}

\begin{table}
\vskip 0.15in
\begin{center}

\footnotesize
\setlength{\tabcolsep}{4pt}
\begin{tabular}{@{}l@{~}l@{~}c@{~}c@{~}cc@{~}c@{~}c@{}}
\toprule
 \multirow{2}{*}{Method} & \multirow{2}{*}{Backbone} & 
 \multicolumn{3}{c}{VG trained} & \multicolumn{3}{c}{ MS-COCO trained}\\ 
 \cmidrule(rl){3-5}
 \cmidrule(rl){6-8}
 &  & VG & Flickr & ReferIt & VG & Flickr & ReferIt\\ 
 \midrule
 Baseline & Random &  11.15 & 27.24 & 24.30 & 11.15 & 27.24 & 24.30 \\  
 Baseline & Center &  20.55 & 47.40 & 30.30 & 20.55 & 47.40 & 30.30 \\
 GAE \cite{Chefer_2021_ICCV} & CLIP & 54.72 & 72.47 & 56.76 & 54.72 & 72.47 & 56.76\\
 \midrule
 FCVC \cite{fang2015captions} & VGG  &  - & - & - & 14.03 & 29.03 & 33.52\\
  VGLS \cite{xiao2017weakly} & VGG  &  - & - & - & 24.40 & - & - \\
 TD \cite{zhang2018top} & Inception-2 &  19.31 & 42.40 & 31.97 &  - & - & - \\
 SSS \cite{javed2018learning} & VGG &  30.03 & 49.10 & 39.98 &  - & - & - \\
 MG \cite{akbari2019multi} & BiLSTM+VGG &  50.18 & 57.91 & 62.76 & 46.99 & 53.29 & 47.89 \\
 MG \cite{akbari2019multi} & ELMo+VGG &  48.76 & 60.08 & 60.01 & 47.94 & 61.66 & 47.52 \\
 GbS \cite{arbelle2021detector} & VGG &   53.40 & 70.48 & 59.44& 52.00 & 72.60 & 56.10 \\
 WWbL~\cite{shaharabany22} & CLIP+VGG &  62.31 & 75.63 & 65.95 & 59.09 & 75.43 & 61.03 \\
Ours & CLIP+VGG &  \textbf{63.51} & \textbf{78.32} & \textbf{67.33} & \textbf{60.05} & \textbf{77.19} & \textbf{63.48} \\ \bottomrule
 \end{tabular}

\end{center}
\caption{Phrase grounding results: ``pointing game" accuracy on Visual Genome (VG), Flickr30K, and ReferIt. The methods in the first three rows do not train.}
\label{tab:grounding_pointing}
\end{table}

\begin{table}
    \centering
    \begin{center}
    \begin{small}
    \begin{tabular}{ llccc }
    \toprule
     \multirow{2}{*}{Training} & \multirow{2}{*}{Model} & \multicolumn{3}{c}{ Test Bbox Accuracy}\\ \cmidrule(l){3-5} & & VG & Flickr & ReferIt \\
      
     \midrule
    \multirow{3}{*}{MS-COCO} & MG \cite{akbari2019multi} & 15.77 & 27.06 & 15.15 \\
    & WWbL \cite{shaharabany22} & 27.22 & 35.75 & 30.08 \\
    & Ours & \textbf{28.77}{\tiny(27.1)} & \textbf{47.26}{\tiny(45.01)} & \textbf{30.63}{\tiny(29.05)} \\
     \midrule
    \multirow{3}{*}{VG} & MG \cite{akbari2019multi}  & 14.45 & 27.78 & 18.85 \\
     & WWbL \cite{shaharabany22} & 27.26 & 36.35 & 32.25 \\
    & Ours & \textbf{31.02}{\tiny(29.23)} & \textbf{42.40}{\tiny(44.91)} & \textbf{35.56}{\tiny(34.56)} \\
     \bottomrule
    \end{tabular}
    \end{small}
    \end{center}
   \caption{Phrase grounding results: bounding box accuracy on Visual Genome (VG), Flickr30K, and ReferIt. {\color{black} The outcomes obtained from network $h$ are presented within brackets.}}
    \label{tab:grounding_res}
    \end{table}

\begin{table}
    \centering
    \begin{center}
    \begin{small}
    \begin{tabular}{@{}l@{~~}c@{~}c@{~}c@{~}c@{~}c@{~}c@{~}c@{}}
    \toprule
     \multirow{2}{*}{\rotatebox{90}{\footnotesize{Train set}}} & \multirow{2}{*}{Model} & \multicolumn{3}{c}{ Test point Accuracy} & \multicolumn{3}{c}{ Test Bbox Accuracy}\\ \cmidrule(l){3-5} \cmidrule(l){6-8} & & VG & Flickr & ReferIt & VG & Flickr & ReferIt \\
      
     \midrule
    \multirow{3}{*}{\rotatebox{90}{COCO}} & MG \cite{akbari2019multi} & 32.91 & 50.154 & 36.34 & 11.48 & 23.75 & 13.31 \\
    & WWbL \cite{shaharabany22} & 44.20 & 61.38 & 43.77 & 17.76 & 32.44 & 21.76 \\
    & Ours & \textbf{46.29} & \textbf{63.43} & \textbf{44.59} & \textbf{22.32} & \textbf{38.00} & \textbf{22.91} \\
     \midrule
    \multirow{3}{*}{\rotatebox{90}{VG}} & MG \cite{akbari2019multi}  & 32.15 & 49.48 & 38.06 & 12.23 & 24.79 & 16.43 \\
     & WWbL \cite{shaharabany22} & 43.91 & 58.59 & 44.89 & 17.77 & 31.46 & 18.89 \\
    & Ours & \textbf{46.77} & \textbf{61.75} & \textbf{44.9} & \textbf{22.40} & \textbf{35.23} & \textbf{23.44} \\
     \bottomrule
    \end{tabular}
    \end{small}
    \end{center}
  \caption{WWbL results: bounding box accuracy on Visual Genome (VG), Flickr30K, and ReferIt.}
    \label{tab:WWbL_res}    
     \vspace{-2mm}
    \end{table}

\begin{table}
    \centering
    \begin{center}
    \begin{small}
    \begin{tabular}{ lccc }
    \toprule
     Model & VOC07 & VOC12 & MS-COCO \\
     \midrule
    Selective Search~\cite{Uijlings13} & 18.8 & 20.9 & 16.0 \\
    EdgeBoxes~\cite{zitnick2014edge} & 31.1 & 31.6 & 28.8 \\
    Kim et al.~\cite{kim2009unsupervised} & 43.9 & 46.4& 35.1 \\
    Zhang et al.~\cite{zhang2020object} & 46.2 & 50.5 & 34.8 \\
    DDT+~\cite{wei2019unsupervised} & 50.2 & 53.1 & 38.2 \\
    rOSD~\cite{vo2020toward} &  54.5 & 55.3 & 48.5 \\
    LOD~\cite{vo2021large}  & 53.6 & 55.1 & 48.5 \\
    DINO-seg~\cite{caron2021emerging} & 45.8 & 46.2 & 42.1 \\
    LOST~\cite{simeoni2021lost} & 61.9 & 64.0 & 50.7 \\
    Ours using LOST & 62.0{\tiny(42.1)} & 66.2{\tiny(53.5)} & 52.0{\tiny(33.7)}\\
    TokenCut~\cite{wang2022tokencut} & 68.8 & 72.1 & 58.8 \\
    Ours using TokenCut& 69.0{\tiny(44.6)} &  72.4{\tiny(54.1)} & 60.7{\tiny(39.5)} \\
    {\color{black}MOVE}~\cite{bielski2022move} & 76.0 & 78.8 & 66.6 \\
    {\color{black}Ours using MOVE}& \textbf{77.5}{\tiny(42.9)} &  \textbf{79.6}{\tiny(54.9)} & \textbf{67.2}{\tiny(48.3)} \\

  \midrule 
    LOD + CAD~\cite{simeoni2021lost} & 56.3 & 61.6 & 52.7 \\
    rOSD + CAD~\cite{simeoni2021lost} & 58.3 & 62.3 & 53.0 \\
    LOST + CAD~\cite{simeoni2021lost} & 65.7 & 70.4 & 57.5 \\
    Ours using LOST + CAD & 66.1 & 71.0 & 58.7\\
    TokenCut~\cite{wang2022tokencut} +CAD & 71.4 & 75.3  & 62.6 \\
    Ours using TokenCut + CAD & 71.9 & 75.6 & 64.4 \\
    {\color{black}MOVE}~\cite{bielski2022move} +CAD & 77.1 & 80.3  & 69.1 \\
    {\color{black}Ours using MOVE~\cite{bielski2022move} +CAD} & \textbf{78.7} & \textbf{81.3} & \textbf{69.3} \\
     \bottomrule
    \end{tabular}
    \end{small}
    \end{center}
        \caption{Object Discovery results: CorLoc score on MS-COCO20K, VOC07 and VOC12. Network $h$ was trained using pseudo labels from either LOST~\cite{simeoni2021lost}, TokenCut~\cite{wang2022tokencut} or MOVE~\cite{bielski2022move}. +CAD indicates training a second-phase   class-agnostic detector with model pseudo-boxes as labels. {\color{black} Network $h$ results are enclosed in brackets.}}
    \label{tab:od_tokencut}
   
    \end{table}

\begin{table}
    \centering
    \begin{center}
    \begin{small}
    \begin{tabular}{@{}l@{~~}c@{~}c@{~}c@{~~}c@{~}c@{~}c@{}}
    \toprule
     \multirow{2}{*}{Ablation} & \multicolumn{3}{c}{ Test point Accuracy} & \multicolumn{3}{c}{ Test Bbox Accuracy}\\ \cmidrule(l){2-4} \cmidrule(l){5-7} & VG & Flickr & ReferIt & VG & Flickr & ReferIt \\

     \midrule
    w/o Box Union & 57.26 & 72.54 & 62.55 & 25.11 & 28.74 & 24.63 \\
    w/o reg. & 53.49 & 68.47 & 61.92 & 26.45 & 42.79 & 29.74 \\
    k=1 & 56.84 & 70.74 & 62.15 & 27.75 & 32.35 & 24.73 \\
    Ours & \textbf{60.05} & \textbf{77.19} & \textbf{63.48} & \textbf{28.77} & \textbf{47.26} & \textbf{30.63} \\
     \bottomrule
    \end{tabular}
    \end{small}
    \end{center}
   \caption{Ablation study for the phrase grounding task. See text for details. All models were trained on MS-COCO14\cite{cocoeccv2014} dataset}
    \label{tab:ablation_pg}
    \end{table}

\begin{table}
    \centering
    \begin{center}
    \begin{small}
    \begin{tabular}{ lcccccc }
    \toprule
     {Ablation}  & VOC07 & VOC12 & MSCOCO20K \\
     \midrule
    w/o reg. & 61.72 & 64.45 & 50.13 \\
    k=1 & \textbf{62.54} & 64.67 & \textbf{52.00} \\
    k=5 & 62.16 & 64.45 & 51.70 \\
    k=10 & 61.92 & \textbf{66.16} & 51.98\\
    k=15 & 61.44 & 64.46 & 50.60 \\
     \bottomrule
    \end{tabular}
    \end{small}
    \end{center}
   {\color{black}\caption{Ablation study for the object discovery task.}}    \label{tab:ablation_object_discovery}
   \vspace{-0mm}
    \end{table}

\section{Experiments}
We present our results for three tasks: weakly supervised phrase grounding (WSPG), ``what is were by looking'' (WWbL), and unsupervised single object discovery. The first two use the same phrase grounding network $g$, and the third one is based on one of two techniques, which both utilize the same pre-trained transformer $f$. 

\noindent\textbf{Datasets} For WSPG and WWbL, the network $g$ is trained on either MSCOCO 2014~\cite{cocoeccv2014} or the Visual Genome (VG) dataset~\cite{krishna2017visual}. Evaluation is carried out on the test splits of Flickr30k\cite{plummer2015flickr30k}, ReferIt\cite{chen2017query, grubinger2006iapr} and VG~\cite{krishna2017visual}. 

VG contains 77,398, 5,000, and 5000 training, validation, and test images, respectively. Each image is linked to natural-language text and annotated bounding boxes.
During the training of MSCOCO2014 we use the training split defined by Akbari et al.~\cite{akbari2019multi}. It consists of 82,783 training samples and 40,504 validation samples, where each sample contains an image and five captions describing the image. 
ReferIt\cite{chen2017query, grubinger2006iapr} consists of 130k expressions referring to 99,535 objects in 20k images. For evaluation, we use the test split of Akbari et al.\cite{akbari2019multi}.
The dataset Flickr30k Entities \cite{plummer2015flickr30k} consists of 224K phrases that depict objects present in more than 31K images, with each image having five corresponding captions. The evaluation is carried out on a the test split of Akbari et al.\cite{akbari2019multi}.
For unsupervised single object discovery, the network $g$ is trained on either MSCOCO 20K, PASCAL-VOC07\cite{pascalvoc2007} or PASCAL-VOC12\cite{pascalvoc2012}. MS-COCO20K has 19,817 images chosen at random from the MSCOCO 2014 dataset\cite{cocoeccv2014}. VOC07 and VOC12 contain 5,011 and 11,540 images respectively, with each image belonging to one of 20 categories. For evaluation, we follow common practice and evaluate the train/val datasets. This evaluation is possible since the task is fully unsupervised.

\noindent\textbf{Implementation details} 
For phrase grounding tasks, the proposed network $h$ backbone is VGG16 \cite{simonyan2014very}, pre-trained on the ImageNet\cite{imagenet} dataset. For the object discovery task, we use $h$ with ResNet-50\cite{resnetcvpr2016} backbone, pre-trained with DINO\cite{caron2021emerging} self-supervision on the ImageNet\cite{imagenet} dataset. For both tasks, $h$ predicts $k=10$ bounding boxes. Refining takes place using an Adam optimizer with a batch size of 36. The learning rate of $h$ is 1e-5, while the learning rates of $g^h$ and $f^h$ are 1e-7 and 5e-7, respectively. The optimizer weight decay regularization is 1e-4. For the first 3000~ iterations, network $h$ is optimized, where $g^h/f^h$ is fixed. Then, for the rest of the training (10k~ iterations), $h$ is fixed while $g^h/f^h$ is optimized. 

\smallskip
\noindent\textbf{Metrics} 
Phrase grounding tasks are evaluated with respect to the accuracy of the pointing game\cite{zhang2018top}, which is calculated based on the output map by finding the 
location of the maximum value, given a query, and checking whether this point falls within the object's region. 

The ``BBox accuracy'' metric extracts a bounding box, given an output mask, and compares it with the ground-truth annotations. A prediction is considered accurate if IOU between the boxes is larger than 0.5. To extract the bounding box from an output map $M$, the procedure of Shaharabany et al.~\cite{shaharabany22} is employed. First, $M$ is binarized using a threshold of 0.5, then contours are extracted from $M$ using the method of Suzuki et al.~\cite{suzuki1985topological}. Based on the contours, a set of bounding boxes is derived by taking the smallest box containing each contour. These bounding boxes are scored by summing the values of M within the contour while ignoring boxes with low scores. Next, a non-maximal suppression process is applied and the minimal bounding box that contains the remaining bounding boxes is chosen. 

The WWbL task is an open-world localization task, with only an image as input (no text input). Using this image, the goal is to both localize and describe all of the elements in the scene. To solve this task, a multi-stage algorithm was introduced by Shaharabany et al.~\cite{shaharabany22}, starting with obtaining object proposals using selective search~\cite{Uijlings13}. Next, BLIP is used to caption these regions.  Captions that are similar to each other are removed using the Community Detection (Cd) clustering method~\cite{blondel2008fast}. Using the learned  phrase grounding model $g$, heatmaps are generated according to the extracted captions. 

Similarly to the phrase grounding task, the WWbL task is evaluated using the same two metrics: pointing game accuracy and bounding box accuracy). For each ground-truth pair of bounding box and caption, the closest caption in CLIP space is selected from the list of automatically generated captions. The associated output map of the phrase grounding method is then compared to the ground truth bounding box using the pointing accuracy metric. In addition, bounding boxes are extracted for the output heatmaps $M$, as described above.

For single object discovery we use the Correct Localization (CorLoc) metric as used by ~\cite{deselaers2010localizing, vo2020toward,vo2021large,vo2019unsupervised, wei2019unsupervised,cho2015unsupervised,siva2013looking}. 
A predicted bounding box is considered as correct if the IOU score between the predicted bounding box and one of the ground truth bounding boxes is above 0.5. We evaluate our model on the same datasets as \cite{wang2022tokencut, simeoni2021lost, bielski2022move}.

\smallskip
\noindent{\bf Results\quad}
Tab.~\ref{tab:grounding_pointing} lists the results for Flickr30k, ReferIt, and VG for the weakly-supervised phrase grounding task. Evidently, our method is superior to all baselines, whether training takes place over VG or MS-COCO.
In addition to the pointing game results, Tab.~\ref{tab:grounding_res} presents bounding box accuracy for the phrase grounding task (this data is not available for most baselines). Here, too, our method outperforms the baseline methods by a wide margin.

Phrase grounding samples are provided in Fig.~\ref{fig:compare_hm}, comparing the results after the refinement process (those with $g^h$) to the results of the baseline $g$. As can be seen, our method encourages the localization maps to match the typical shape of image objects. As a result, the predicted bounding box after refining the model is often closer to the actual objects in the image.

The WWbL results are listed in Tab.~\ref{tab:WWbL_res}, which depicts the performance obtained by our $g^{h}$, WWbL~\cite{shaharabany22}, and a baseline that employs the phrase grounding method MG~\cite{akbari2019multi} as part of the WWbL captioning procedure described above. Out of the three models, our refined model $g^{h}$ achieves the best scores, for all benchmarks and both metrics.

{\color{black}Tab.~\ref{tab:od_tokencut} summarize the results on the VOC07, VOC12, and MS-COCO20K datasets for the single object discovery task. When utilizing the MOVE~\cite{bielski2022move} model, our method achieves superior performance compared to all other models across all datasets. This superiority holds true when comparing all methods without CAD and when comparing all methods with CAD. Furthermore, our method consistently outperforms other approaches when refining the DINO model f using both TokenCut~\cite{wang2022tokencut} boxes and LOST~\cite{simeoni2021lost} boxes on all datasets.
}

Fig.~\ref{fig:object_discovery_f_fh_cad} depicts typical samples of our results for the unsupervised object discovery task, when combining our method with either LOST~\cite{simeoni2021lost} or TokenCut~\cite{wang2022tokencut}. Evidently, our refining process improves object and background separation and produces a denser output mask,  which covers the object more completely. Furthermore, the extracted bounding boxes become more accurate.

\smallskip
\noindent\textbf{Ablation study} In order to validate the individual components of our approach, we conducted an ablation study. 

For the phrase grounding task, this study is reported in Tab.~\ref{tab:ablation_pg}. 
The first ablation replaces the loss $L_{h_{BU}}$ with the loss $L_h$, i.e., no union of the detection boxes is performed.  The second ablation employs only the loss of $h$, $L_{h_{BU}}$, and disregards the loss terms that were used to train network $g$. The third ablation employs a single detection box ($k=1$) instead of the default of $k=10$. As can be seen, these three variants reduce performance across all metrics and datasets. The exact reduction in performance varies across the datasets. 

{\color{black}To extensively explore the task of unsupervised object discovery, we conducted a comprehensive ablation study by varying multiple values of k, see Tab.} \ref{tab:ablation_object_discovery}. This ablation was performed using LOST, which is quicker than TokenCut and without the extra overhead of training CAD. Evidently, removing the regularization term, leaving only the loss $L_h$ (there is no box union in this task, since both LOST and TokenCut return a single box) hurts performance. However, as can be expected, using $k=1$, instead of the value of $k=10$ that is used throughout our experiments, better fits this scenario and leads to better performance on VOC07 (and virtually the same on MSCOCO20K).

\smallskip
\noindent{\bf Training time\quad} The time it takes to train our method on medium-sized datasets is reported in Tab.~\ref{tab:runtime}. For both original networks, $f$ and $g$, we use pretrained networks and report the published values. Training $h,f^h,g^h$ reflects our runs on GeForce RTX 2080Ti GPUs ($f$ which is DINO, was trained on much more involved hardware, while $g$ was trained on similar hardware).  As can be seen, training $h$ and refining $f$ or $g$ to obtain $f^h$ or $g^h$ is much quicker than the training of the $f$ and $g$ baselines. The difference in training time between LOST and TokenCut stems from the inference done during training, which is much quicker for LOST.

\begin{table}
    \centering
    \begin{center}
    \begin{tabular}{@{}lccc@{}}
    \toprule
     Network  & Phrase Grounding & \multicolumn{2}{c}{Object discovery}\\
     & & LOST & TokenCut\\
     \midrule
    $f$ or $g$ & 28 x [4] & 72.6 x [16] & 72.6 x [16] \\
    $h$ & 0.5 x [1]  & 0.5 x [1] & 2.5 x [1] \\
    $f^h$ or $g^h$ & 3.2 x [4] & 5.3 x [1] & 20.5 x [1] \\
     \bottomrule
    \end{tabular}
    \end{center}
   \caption{ Training time (hours) for phrase grounding and unsupervised object discovery. Within brackets is the number of GPUS used during training.}
    \label{tab:runtime}
    \end{table}

\section{Conclusions}

We present a novel method, in which a primary network is used in a symbiotic manner with a detection network. The first network is used to extract a feature map and detection boxes, which are used as the input and output of the second. The second network is then used to allow the first network to be refined using the boxes extracted from its output. All training phases are performed on the same training set, within the bounds of the allowed level of supervision. Tested on a wide variety of tasks and benchmarks, the proposed method consistently improves localization accuracy.

\newpage
{\small
\bibliographystyle{ieee_fullname}
\bibliography{main}

\begin{thebibliography}{10}\itemsep=-1pt

\bibitem{akbari2019multi}
Hassan Akbari, Svebor Karaman, Surabhi Bhargava, Brian Chen, Carl Vondrick, and
  Shih-Fu Chang.
\newblock Multi-level multimodal common semantic space for image-phrase
  grounding.
\newblock In {\em Proceedings of the IEEE/CVF Conference on Computer Vision and
  Pattern Recognition}, pages 12476--12486, 2019.

\bibitem{arbelle2021detector}
Assaf Arbelle, Sivan Doveh, Amit Alfassy, Joseph Shtok, Guy Lev, Eli Schwartz,
  Hilde Kuehne, Hila~Barak Levi, Prasanna Sattigeri, Rameswar Panda, et~al.
\newblock Detector-free weakly supervised grounding by separation.
\newblock In {\em Proceedings of the IEEE/CVF International Conference on
  Computer Vision}, pages 1801--1812, 2021.

\bibitem{bai2022point}
Yutong Bai, Xinlei Chen, Alexander Kirillov, Alan Yuille, and Alexander~C Berg.
\newblock Point-level region contrast for object detection pre-training.
\newblock In {\em Proceedings of the IEEE/CVF Conference on Computer Vision and
  Pattern Recognition}, pages 16061--16070, 2022.

\bibitem{bielski2019emergence}
Adam Bielski and Paolo Favaro.
\newblock Emergence of object segmentation in perturbed generative models.
\newblock {\em Advances in Neural Information Processing Systems}, 32, 2019.

\bibitem{bielski2022move}
Adam Bielski and Paolo Favaro.
\newblock Move: Unsupervised movable object segmentation and detection.
\newblock {\em arXiv preprint arXiv:2210.07920}, 2022.

\bibitem{blondel2008fast}
Vincent~D Blondel, Jean-Loup Guillaume, Renaud Lambiotte, and Etienne Lefebvre.
\newblock Fast unfolding of communities in large networks.
\newblock {\em Journal of statistical mechanics: theory and experiment},
  2008(10):P10008, 2008.

\bibitem{bolelli2019connectedcomponents}
Federico Bolelli, Stefano Allegretti, Lorenzo Baraldi, and Costantino Grana.
\newblock Spaghetti labeling: Directed acyclic graphs for block-based connected
  components labeling.
\newblock {\em IEEE Transactions on Image Processing}, PP:1--1, 10 2019.

\bibitem{carion20}
Nicolas Carion, Francisco Massa, Gabriel Synnaeve, Nicolas Usunier, Alexander
  Kirillov, and Sergey Zagoruyko.
\newblock End-to-end object detection with transformers.
\newblock In {\em Computer Vision – ECCV 2020: 16th European Conference,
  Glasgow, UK, August 23–28, 2020, Proceedings, Part I}, page 213–229,
  2020.

\bibitem{caron2021emerging}
Mathilde Caron, Hugo Touvron, Ishan Misra, Herv{\'e} J{\'e}gou, Julien Mairal,
  Piotr Bojanowski, and Armand Joulin.
\newblock Emerging properties in self-supervised vision transformers.
\newblock In {\em ICCV}, 2021.

\bibitem{Chefer_2021_ICCV}
Hila Chefer, Shir Gur, and Lior Wolf.
\newblock Generic attention-model explainability for interpreting bi-modal and
  encoder-decoder transformers.
\newblock In {\em Proceedings of the IEEE/CVF International Conference on
  Computer Vision (ICCV)}, pages 397--406, October 2021.

\bibitem{chefer2021transformer}
Hila Chefer, Shir Gur, and Lior Wolf.
\newblock Transformer interpretability beyond attention visualization.
\newblock In {\em Proceedings of the IEEE/CVF Conference on Computer Vision and
  Pattern Recognition}, pages 782--791, 2021.

\bibitem{chen2017query}
Kan Chen, Rama Kovvuri, and Ram Nevatia.
\newblock Query-guided regression network with context policy for phrase
  grounding.
\newblock In {\em ICCV}, 2017.

\bibitem{chen2021semi}
Xiaokang Chen, Yuhui Yuan, Gang Zeng, and Jingdong Wang.
\newblock Semi-supervised semantic segmentation with cross pseudo supervision.
\newblock In {\em Proceedings of the IEEE/CVF Conference on Computer Vision and
  Pattern Recognition}, pages 2613--2622, 2021.

\bibitem{chiu2016named}
Jason~PC Chiu and Eric Nichols.
\newblock Named entity recognition with bidirectional lstm-cnns.
\newblock {\em Transactions of the association for computational linguistics},
  4:357--370, 2016.

\bibitem{cho2015unsupervised}
Minsu Cho, Suha Kwak, Cordelia Schmid, and Jean Ponce.
\newblock Unsupervised object discovery and localization in the wild:
  Part-based matching with bottom-up region proposals, 2015.

\bibitem{choe2021region}
Junsuk Choe, Dongyoon Han, Sangdoo Yun, Jung-Woo Ha, Seong~Joon Oh, and
  Hyunjung Shim.
\newblock Region-based dropout with attention prior for weakly supervised
  object localization.
\newblock {\em Pattern Recognition}, 116:107949, 2021.

\bibitem{choe2020evaluating}
Junsuk Choe, Seong~Joon Oh, Seungho Lee, Sanghyuk Chun, Zeynep Akata, and
  Hyunjung Shim.
\newblock Evaluating weakly supervised object localization methods right.
\newblock In {\em Proceedings of the IEEE/CVF Conference on Computer Vision and
  Pattern Recognition}, pages 3133--3142, 2020.

\bibitem{imagenet}
J. Deng, W. Dong, R. Socher, L.-J. Li, K. Li, and L. Fei-Fei.
\newblock {ImageNet: A Large-Scale Hierarchical Image Database}.
\newblock In {\em Computer Vision and Pattern Recognition (CVPR)}, 2009.

\bibitem{deselaers2010localizing}
Thomas Deselaers, Bogdan Alexe, and Vittorio Ferrari.
\newblock Localizing objects while learning their appearance.
\newblock In Kostas Daniilidis, Petros Maragos, and Nikos Paragios, editors,
  {\em Computer Vision -- ECCV 2010}, pages 452--466, Berlin, Heidelberg, 2010.
  Springer Berlin Heidelberg.

\bibitem{pascalvoc2007}
M. Everingham, L. Van~Gool, C.~K.~I. Williams, J. Winn, and A. Zisserman.
\newblock The {PASCAL} {V}isual {O}bject {C}lasses {C}hallenge 2007 {R}esults.
\newblock pascal-network.org/challenges/VOC/voc2007.

\bibitem{pascalvoc2012}
M. Everingham, L. Van~Gool, C.~K.~I. Williams, J. Winn, and A. Zisserman.
\newblock The {PASCAL} {V}isual {O}bject {C}lasses {C}hallenge 2012 {(VOC2012)}
  {R}esults.
\newblock http://www.pascal-network.org/challenges/VOC/voc2012.

\bibitem{fang2015captions}
Hao Fang, Saurabh Gupta, Forrest Iandola, Rupesh~K Srivastava, Li Deng, Piotr
  Doll{\'a}r, Jianfeng Gao, Xiaodong He, Margaret Mitchell, John~C Platt,
  et~al.
\newblock From captions to visual concepts and back.
\newblock In {\em Proceedings of the IEEE conference on computer vision and
  pattern recognition}, pages 1473--1482, 2015.

\bibitem{goodfellow2020generative}
Ian Goodfellow, Jean Pouget-Abadie, Mehdi Mirza, Bing Xu, David Warde-Farley,
  Sherjil Ozair, Aaron Courville, and Yoshua Bengio.
\newblock Generative adversarial networks.
\newblock {\em Communications of the ACM}, 63(11):139--144, 2020.

\bibitem{grubinger2006iapr}
Michael Grubinger, Paul Clough, Henning M{\"u}ller, and Thomas Deselaers.
\newblock The iapr tc-12 benchmark: A new evaluation resource for visual
  information systems.
\newblock In {\em International workshop ontoImage}, volume~2, 2006.

\bibitem{resnetcvpr2016}
Kaiming He, Xiangyu Zhang, Shaoqing Ren, and Jian Sun.
\newblock Deep residual learning for image recognition.
\newblock In {\em CVPR}, pages 770--778, 2016.

\bibitem{javed2018learning}
Syed~Ashar Javed, Shreyas Saxena, and Vineet Gandhi.
\newblock Learning unsupervised visual grounding through semantic
  self-supervision.
\newblock {\em arXiv preprint arXiv:1803.06506}, 2018.

\bibitem{kamath2021mdetr}
Aishwarya Kamath, Mannat Singh, Yann LeCun, Gabriel Synnaeve, Ishan Misra, and
  Nicolas Carion.
\newblock Mdetr-modulated detection for end-to-end multi-modal understanding.
\newblock In {\em Proceedings of the IEEE/CVF International Conference on
  Computer Vision}, pages 1780--1790, 2021.

\bibitem{kim2009unsupervised}
Gunhee Kim and Antonio Torralba.
\newblock Unsupervised detection of regions of interest using iterative link
  analysis.
\newblock {\em Advances in neural information processing systems}, 22, 2009.

\bibitem{krishna2017visual}
Ranjay Krishna, Yuke Zhu, Oliver Groth, Justin Johnson, Kenji Hata, Joshua
  Kravitz, Stephanie Chen, Yannis Kalantidis, Li-Jia Li, David~A Shamma, et~al.
\newblock Visual genome: Connecting language and vision using crowdsourced
  dense image annotations.
\newblock {\em International journal of computer vision}, 123(1):32--73, 2017.

\bibitem{Kuhn55}
Harold~W. Kuhn.
\newblock {The Hungarian Method for the Assignment Problem}.
\newblock {\em Naval Research Logistics Quarterly}, 2(1--2):83--97, 1955.

\bibitem{li2021grounded}
Liunian~Harold Li, Pengchuan Zhang, Haotian Zhang, Jianwei Yang, Chunyuan Li,
  Yiwu Zhong, Lijuan Wang, Lu Yuan, Lei Zhang, Jenq-Neng Hwang, et~al.
\newblock Grounded language-image pre-training.
\newblock {\em arXiv preprint arXiv:2112.03857}, 2021.

\bibitem{cocoeccv2014}
Tsung-Yi Lin, Michael Maire, Serge Belongie, James Hays, Pietro Perona, Deva
  Ramanan, Piotr Doll{\'a}r, and C~Lawrence Zitnick.
\newblock Microsoft {COCO}: Common objects in context.
\newblock In {\em ECCV}, volume 8693 of {\em LNCS}, pages 740--755, 2014.

\bibitem{liu2019generative}
Lanlan Liu, Michael Muelly, Jia Deng, Tomas Pfister, and Li-Jia Li.
\newblock Generative modeling for small-data object detection.
\newblock In {\em Proceedings of the IEEE/CVF International Conference on
  Computer Vision}, pages 6073--6081, 2019.

\bibitem{plummer2015flickr30k}
Bryan~A Plummer, Liwei Wang, Chris~M Cervantes, Juan~C Caicedo, Julia
  Hockenmaier, and Svetlana Lazebnik.
\newblock Flickr30k entities: Collecting region-to-phrase correspondences for
  richer image-to-sentence models.
\newblock In {\em Proceedings of the IEEE international conference on computer
  vision}, pages 2641--2649, 2015.

\bibitem{polyak2018unsupervised}
Adam Polyak, Yaniv Taigman, and Lior Wolf.
\newblock Unsupervised generation of free-form and parameterized avatars.
\newblock {\em IEEE transactions on pattern analysis and machine intelligence},
  42(2):444--459, 2018.

\bibitem{qin19}
Zhenyue Qin, Dongwoo Kim, and Tom Gedeon.
\newblock Rethinking softmax with cross-entropy: Neural network classifier as
  mutual information estimator.
\newblock {\em arXiv preprint arXiv:1911.10688}, 2019.

\bibitem{radford21}
Alec Radford, Jong~Wook Kim, Chris Hallacy, Aditya Ramesh, Gabriel Goh,
  Sandhini Agarwal, Girish Sastry, Amanda Askell, Pamela Mishkin, Jack Clark,
  et~al.
\newblock Learning transferable visual models from natural language
  supervision.
\newblock {\em arXiv preprint arXiv:2103.00020}, 2021.

\bibitem{radford2019language}
Alec Radford, Jeffrey Wu, Rewon Child, David Luan, Dario Amodei, Ilya
  Sutskever, et~al.
\newblock Language models are unsupervised multitask learners.
\newblock {\em OpenAI blog}, 1(8):9, 2019.

\bibitem{renNIPS15fasterrcnn}
Shaoqing Ren, Kaiming He, Ross Girshick, and Jian Sun.
\newblock Faster {R-CNN}: Towards real-time object detection with region
  proposal networks.
\newblock In {\em Advances in Neural Information Processing Systems ({NIPS})},
  2015.

\bibitem{Rezatofighi19}
Hamid Rezatofighi, Nathan Tsoi, JunYoung Gwak, Amir Sadeghian, Ian Reid, and
  Silvio Savarese.
\newblock Generalized intersection over union.
\newblock In {\em The IEEE Conference on Computer Vision and Pattern
  Recognition (CVPR)}, June 2019.

\bibitem{sarzynska2021detecting}
Justyna Sarzynska-Wawer, Aleksander Wawer, Aleksandra Pawlak, Julia
  Szymanowska, Izabela Stefaniak, Michal Jarkiewicz, and Lukasz Okruszek.
\newblock Detecting formal thought disorder by deep contextualized word
  representations.
\newblock {\em Psychiatry Research}, 304:114135, 2021.

\bibitem{shaharabany2022looking}
Tal Shaharabany, Yoad Tewel, and Lior Wolf.
\newblock What is where by looking: Weakly-supervised open-world
  phrase-grounding without text inputs.
\newblock {\em arXiv preprint arXiv:2206.09358}, 2022.

\bibitem{shaharabany22}
Tal Shaharabany, Yoad Tewel, and Lior Wolf.
\newblock What is where by looking: Weakly-supervised open-world
  phrase-grounding without text inputs.
\newblock In Alice~H. Oh, Alekh Agarwal, Danielle Belgrave, and Kyunghyun Cho,
  editors, {\em Advances in Neural Information Processing Systems}, 2022.

\bibitem{shi2000normalized}
Jianbo Shi and J. Malik.
\newblock Normalized cuts and image segmentation.
\newblock {\em IEEE Transactions on Pattern Analysis and Machine Intelligence},
  22(8):888--905, 2000.

\bibitem{shin2022selfmask}
Gyungin Shin, Samuel Albanie, and Weidi Wie.
\newblock Unsupervised salient object detection with spectral cluster voting.
\newblock {\em arXiv preprint arXiv:2203.12614}, 2022.

\bibitem{shin2022namedmask}
Gyungin Shin, Weidi Xie, and Samuel Albanie.
\newblock Namedmask: Distilling segmenters from complementary foundation
  models.
\newblock In {\em CVPRW}, 2023.

\bibitem{simeoni2021lost}
Oriane Sim\'eoni, Gilles Puy, Huy~V. Vo, Simon Roburin, Spyros Gidaris, Andrei
  Bursuc, Patrick P\'erez, Renaud Marlet, and Jean Ponce.
\newblock Localizing objects with self-supervised transformers and no labels.
\newblock In {\em Proceedings of the British Machine Vision Conference (BMVC)},
  November 2021.

\bibitem{simonyan2014very}
Karen Simonyan and Andrew Zisserman.
\newblock Very deep convolutional networks for large-scale image recognition.
\newblock {\em arXiv preprint arXiv:1409.1556}, 2014.

\bibitem{siméoni2023unsupervised}
Oriane Siméoni, Chloé Sekkat, Gilles Puy, Antonin Vobecky, Éloi Zablocki,
  and Patrick Pérez.
\newblock Unsupervised object localization: Observing the background to
  discover objects, 2023.

\bibitem{siva2013looking}
Parthipan Siva, Chris Russell, Tao Xiang, and Lourdes Agapito.
\newblock Looking beyond the image: Unsupervised learning for object saliency
  and detection.
\newblock In {\em 2013 IEEE Conference on Computer Vision and Pattern
  Recognition}, pages 3238--3245, 2013.

\bibitem{suzuki1985topological}
Satoshi Suzuki et~al.
\newblock Topological structural analysis of digitized binary images by border
  following.
\newblock {\em Computer vision, graphics, and image processing}, 30(1):32--46,
  1985.

\bibitem{Uijlings13}
J.R.R. Uijlings, K.E.A. van~de Sande, T. Gevers, and A.W.M. Smeulders.
\newblock Selective search for object recognition.
\newblock {\em International Journal of Computer Vision}, 2013.

\bibitem{vo2019unsupervised}
Huy~V. Vo, Francis Bach, Minsu Cho, Kai Han, Yann LeCun, Patrick Perez, and
  Jean Ponce.
\newblock Unsupervised image matching and object discovery as optimization,
  2019.

\bibitem{vo2020toward}
Huy~V. Vo, Patrick Pérez, and Jean Ponce.
\newblock Toward unsupervised, multi-object discovery in large-scale image
  collections, 2020.

\bibitem{vo2021large}
Huy~V. Vo, Elena Sizikova, Cordelia Schmid, Patrick Pérez, and Jean Ponce.
\newblock Large-scale unsupervised object discovery, 2021.

\bibitem{voynov2021object}
Andrey Voynov, Stanislav Morozov, and Artem Babenko.
\newblock Object segmentation without labels with large-scale generative
  models.
\newblock In {\em International Conference on Machine Learning}, pages
  10596--10606. PMLR, 2021.

\bibitem{wang2023cut}
Xudong Wang, Rohit Girdhar, Stella~X Yu, and Ishan Misra.
\newblock Cut and learn for unsupervised object detection and instance
  segmentation.
\newblock In {\em Proceedings of the IEEE/CVF Conference on Computer Vision and
  Pattern Recognition}, pages 3124--3134, 2023.

\bibitem{wang2022tokencut}
Yangtao Wang, Xi Shen, Shell~Xu Hu, Yuan Yuan, James~L. Crowley, and Dominique
  Vaufreydaz.
\newblock Self-supervised transformers for unsupervised object discovery using
  normalized cut.
\newblock In {\em Conference on Computer Vision and Pattern Recognition}, 2022.

\bibitem{wei2019unsupervised}
Xiu-Shen Wei, Chen-Lin Zhang, Jianxin Wu, Chunhua Shen, and Zhi-Hua Zhou.
\newblock Unsupervised object discovery and co-localization by deep descriptor
  transformation.
\newblock {\em Pattern Recognition}, 88:113--126, 2019.

\bibitem{wu2019detectron2}
Yuxin Wu, Alexander Kirillov, Francisco Massa, Wan-Yen Lo, and Ross Girshick.
\newblock Detectron2.
\newblock \url{https://github.com/facebookresearch/detectron2}, 2019.

\bibitem{xiao2017weakly}
Fanyi Xiao, Leonid Sigal, and Yong Jae~Lee.
\newblock Weakly-supervised visual grounding of phrases with linguistic
  structures.
\newblock In {\em Proceedings of the IEEE Conference on Computer Vision and
  Pattern Recognition}, pages 5945--5954, 2017.

\bibitem{zhang2018top}
Jianming Zhang, Sarah~Adel Bargal, Zhe Lin, Jonathan Brandt, Xiaohui Shen, and
  Stan Sclaroff.
\newblock Top-down neural attention by excitation backprop.
\newblock {\em International Journal of Computer Vision}, 126(10):1084--1102,
  2018.

\bibitem{zhang2020object}
Tianshu Zhang, Buzhen Huang, and Yangang Wang.
\newblock Object-occluded human shape and pose estimation from a single color
  image.
\newblock In {\em Proceedings of the IEEE/CVF conference on computer vision and
  pattern recognition}, pages 7376--7385, 2020.

\bibitem{zhang2021refining}
Xiao Zhang, Yixiao Ge, Yu Qiao, and Hongsheng Li.
\newblock Refining pseudo labels with clustering consensus over generations for
  unsupervised object re-identification.
\newblock In {\em Proceedings of the IEEE/CVF Conference on Computer Vision and
  Pattern Recognition}, pages 3436--3445, 2021.

\bibitem{zitnick2014edge}
C~Lawrence Zitnick and Piotr Doll{\'a}r.
\newblock Edge boxes: Locating object proposals from edges.
\newblock In {\em Computer Vision--ECCV 2014}, pages 391--405. Springer, 2014.

\end{thebibliography}
}

\clearpage
\appendix
\section*{Supplementary Material}
This appendix presents visual results that demonstrate the effectiveness of our refined models $g^h$ and $f^h$ in various tasks, including weakly supervised and unsupervised localization, What-is-where-by-looking, and unsupervised single object discovery. By building upon existing models $g$ and $f$, we have showcased improvements in output localization maps and bounding boxes.

Our comprehensive comparisons span multiple datasets, including MS-COCO14~\cite{cocoeccv2014}, Visual-Genome~\cite{krishna2017visual}, Flickr30K~\cite{plummer2015flickr30k}, ReferIt~\cite{chen2017query, grubinger2006iapr}, PASCAL-VOC07~\cite{pascalvoc2007}, PASCAL-VOC12~\cite{pascalvoc2012}, and MS-COCO20K~\cite{cocoeccv2014}. These comparisons serve to highlight the adaptability and robustness of our refined models across different tasks and datasets. The visual results provide strong evidence of our models' superiority in generating more accurate localization maps and bounding boxes compared to their base models.

The code and scripts for reproducing the paper's results are attached to this supplementary.

\section*{Weakly supervised phrase-grounding visual results}

\begin{figure*}[h!]
    \setlength{\tabcolsep}{3pt}
    \renewcommand{\arraystretch}{1}
    \centering
\begin{tabular}{@{}p{.1cm}@{~}c@{}c@{}c@{~}c@{}@{}p{.1cm}@{~}c@{}c@{}c@{~}c@{}}

&\rotatebox{90}{\footnotesize a young woman} & \includegraphics[valign=B,width=0.1404\linewidth]{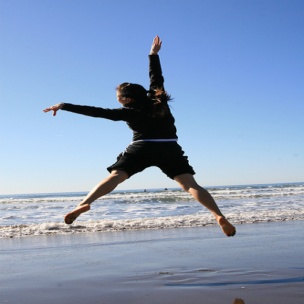} & \includegraphics[valign=B,width=0.1404\linewidth]{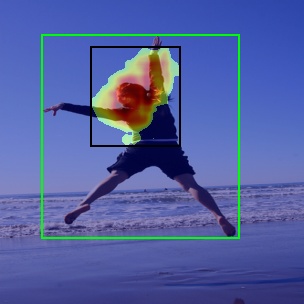} & \includegraphics[valign=B,width=0.1404\linewidth]{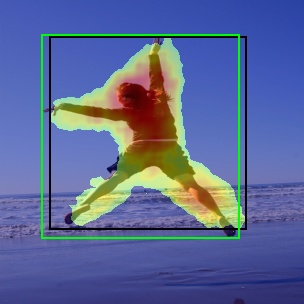} &
&\rotatebox{90}{\footnotesize foreign folk dancers} & \includegraphics[valign=B,width=0.1404\linewidth]{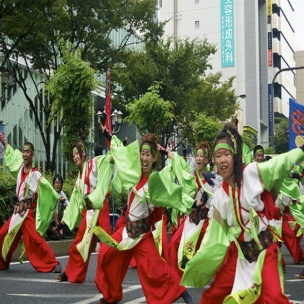} & \includegraphics[valign=B,width=0.1404\linewidth]{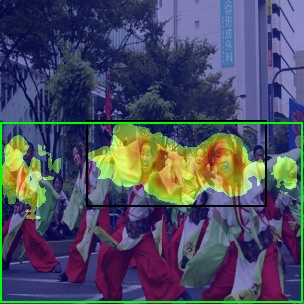} & \includegraphics[valign=B,width=0.1404\linewidth]{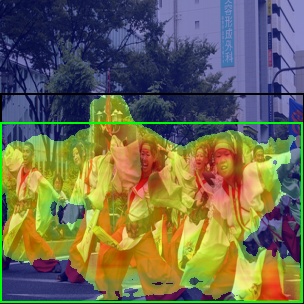} \\
&\rotatebox{90}{\footnotesize two men} & \includegraphics[valign=B,width=0.1404\linewidth]{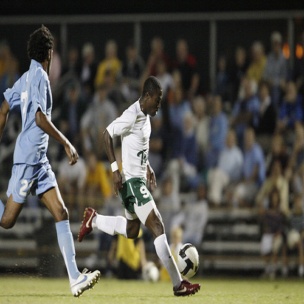} & \includegraphics[valign=B,width=0.1404\linewidth]{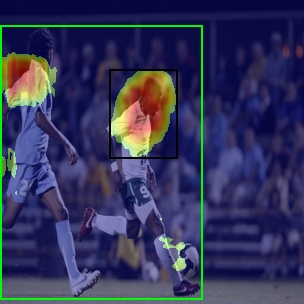} & \includegraphics[valign=B,width=0.1404\linewidth]{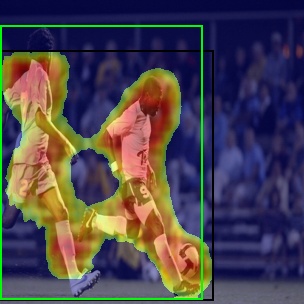} &
&\rotatebox{90}{\footnotesize a lady} & \includegraphics[valign=B,width=0.1404\linewidth]{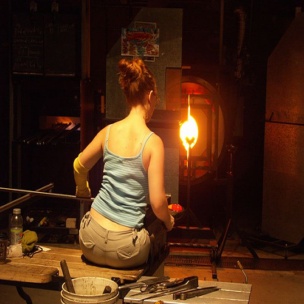} & \includegraphics[valign=B,width=0.1404\linewidth]{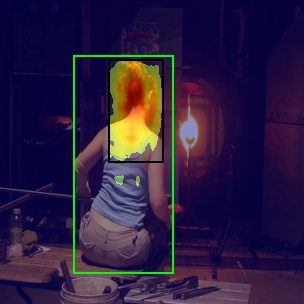} & \includegraphics[valign=B,width=0.1404\linewidth]{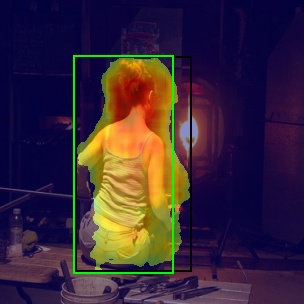} \\
\end{tabular}
    \caption{Phrase-grounding results on Flickr30K\cite{plummer2015flickr30k} dataset. Model $g^h$ was trained on MS-COCO14\cite{cocoeccv2014} dataset. (a) the phrase (b) the input image (c) results (black) for network $g$ \cite{shaharabany22} compared to ground-truth box (green) (d) same for refined network $g^{h}$. (e) same as a (f) same as b (g) same as c (h) same as d}
    \label{fig:appendix_pg_coco_flickr}
\end{figure*}

\begin{figure*}[h!]
    \setlength{\tabcolsep}{3pt} 
    \renewcommand{\arraystretch}{1}
    \centering
    \begin{tabular}{@{}p{.1cm}@{~}c@{}c@{}c@{~}c@{}@{}p{.1cm}@{~}c@{}c@{}c@{~}c@{}}

&\rotatebox{90}{\footnotesize woman} & \includegraphics[valign=B,width=0.1404\linewidth]{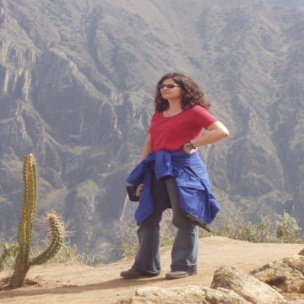} & \includegraphics[valign=B,width=0.1404\linewidth]{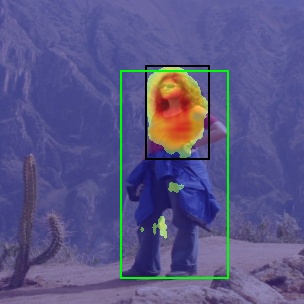} & \includegraphics[valign=B,width=0.1404\linewidth]{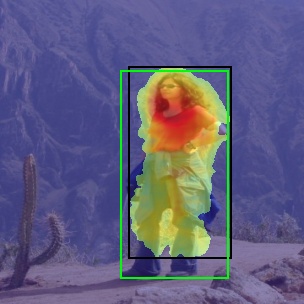} &
&\rotatebox{90}{\footnotesize bridge} & \includegraphics[valign=B,width=0.1404\linewidth]{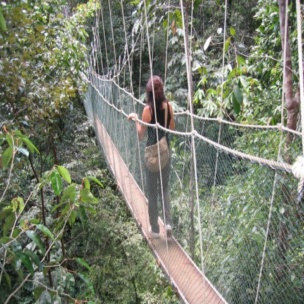} & \includegraphics[valign=B,width=0.1404\linewidth]{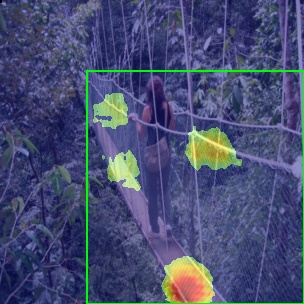} & \includegraphics[valign=B,width=0.1404\linewidth]{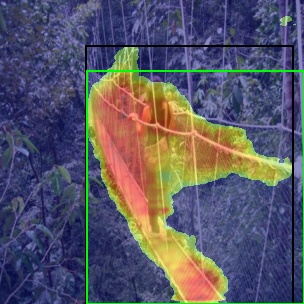} \\
&\rotatebox{90}{\footnotesize train} & \includegraphics[valign=B,width=0.1404\linewidth]{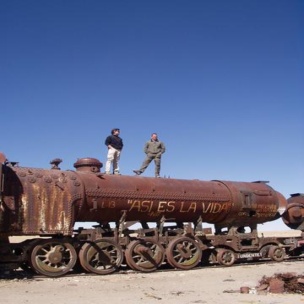} & \includegraphics[valign=B,width=0.1404\linewidth]{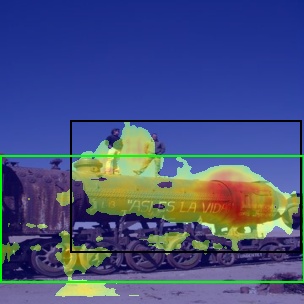} & \includegraphics[valign=B,width=0.1404\linewidth]{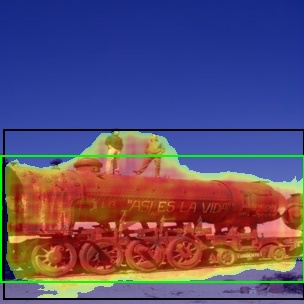} &
&\rotatebox{90}{\footnotesize any of the people} & \includegraphics[valign=B,width=0.1404\linewidth]{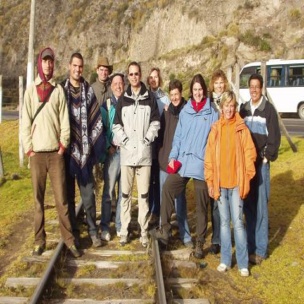} & \includegraphics[valign=B,width=0.1404\linewidth]{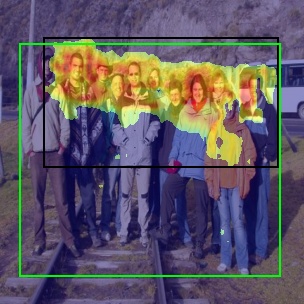} & \includegraphics[valign=B,width=0.1404\linewidth]{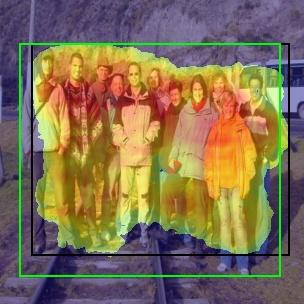} \\
\end{tabular}
    \caption{Phrase-grounding results on ReferIt\cite{chen2017query, grubinger2006iapr} dataset. Model $g^h$ was trained on MS-COCO14\cite{cocoeccv2014} dataset. (a) the phrase (b) the input image (c) results (black) for network $g$ \cite{shaharabany22} compared to ground-truth box (green) (d) same for refined network $g^{h}$. (e) same as a (f) same as b (g) same as c (h) same as d}
    \label{fig:appendix_pg_coco_referit}
\end{figure*}

\begin{figure*}[h!]
    \setlength{\tabcolsep}{3pt}
    \renewcommand{\arraystretch}{1}
    \centering
    \begin{tabular}{@{}p{.1cm}@{~}c@{}c@{}c@{~}c@{}@{}p{.1cm}@{~}c@{}c@{}c@{~}c@{}}

&\rotatebox{90}{\footnotesize ... umbrella} & \includegraphics[valign=B,width=0.1404\linewidth]{vis/appendix/pg/coco/vg/1047_m1_image_of_green_and_white_umbrella-img.jpg} & \includegraphics[valign=B,width=0.1404\linewidth]{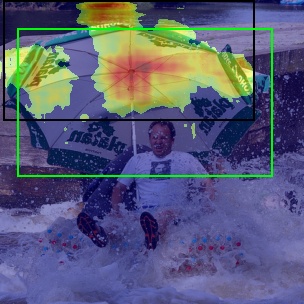} & \includegraphics[valign=B,width=0.1404\linewidth]{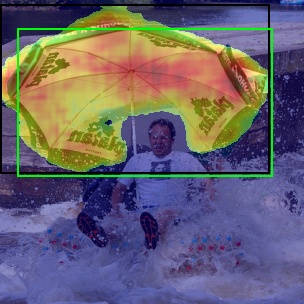} &
&\rotatebox{90}{\footnotesize a small bird ... branch} & \includegraphics[valign=B,width=0.1404\linewidth]{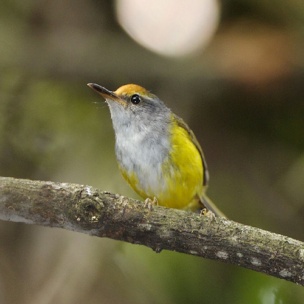} & \includegraphics[valign=B,width=0.1404\linewidth]{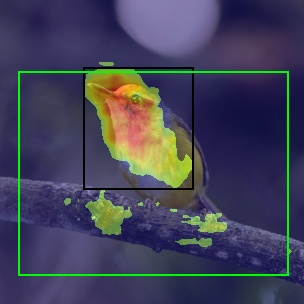} & \includegraphics[valign=B,width=0.1404\linewidth]{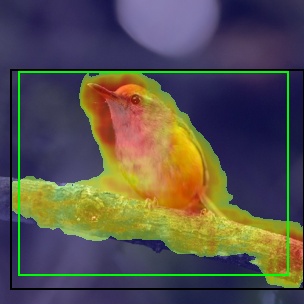} \\
&\rotatebox{90}{\footnotesize ... man} & \includegraphics[valign=B,width=0.1404\linewidth]{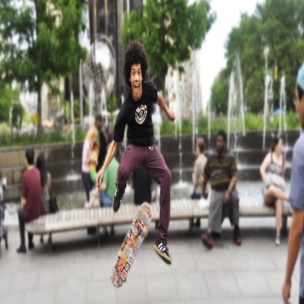} & \includegraphics[valign=B,width=0.1404\linewidth]{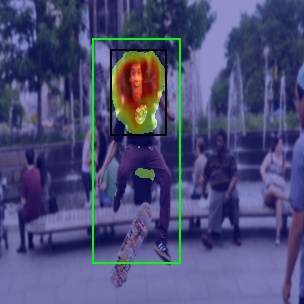} & \includegraphics[valign=B,width=0.1404\linewidth]{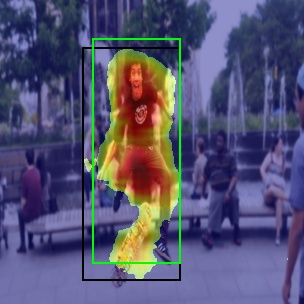} &
&\rotatebox{90}{\footnotesize boy ... shirt} & \includegraphics[valign=B,width=0.1404\linewidth]{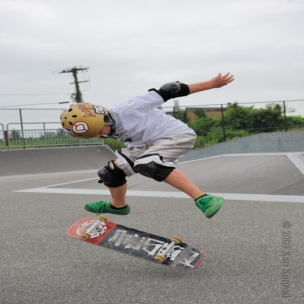} & \includegraphics[valign=B,width=0.1404\linewidth]{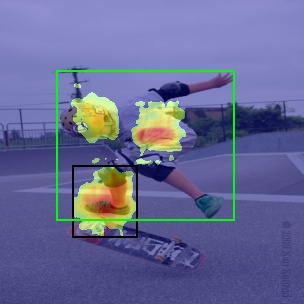} & \includegraphics[valign=B,width=0.1404\linewidth]{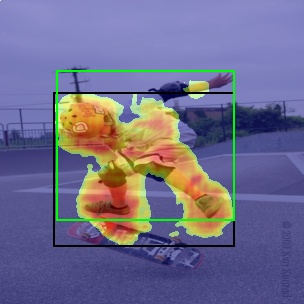} \\
\end{tabular}
    \caption{Phrase-grounding results on Visual Genome~\cite{krishna2017visual} dataset. Model $g^h$ was trained on MS-COCO14\cite{cocoeccv2014} dataset. (a) the phrase (b) the input image (c) results (black) for network $g$ \cite{shaharabany22} compared to ground-truth box (green) (d) same for refined network $g^{h}$. (e) same as a (f) same as b (g) same as c (h) same as d}
    \label{fig:appendix_pg_coco_vg}
\end{figure*}

\begin{figure*}[h!]
    \setlength{\tabcolsep}{3pt} 
    \renewcommand{\arraystretch}{1}
    \centering
    \begin{tabular}{@{}p{.1cm}@{~}c@{}c@{}c@{~}c@{}@{}p{.1cm}@{~}c@{}c@{}c@{~}c@{}}

&\rotatebox{90}{\footnotesize a bucking bull} & \includegraphics[valign=B,width=0.1404\linewidth]{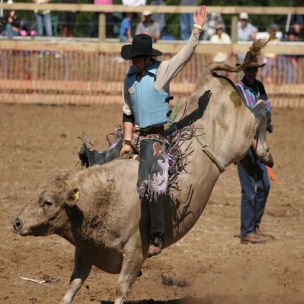} & \includegraphics[valign=B,width=0.1404\linewidth]{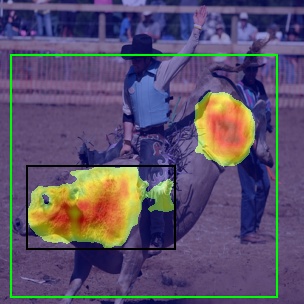} & \includegraphics[valign=B,width=0.1404\linewidth]{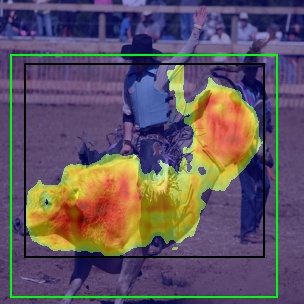} &
&\rotatebox{90}{\footnotesize a segway scooter} & \includegraphics[valign=B,width=0.1404\linewidth]{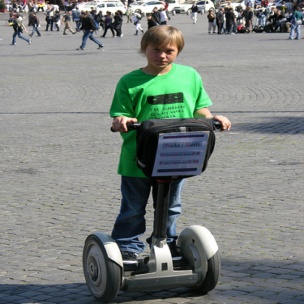} & \includegraphics[valign=B,width=0.1404\linewidth]{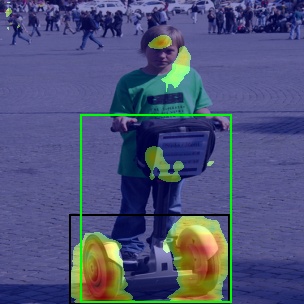} & \includegraphics[valign=B,width=0.1404\linewidth]{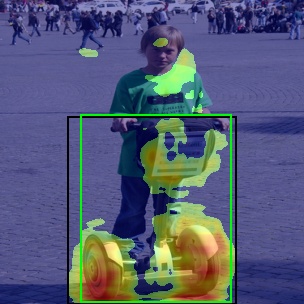} \\
&\rotatebox{90}{\footnotesize football players} & \includegraphics[valign=B,width=0.1404\linewidth]{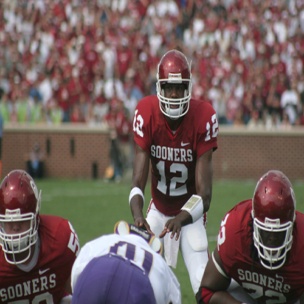} & \includegraphics[valign=B,width=0.1404\linewidth]{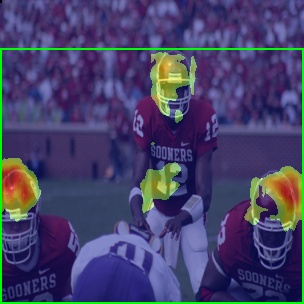} & \includegraphics[valign=B,width=0.1404\linewidth]{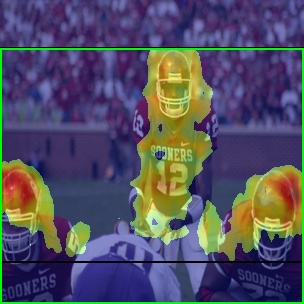} &
&\rotatebox{90}{\footnotesize a young man} & \includegraphics[valign=B,width=0.1404\linewidth]{vis/appendix/pg/vg/flickr/361_m1_image_of_a_young_man-img.jpg} & \includegraphics[valign=B,width=0.1404\linewidth]{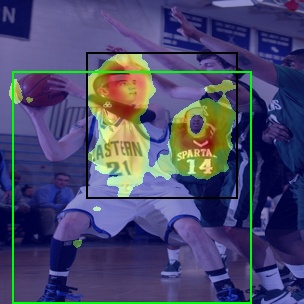} & \includegraphics[valign=B,width=0.1404\linewidth]{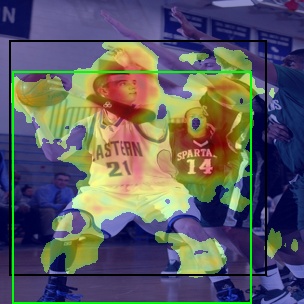} \\
\end{tabular}
    \caption{Phrase-grounding results on Flickr30K\cite{plummer2015flickr30k} dataset. Model $g^h$ was trained on Visual Genome~\cite{krishna2017visual} dataset. (a) the phrase (b) the input image (c) results (black) for network $g$ \cite{shaharabany22} compared to ground-truth box (green) (d) same for refined network $g^{h}$. (e) same as a (f) same as b (g) same as c (h) same as d}
    \label{fig:appendix_pg_vg_flickr}
\end{figure*}

\begin{figure*}[h!]
    \setlength{\tabcolsep}{3pt} 
    \renewcommand{\arraystretch}{1}
    \centering

    \begin{tabular}{@{}p{.1cm}@{~}c@{}c@{}c@{~}c@{}@{}p{.1cm}@{~}c@{}c@{}c@{~}c@{}}

&\rotatebox{90}{\footnotesize couple} & \includegraphics[valign=B,width=0.1404\linewidth]{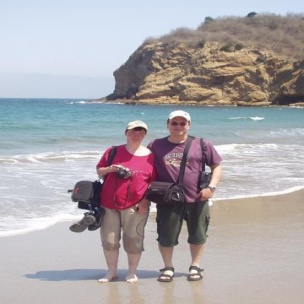} & \includegraphics[valign=B,width=0.1404\linewidth]{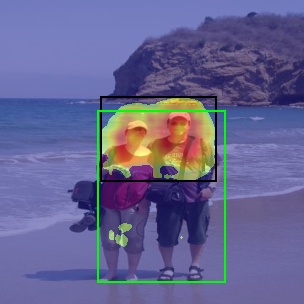} & \includegraphics[valign=B,width=0.1404\linewidth]{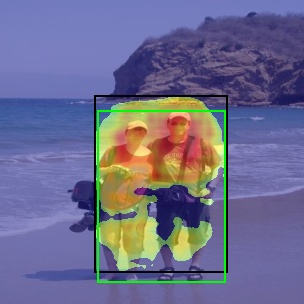} &
&\rotatebox{90}{\footnotesize snow front} & \includegraphics[valign=B,width=0.1404\linewidth]{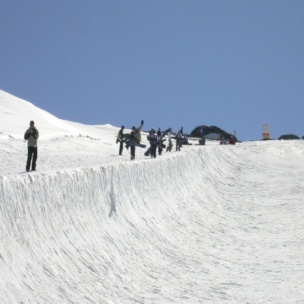} & \includegraphics[valign=B,width=0.1404\linewidth]{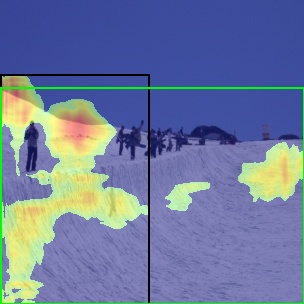} & \includegraphics[valign=B,width=0.1404\linewidth]{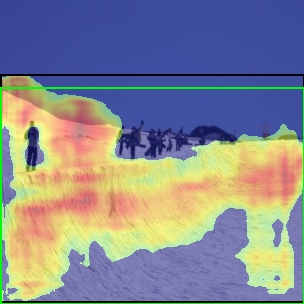} \\
&\rotatebox{90}{\footnotesize guy} & \includegraphics[valign=B,width=0.1404\linewidth]{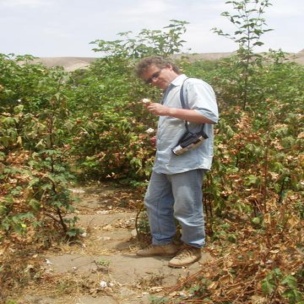} & \includegraphics[valign=B,width=0.1404\linewidth]{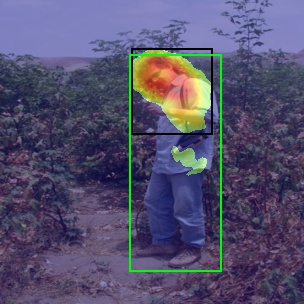} & \includegraphics[valign=B,width=0.1404\linewidth]{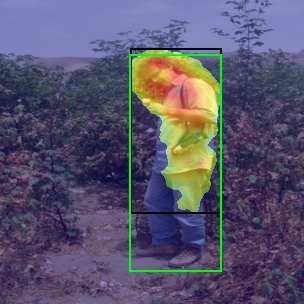} &
&\rotatebox{90}{\footnotesize black object} & \includegraphics[valign=B,width=0.1404\linewidth]{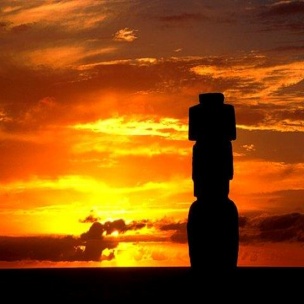} & \includegraphics[valign=B,width=0.1404\linewidth]{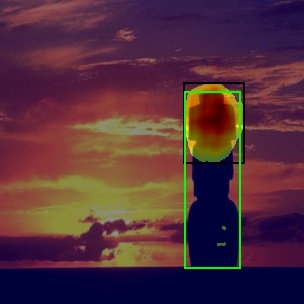} & \includegraphics[valign=B,width=0.1404\linewidth]{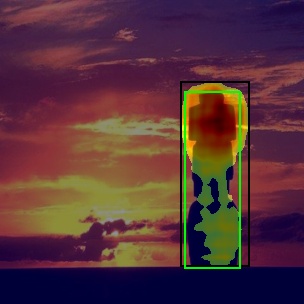} \\
\end{tabular}

    \caption{Phrase-grounding results on ReferIt\cite{chen2017query, grubinger2006iapr} dataset. Model $g^h$ was trained on Visual Genome~\cite{krishna2017visual} dataset. (a) the phrase (b) the input image (c) results (black) for network $g$ \cite{shaharabany22} compared to ground-truth box (green) (d) same for refined network $g^{h}$. (e) same as a (f) same as b (g) same as c (h) same as d}
    \label{fig:appendix_pg_vg_referit}
\end{figure*}

\begin{figure*}[h!]
    \setlength{\tabcolsep}{3pt} 
    \renewcommand{\arraystretch}{1}
    \centering

    \begin{tabular}{@{}p{.1cm}@{~}c@{}c@{}c@{~}c@{}@{}p{.1cm}@{~}c@{}c@{}c@{~}c@{}}
&\rotatebox{90}{\footnotesize biker ...} & \includegraphics[valign=B,width=0.1404\linewidth]{vis/appendix/pg/vg/vg/17013_m1_image_of_biker_sitting_on_motorcycle-img.jpg} & \includegraphics[valign=B,width=0.1404\linewidth]{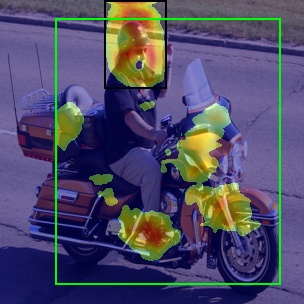} & \includegraphics[valign=B,width=0.1404\linewidth]{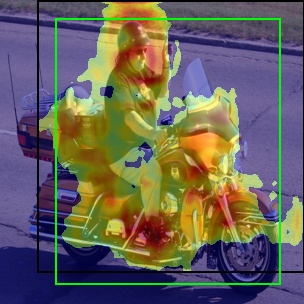} &
&\rotatebox{90}{\footnotesize a sheep barn} & \includegraphics[valign=B,width=0.1404\linewidth]{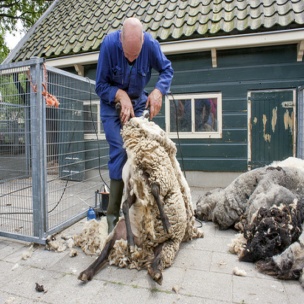} & \includegraphics[valign=B,width=0.1404\linewidth]{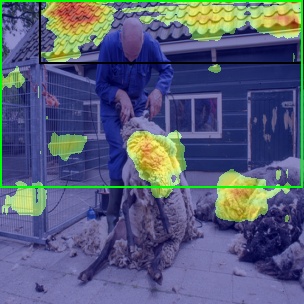} & \includegraphics[valign=B,width=0.1404\linewidth]{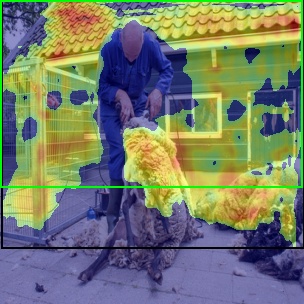} \\
&\rotatebox{90}{\footnotesize woman playing tennis} & \includegraphics[valign=B,width=0.1404\linewidth]{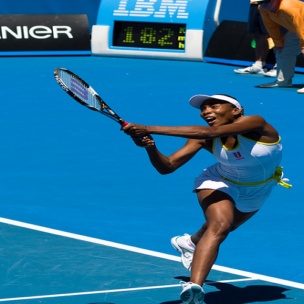} & \includegraphics[valign=B,width=0.1404\linewidth]{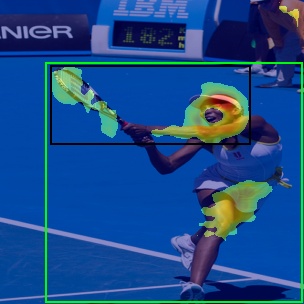} & \includegraphics[valign=B,width=0.1404\linewidth]{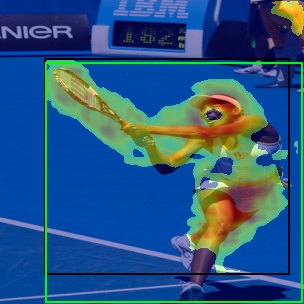} &
&\rotatebox{90}{\footnotesize ... jumping} & \includegraphics[valign=B,width=0.1404\linewidth]{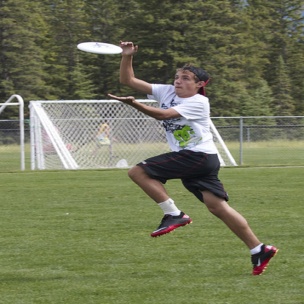} & \includegraphics[valign=B,width=0.1404\linewidth]{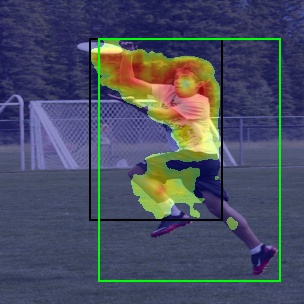} & \includegraphics[valign=B,width=0.1404\linewidth]{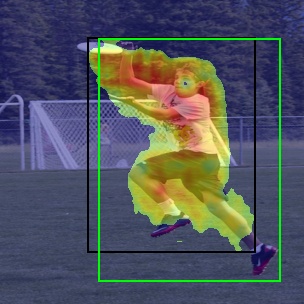} \\
\end{tabular}
    \caption{Phrase-grounding results on Visual Genome~\cite{krishna2017visual} dataset. Model $g^h$ was trained on the same dataset. (a) the phrase (b) the input image (c) results (black) for network $g$ \cite{shaharabany22} compared to ground-truth box (green) (d) same for refined network $g^{h}$. (e) same as a (f) same as b (g) same as c (h) same as d}
    \label{fig:appendix_pg_vg_vg}
\end{figure*}

    We present visual outcomes of our model, $g^h$, which is built upon the previously published model $g$ by \cite{shaharabany22}. We compare the localization maps and bounding box outputs generated by both models and evaluate each bounding box against the ground truth. We showcase the results for models trained on the MS-COCO14~\cite{cocoeccv2014} and Visual-Genome~\cite{krishna2017visual} datasets. For each model, we display visualizations on the Flickr30K\cite{plummer2015flickr30k}, ReferIt \cite{chen2017query, grubinger2006iapr}, and Visual-Genome~\cite{krishna2017visual} datasets. Figures~\ref{fig:appendix_pg_coco_flickr}, \ref{fig:appendix_pg_coco_referit}, \ref{fig:appendix_pg_coco_vg} illustrate the results for the MS-COCO-based model, while the outcomes for the VG-based model can be found in Figures~\ref{fig:appendix_pg_vg_flickr}, \ref{fig:appendix_pg_vg_referit}, \ref{fig:appendix_pg_vg_vg}.

\section*{What is where by looking visual results}
\begin{figure*}
    \begin{tabular}{@{}c@{}c@{}c@{}c@{}c@{~}}

    &\multicolumn{2}{c}{\small{a woman wearing a hat}}& \multicolumn{2}{c}{\small{a woman in a kimono}}\\
    \cmidrule(lr){2-3}
    \cmidrule(lr){4-5}
    \includegraphics[width=0.19324\linewidth]{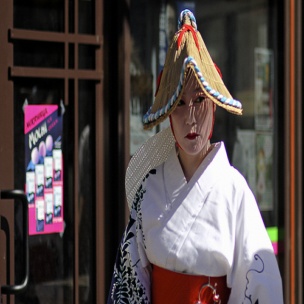} &
    \includegraphics[width=0.19324\linewidth]{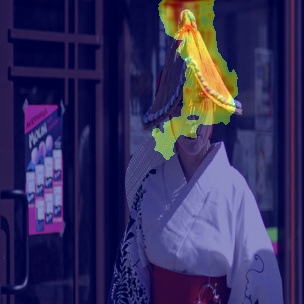} &
    \includegraphics[width=0.19324\linewidth]{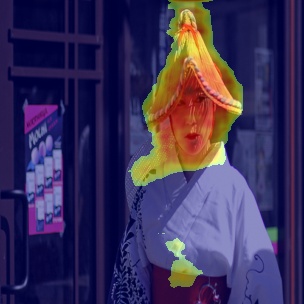} &
        \includegraphics[width=0.19324\linewidth]{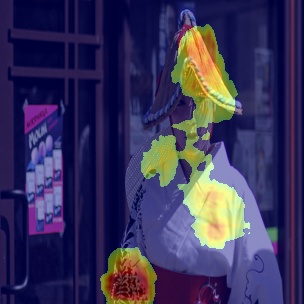} &
    \includegraphics[width=0.19324\linewidth]{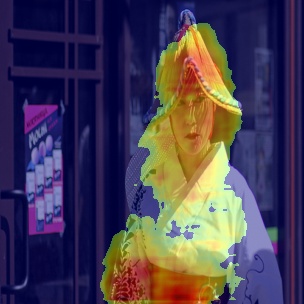}  \\

    &\multicolumn{2}{c}{\small{a bunch of balloons}}& \multicolumn{2}{c}{\small{a group of people walking down the street}}\\
    \cmidrule(lr){2-3}
    \cmidrule(lr){4-5}
    \includegraphics[width=0.19324\linewidth]{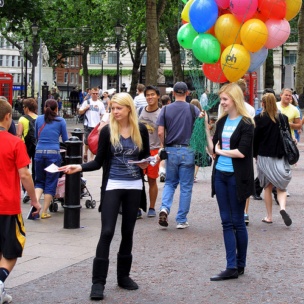} &
    \includegraphics[width=0.19324\linewidth]{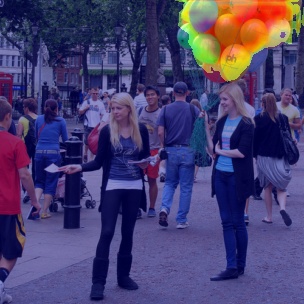} &
    \includegraphics[width=0.19324\linewidth]{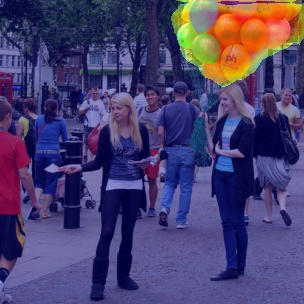} &
    \includegraphics[width=0.19324\linewidth]{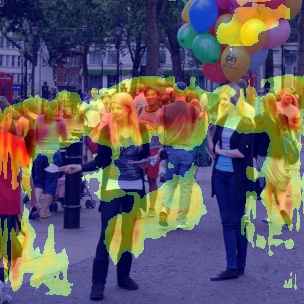} &
    \includegraphics[width=0.19324\linewidth]{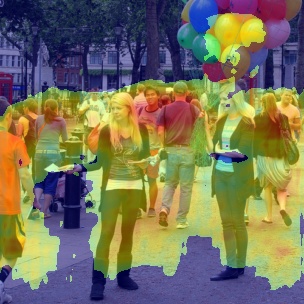}
\\

    &\multicolumn{2}{c}{\small{a police officer}}& \multicolumn{2}{c}{\small{a person wearing a safety vest}}\\
\end{tabular}
    
    \caption{What-is-where-by-looking results on Flickr30K\cite{plummer2015flickr30k} dataset. Model $g^h$ was trained on MS-COCO14\cite{cocoeccv2014} dataset. (a) the input image (b) results for network $g$ \cite{shaharabany22} (c) results for network $g^h$  (d-e) same as b-c, using different phrase}
    \label{fig:appendix_wwbl}
\end{figure*}

    We present visual outcomes for the What-is-where-by-looking task using our improved model $g^h$, which is derived from the previously published model $g$ by \cite{shaharabany22}. We compare the localization maps generated by both models, using the same image but different phrases. In Figure~\ref{fig:appendix_wwbl}, we display the results for the Flickr30K\cite{plummer2015flickr30k} dataset, with models $g$ and $g^h$ trained on the MS-COCO14~\cite{cocoeccv2014} dataset.

\section*{Unsupervised single object discovery visual results}
\begin{figure*}[t!]
    \setlength{\tabcolsep}{3pt}
    \renewcommand{\arraystretch}{1}
     \centering
   \begin{tabular}{cccccc}

\includegraphics[valign=B,width=0.102432\linewidth]{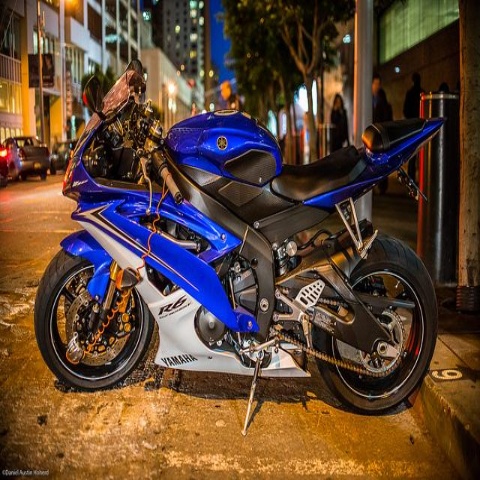} & \includegraphics[valign=B,width=0.102432\linewidth]{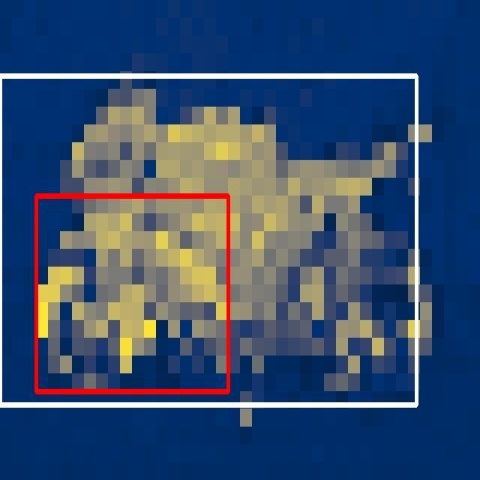} & \includegraphics[valign=B,width=0.102432\linewidth]{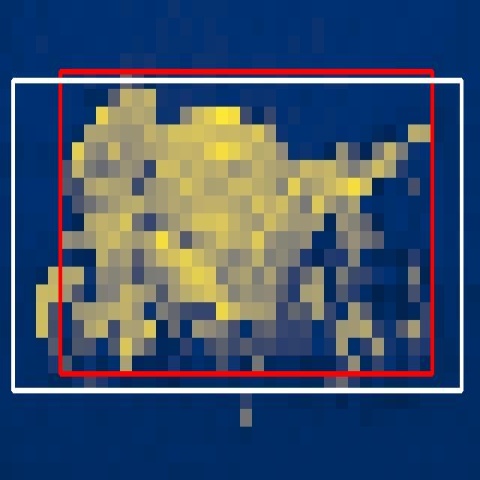} &
\includegraphics[valign=B,width=0.102432\linewidth]{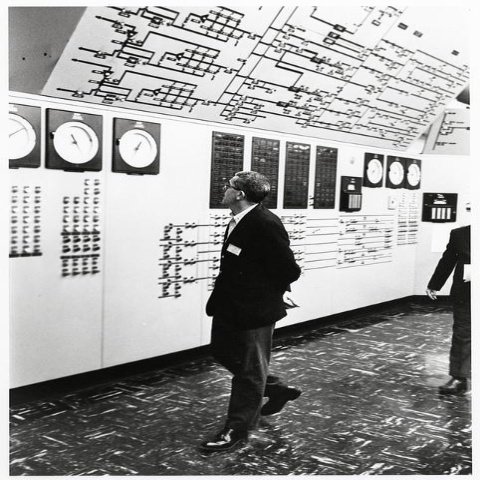} & \includegraphics[valign=B,width=0.102432\linewidth]{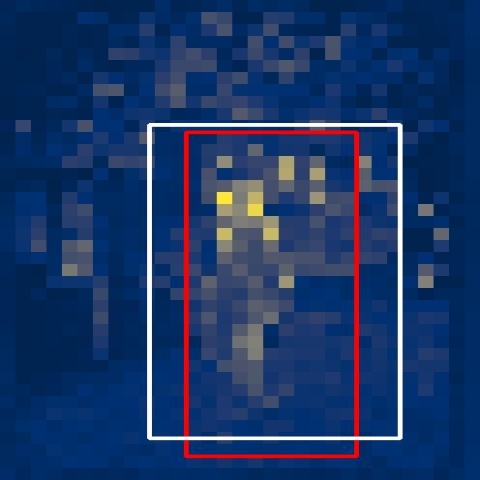} & \includegraphics[valign=B,width=0.102432\linewidth]{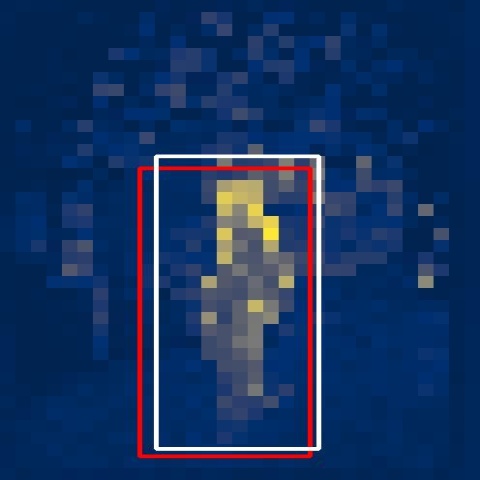} \\
\includegraphics[valign=B,width=0.102432\linewidth]{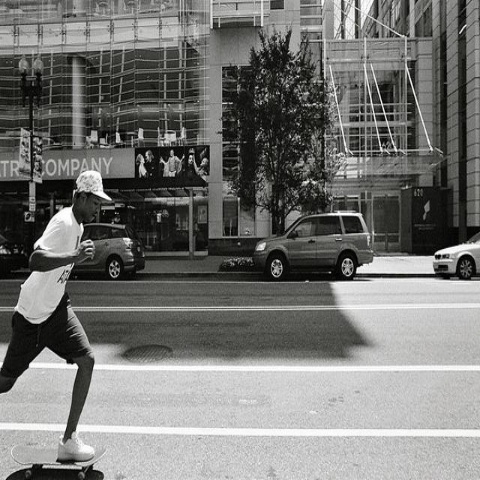} & \includegraphics[valign=B,width=0.102432\linewidth]{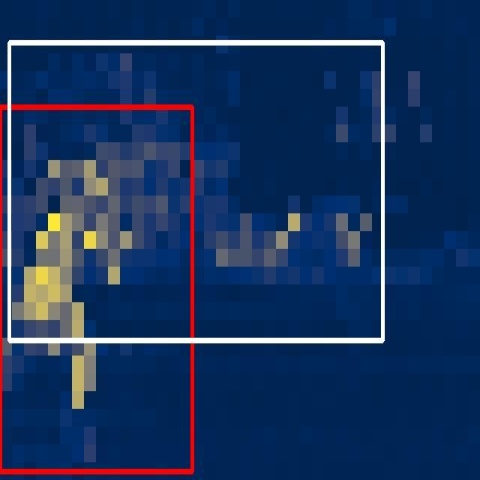} & \includegraphics[valign=B,width=0.102432\linewidth]{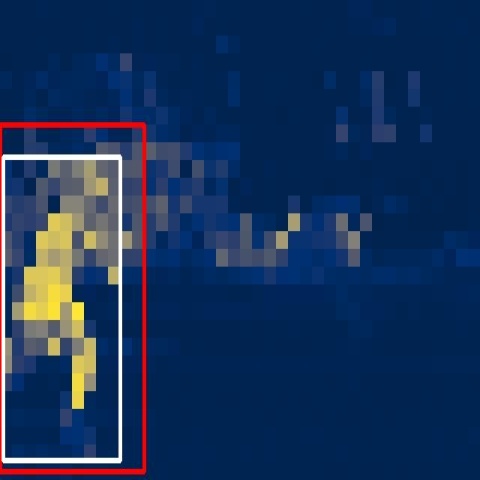} &
\includegraphics[valign=B,width=0.102432\linewidth]{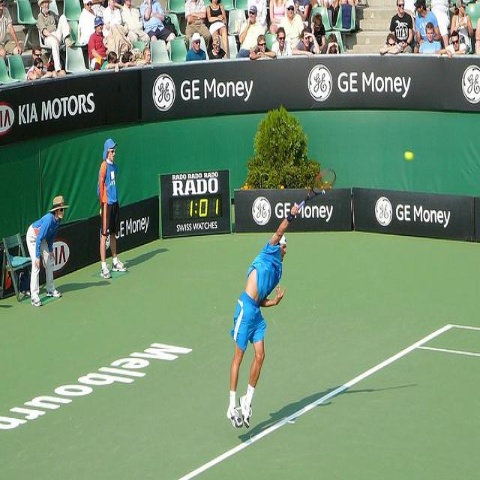} & \includegraphics[valign=B,width=0.102432\linewidth]{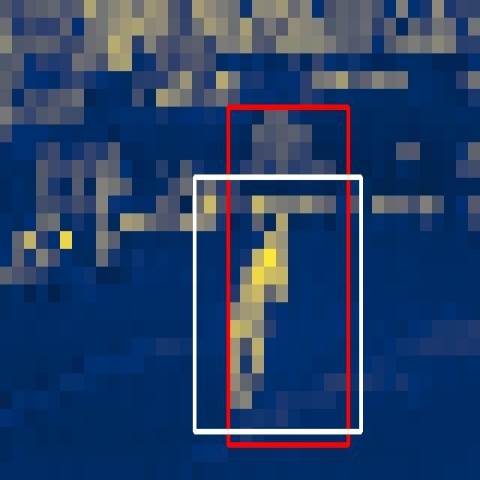} & \includegraphics[valign=B,width=0.102432\linewidth]{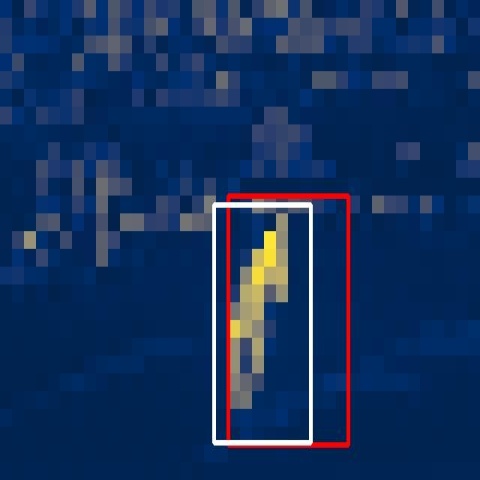} \\
\includegraphics[valign=B,width=0.102432\linewidth]{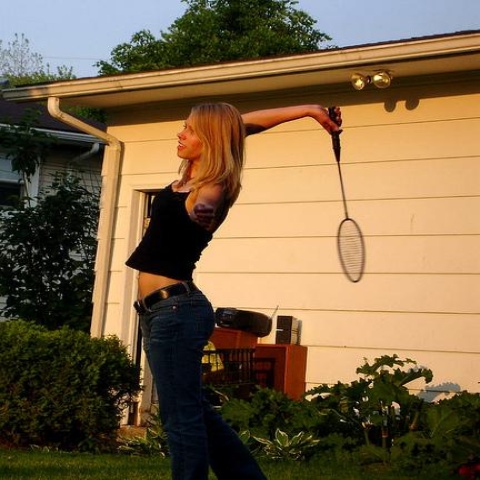} & \includegraphics[valign=B,width=0.102432\linewidth]{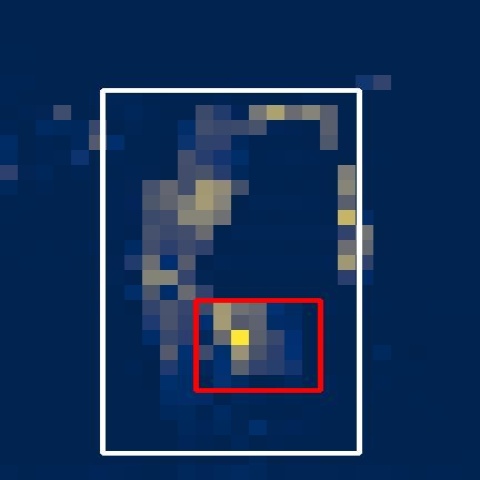} & \includegraphics[valign=B,width=0.102432\linewidth]{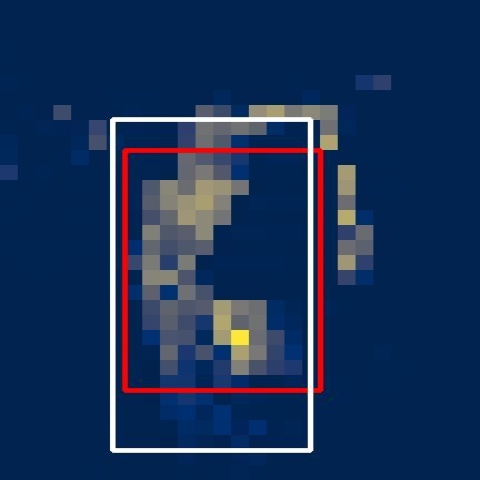} &
\includegraphics[valign=B,width=0.102432\linewidth]{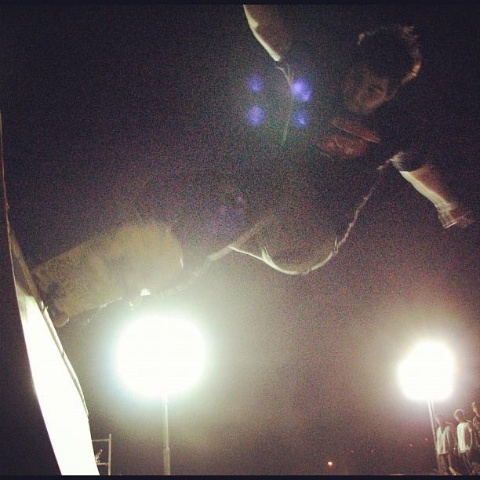} & \includegraphics[valign=B,width=0.102432\linewidth]{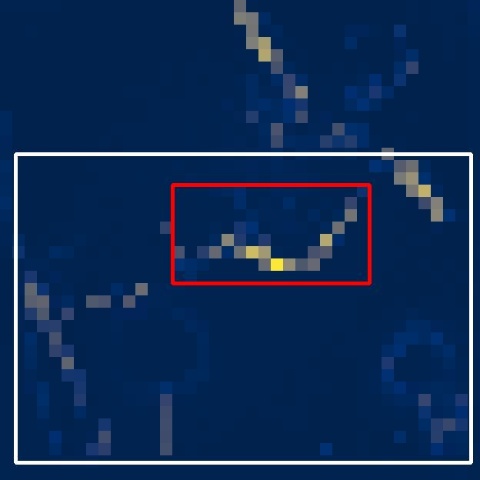} & \includegraphics[valign=B,width=0.102432\linewidth]{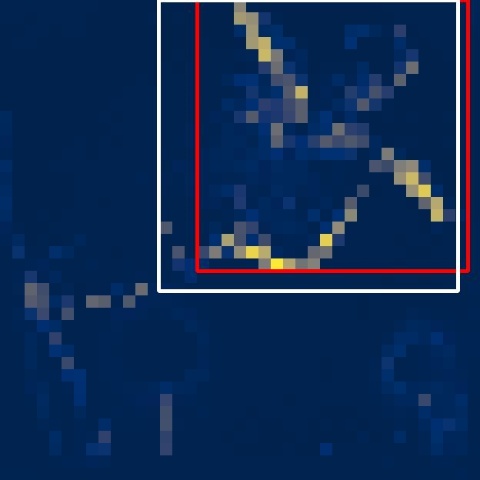} \\
\end{tabular}

   \caption{Single object discovery results on MS-COCO14\cite{cocoeccv2014} dataset. (a) the input image (b) the inverse degree of the LOST~\cite{simeoni2021lost}; the red bounding box is directly from LOST, the white is the prediction of CAD trained on top of it (c) same with our refined model $f^h$ and LOST (d) same as a (e) same as b (f) same as c}
   \label{fig:appendix_uod_lost_coco}
\end{figure*}

\begin{figure*}[h!]
    \setlength{\tabcolsep}{3pt} 
    \renewcommand{\arraystretch}{1}
     \centering
   \begin{tabular}{cccccc}
\includegraphics[valign=B,width=0.102432\linewidth]{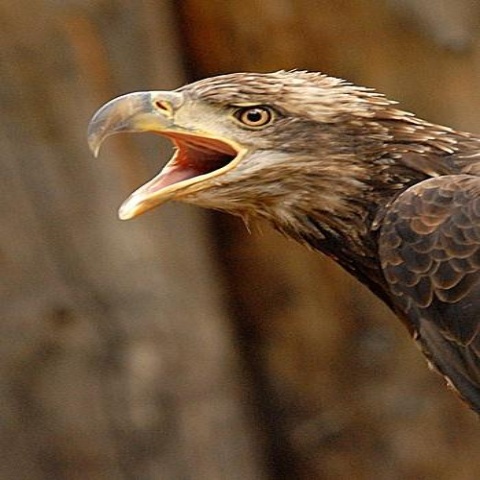} & \includegraphics[valign=B,width=0.102432\linewidth]{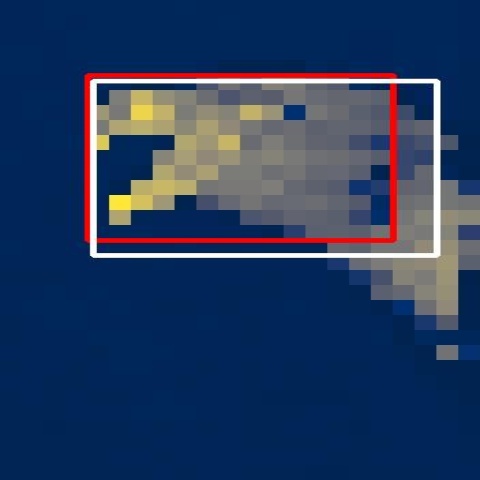} & \includegraphics[valign=B,width=0.102432\linewidth]{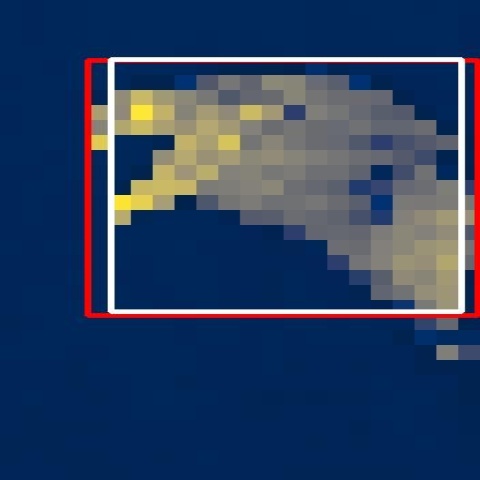} &
\includegraphics[valign=B,width=0.102432\linewidth]{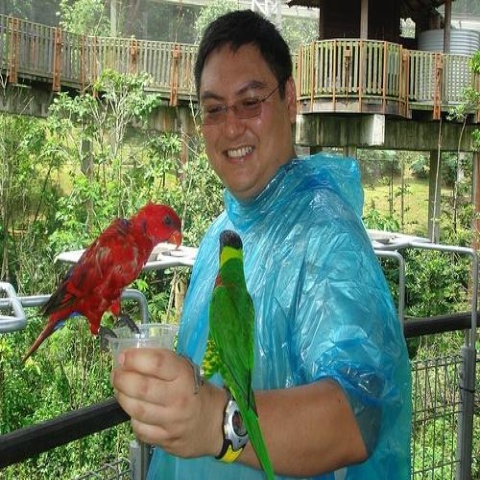} & \includegraphics[valign=B,width=0.102432\linewidth]{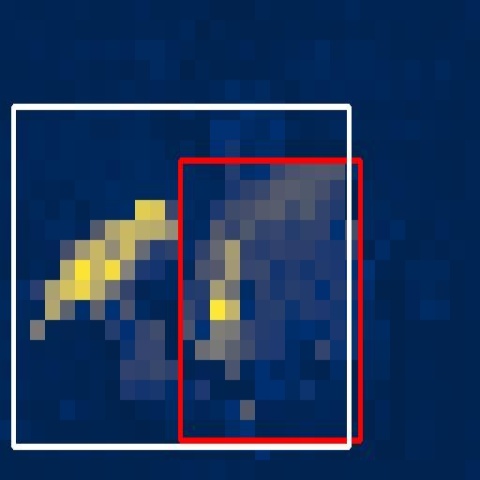} & \includegraphics[valign=B,width=0.102432\linewidth]{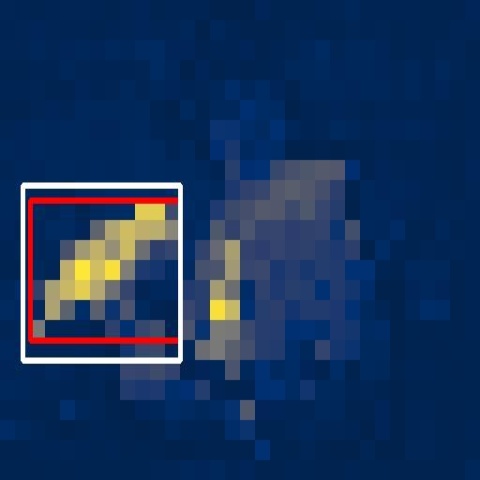} \\
\includegraphics[valign=B,width=0.102432\linewidth]{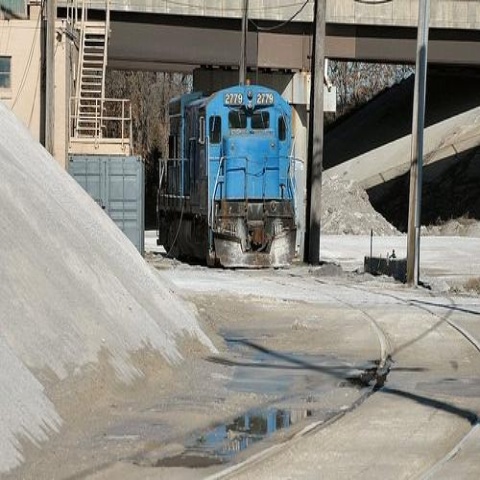} & \includegraphics[valign=B,width=0.102432\linewidth]{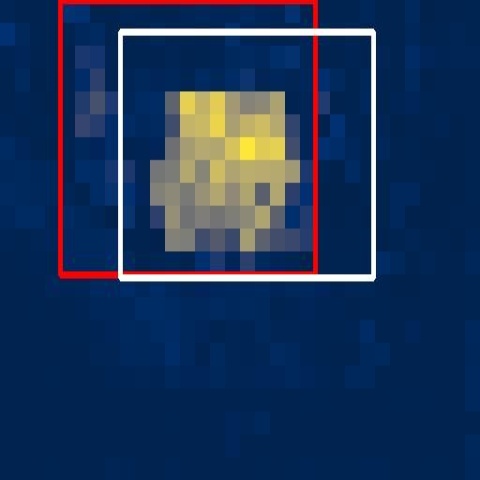} & \includegraphics[valign=B,width=0.102432\linewidth]{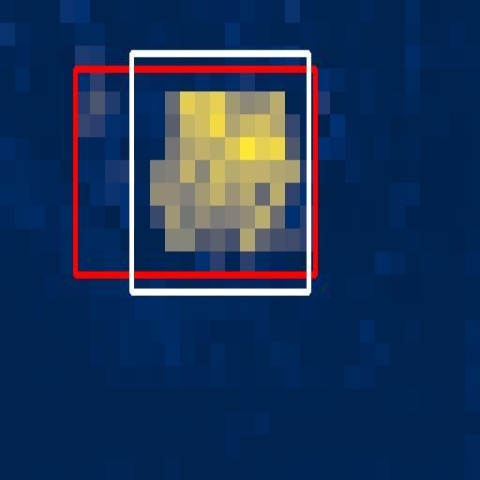} &
\includegraphics[valign=B,width=0.102432\linewidth]{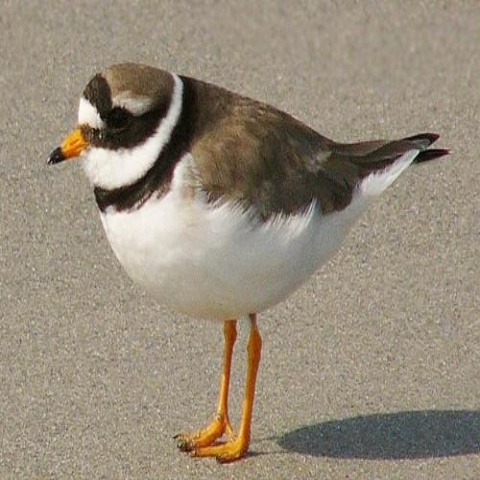} & \includegraphics[valign=B,width=0.102432\linewidth]{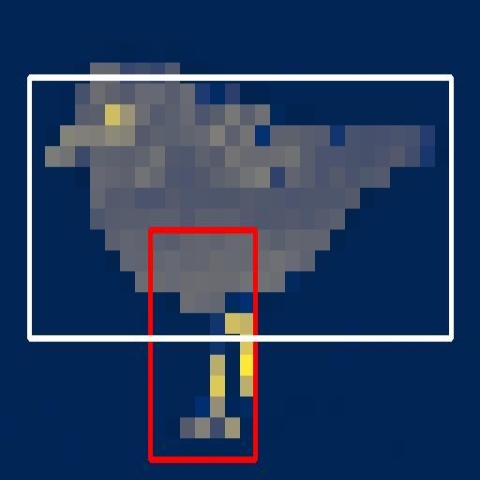} & \includegraphics[valign=B,width=0.102432\linewidth]{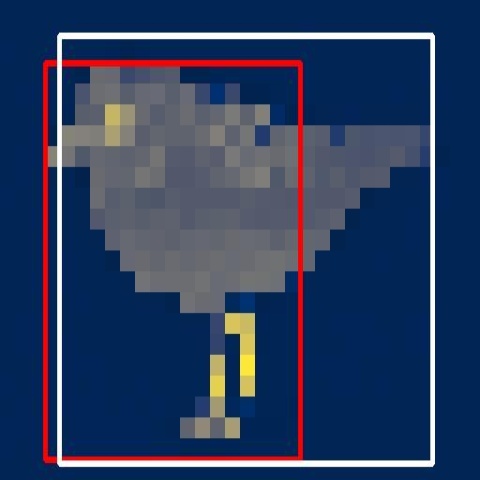} \\
\includegraphics[valign=B,width=0.102432\linewidth]{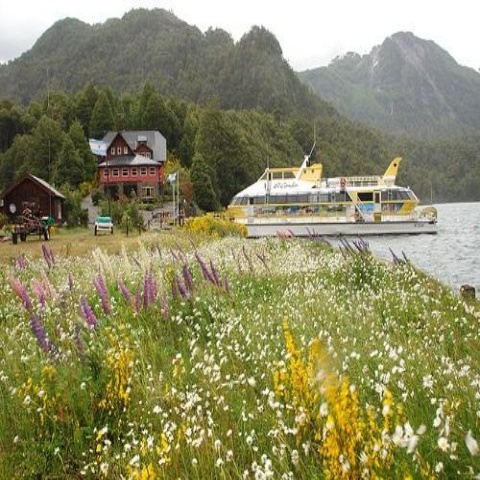} & \includegraphics[valign=B,width=0.102432\linewidth]{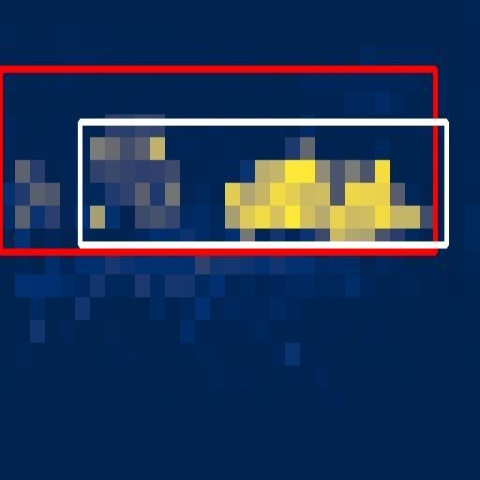} & \includegraphics[valign=B,width=0.102432\linewidth]{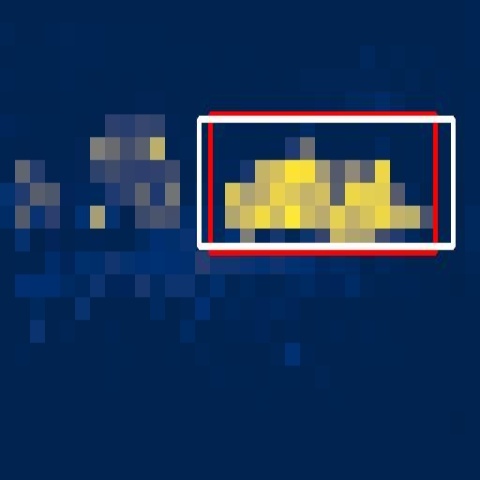} &
\includegraphics[valign=B,width=0.102432\linewidth]{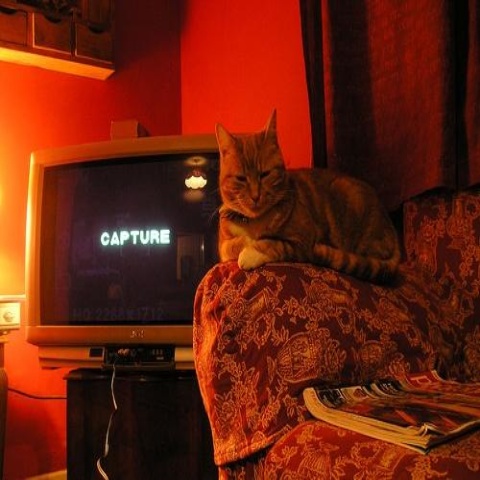} & \includegraphics[valign=B,width=0.102432\linewidth]{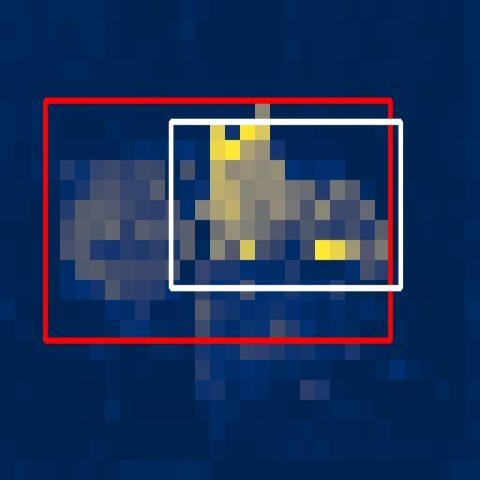} & \includegraphics[valign=B,width=0.102432\linewidth]{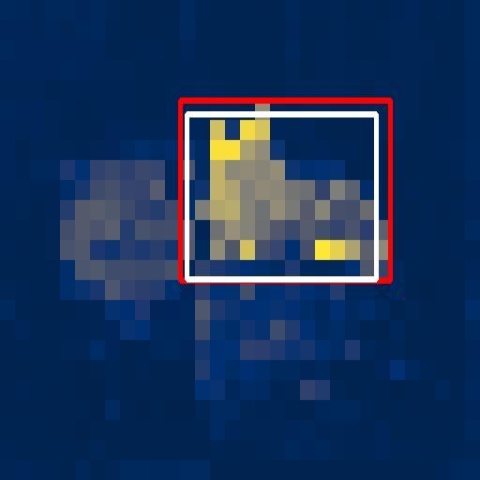} \\
\end{tabular}

   \caption{Single object discovery results on PASCAL-VOC07\cite{pascalvoc2007} dataset. (a) the input image (b) the inverse degree of the LOST~\cite{simeoni2021lost}; the red bounding box is directly from LOST, the white is the prediction of CAD trained on top of it (c) same with our refined model $f^h$ and LOST (d) same as a (e) same as b (f) same as c}
   \label{fig:appendix_uod_lost_voc07}
\end{figure*}

\begin{figure*}[h!]
    \setlength{\tabcolsep}{3pt} 
    \renewcommand{\arraystretch}{1}
     \centering
\begin{tabular}{cccccc}
\includegraphics[valign=B,width=0.1432\linewidth]{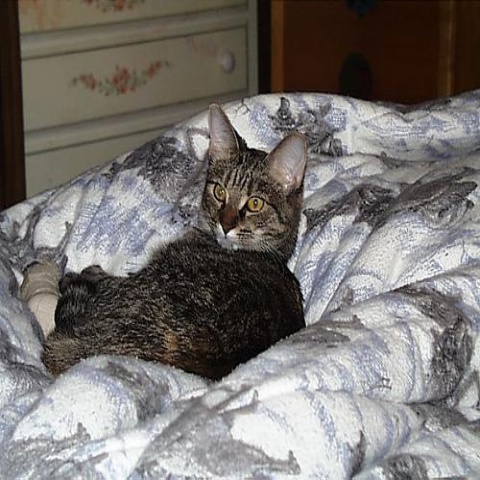} & \includegraphics[valign=B,width=0.1432\linewidth]{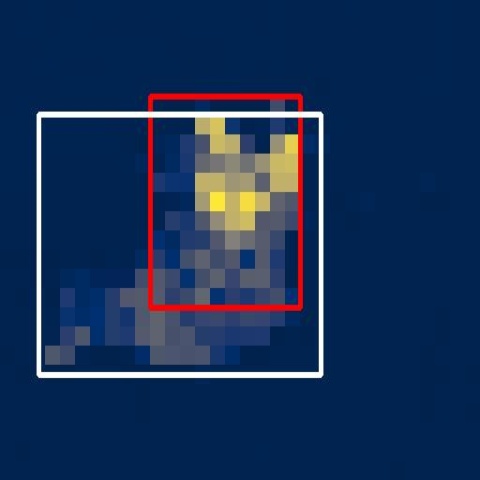} & \includegraphics[valign=B,width=0.1432\linewidth]{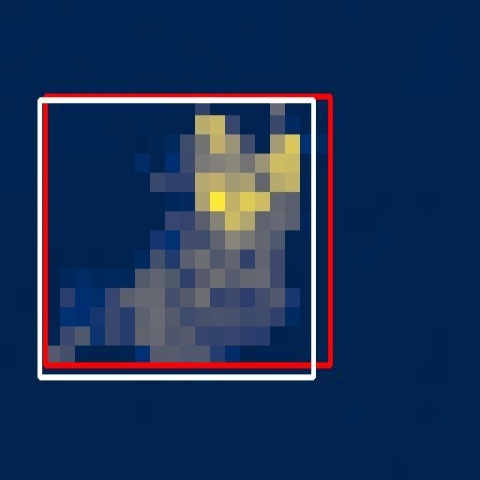} &
\includegraphics[valign=B,width=0.1432\linewidth]{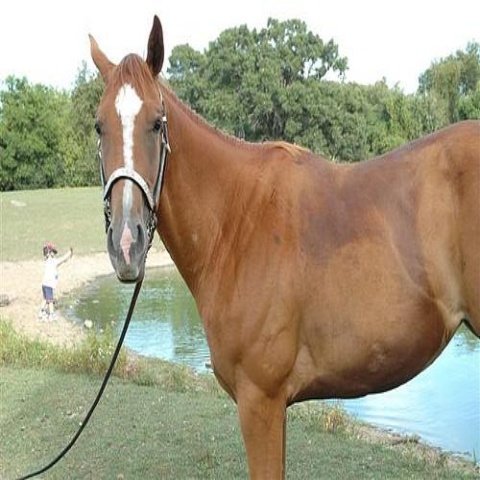} & \includegraphics[valign=B,width=0.1432\linewidth]{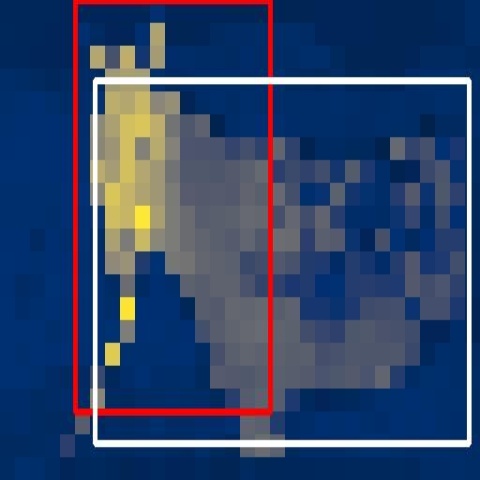} & \includegraphics[valign=B,width=0.1432\linewidth]{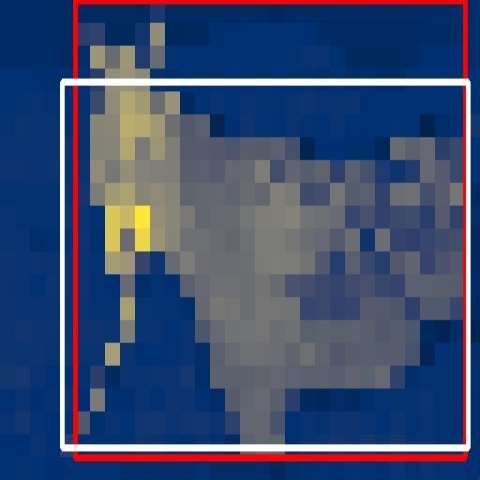} \\
\includegraphics[valign=B,width=0.1432\linewidth]{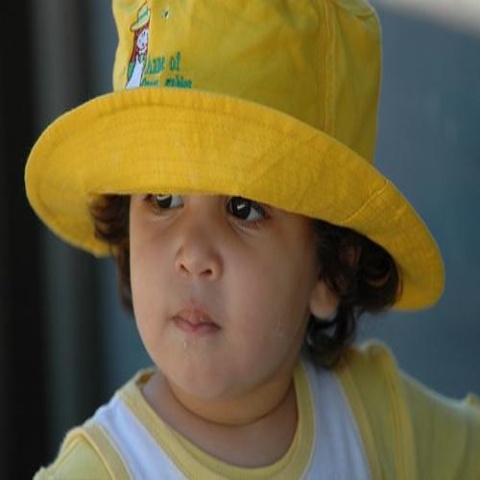} & \includegraphics[valign=B,width=0.1432\linewidth]{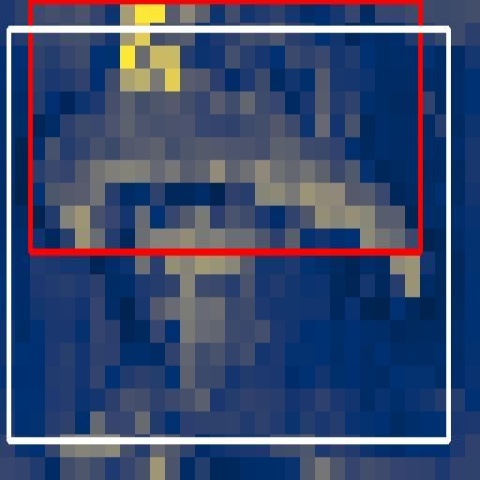} & \includegraphics[valign=B,width=0.1432\linewidth]{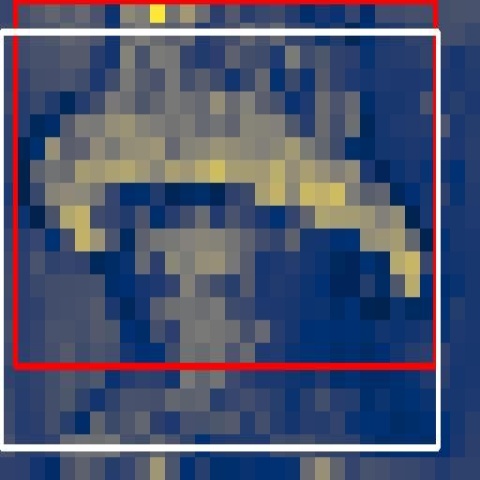} &
\includegraphics[valign=B,width=0.1432\linewidth]{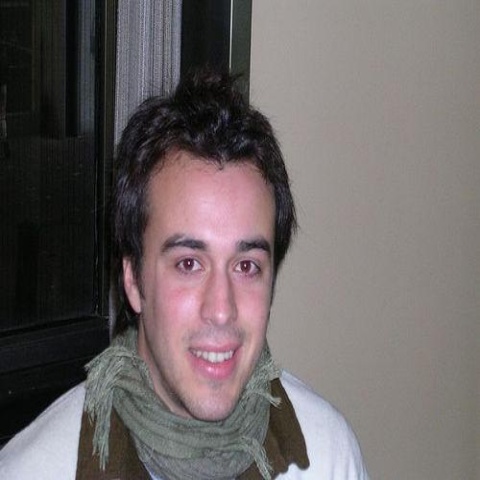} & \includegraphics[valign=B,width=0.1432\linewidth]{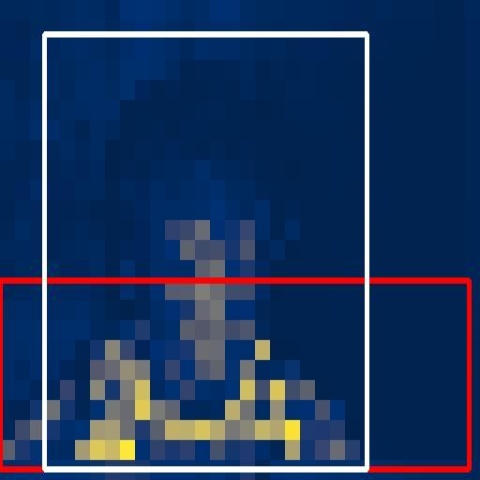} & \includegraphics[valign=B,width=0.1432\linewidth]{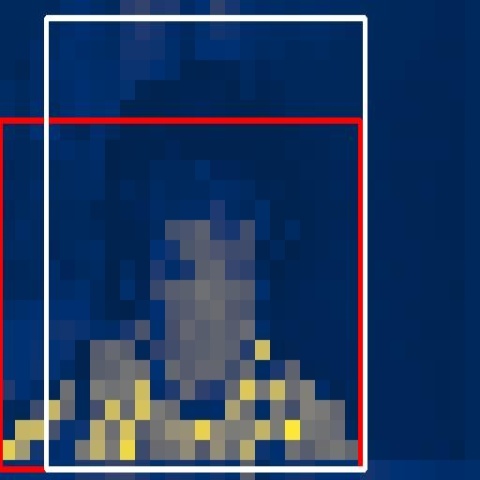} \\
\includegraphics[valign=B,width=0.1432\linewidth]{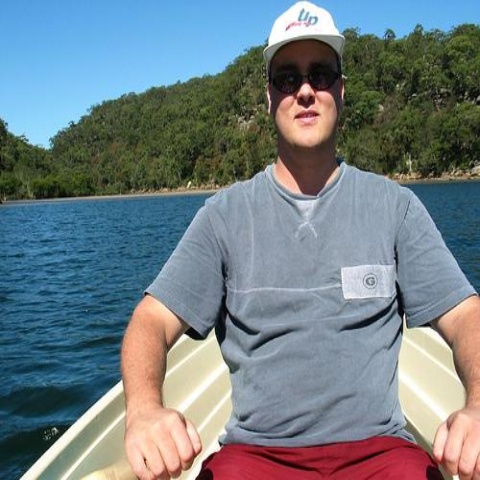} & \includegraphics[valign=B,width=0.1432\linewidth]{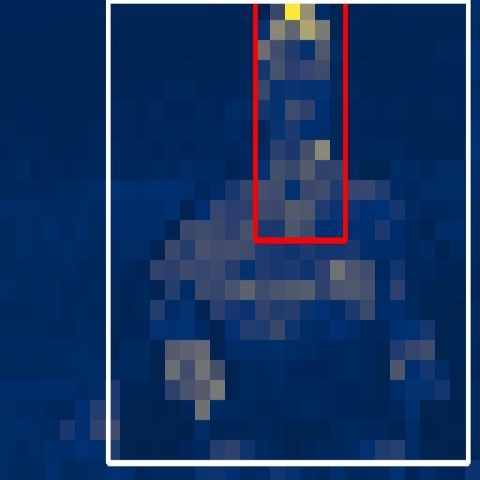} & \includegraphics[valign=B,width=0.1432\linewidth]{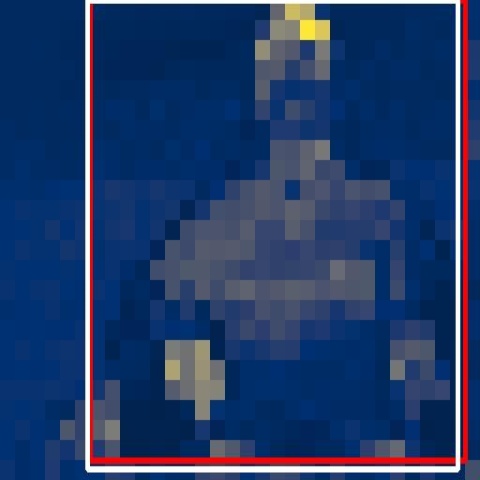} &
\includegraphics[valign=B,width=0.1432\linewidth]{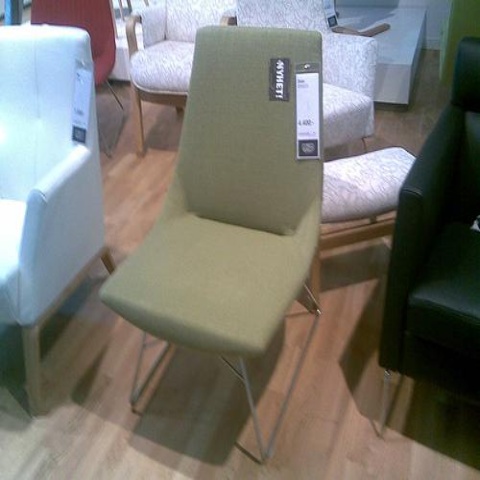} & \includegraphics[valign=B,width=0.1432\linewidth]{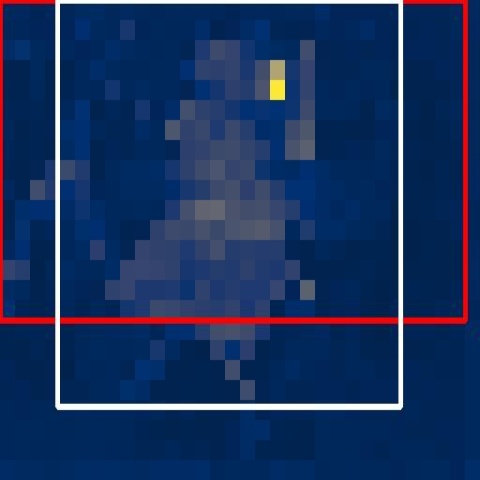} & \includegraphics[valign=B,width=0.1432\linewidth]{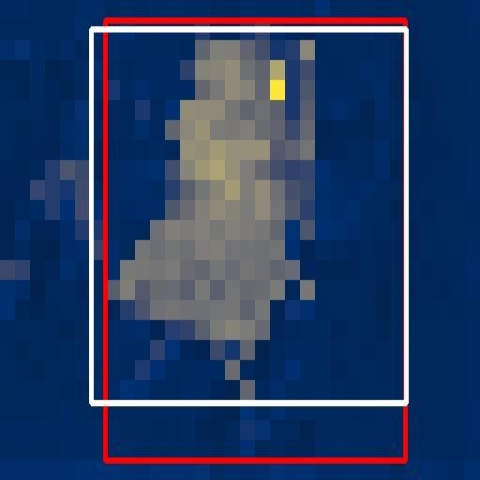} \\
\end{tabular}
   \caption{Single object discovery results on PASCAL-VOC12\cite{pascalvoc2012} dataset. (a) the input image (b) the inverse degree of the LOST~\cite{simeoni2021lost}; the red bounding box is directly from LOST, the white is the prediction of CAD trained on top of it (c) same with our refined model $f^h$ and LOST (d) same as a (e) same as b (f) same as c}
   \label{fig:appendix_uod_lost_voc12}
\end{figure*}

\begin{figure*}[h!]
    \setlength{\tabcolsep}{3pt}
    \renewcommand{\arraystretch}{1} 
     \centering
\begin{tabular}{cccccc}
\includegraphics[valign=B,width=0.1432\linewidth]{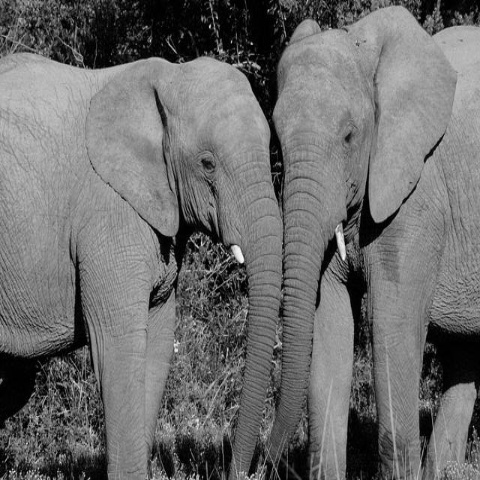} & \includegraphics[valign=B,width=0.1432\linewidth]{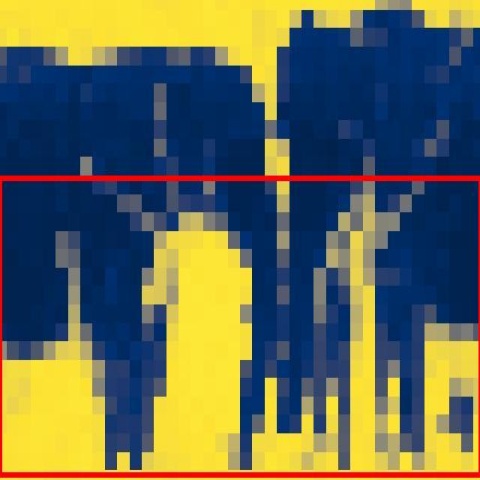} & \includegraphics[valign=B,width=0.1432\linewidth]{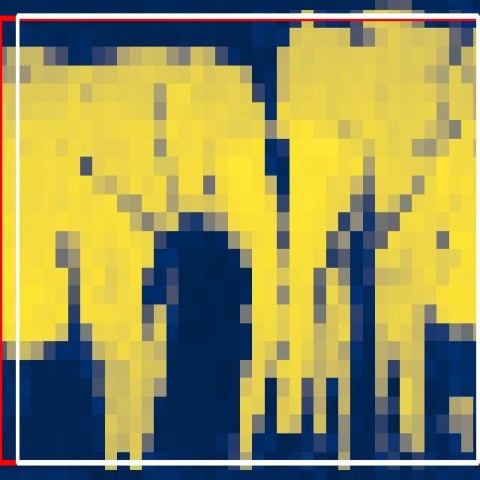} &
\includegraphics[valign=B,width=0.1432\linewidth]{vis/appendix/uod/tokencut/coco/52650_im_tokencut.jpg} & \includegraphics[valign=B,width=0.1432\linewidth]{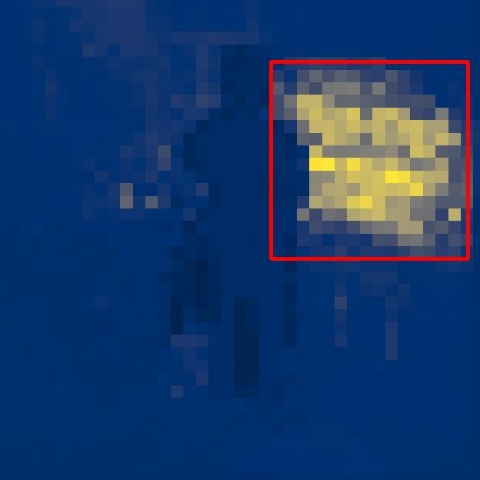} & \includegraphics[valign=B,width=0.1432\linewidth]{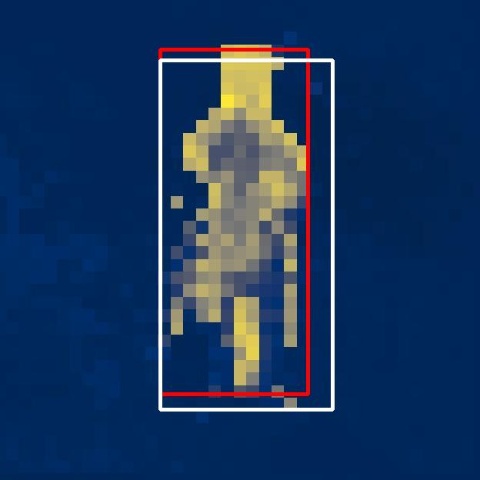} \\
\includegraphics[valign=B,width=0.1432\linewidth]{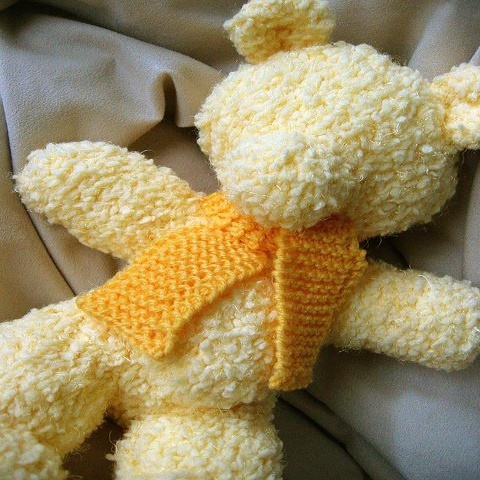} & \includegraphics[valign=B,width=0.1432\linewidth]{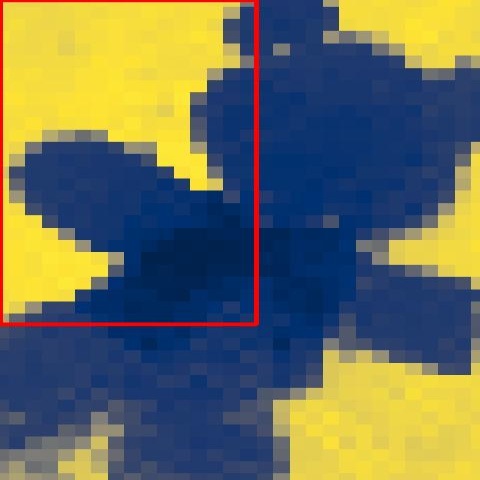} & \includegraphics[valign=B,width=0.1432\linewidth]{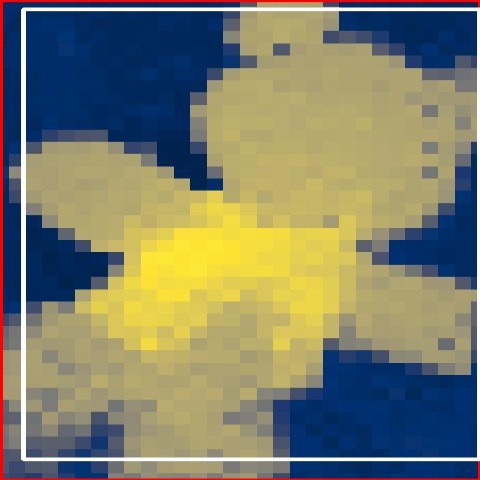} &
\includegraphics[valign=B,width=0.1432\linewidth]{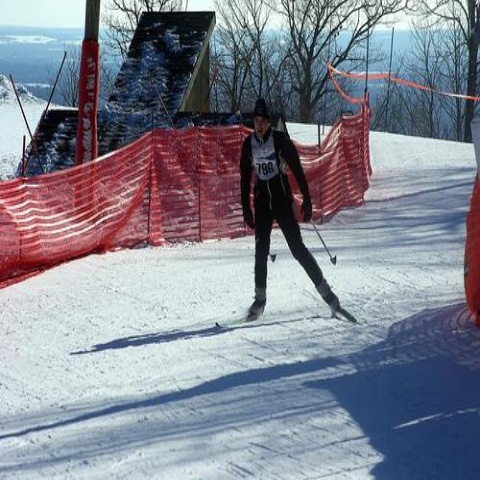} & \includegraphics[valign=B,width=0.1432\linewidth]{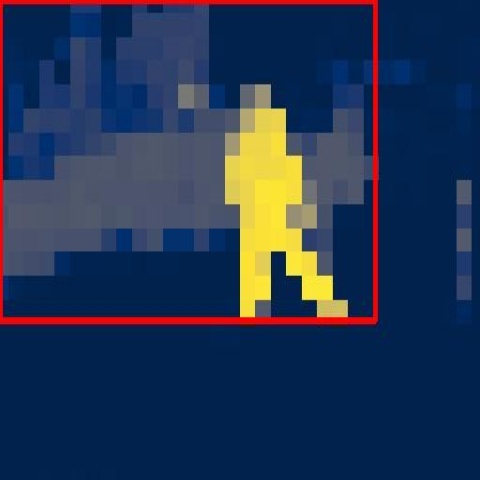} & \includegraphics[valign=B,width=0.1432\linewidth]{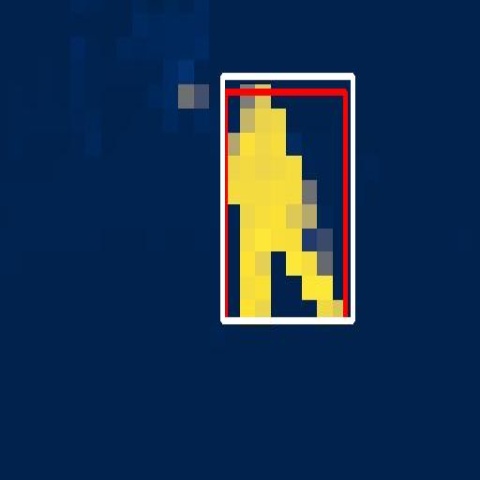} \\
\includegraphics[valign=B,width=0.1432\linewidth]{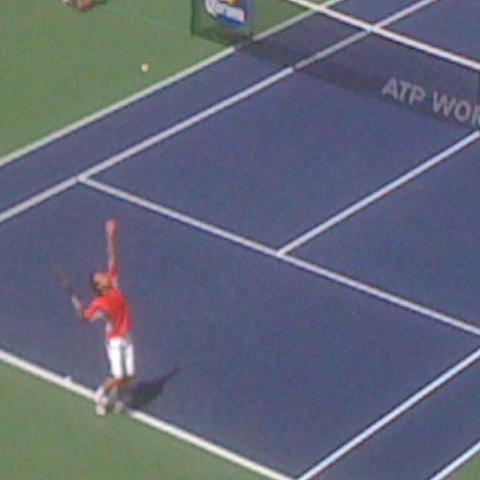} & \includegraphics[valign=B,width=0.1432\linewidth]{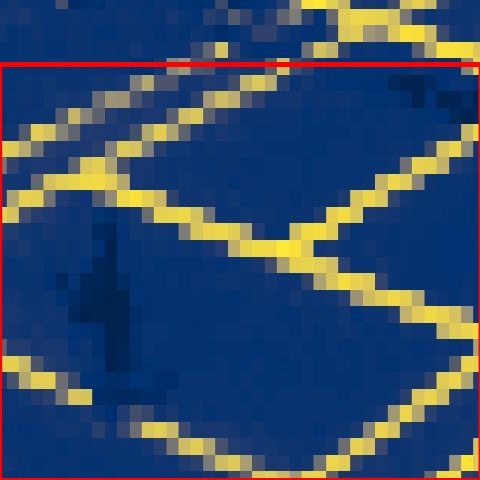} & \includegraphics[valign=B,width=0.1432\linewidth]{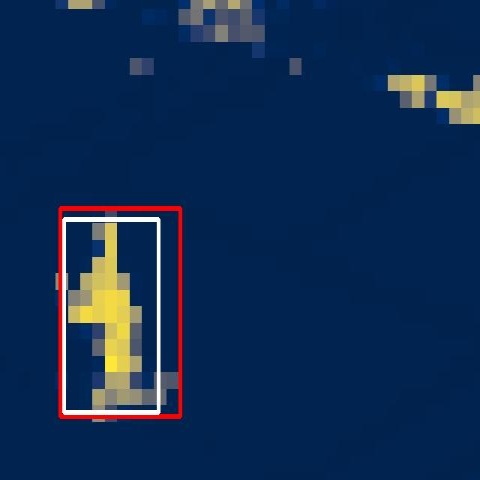} &
\includegraphics[valign=B,width=0.1432\linewidth]{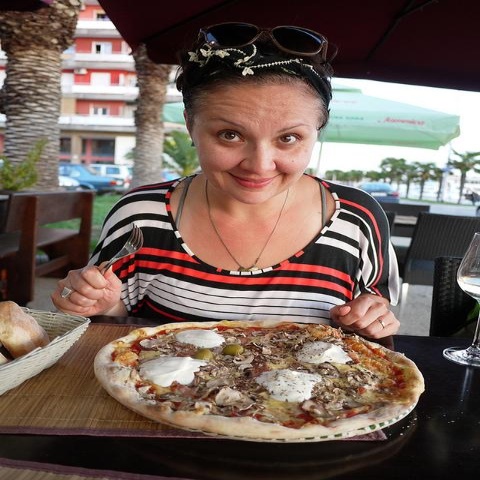} & \includegraphics[valign=B,width=0.1432\linewidth]{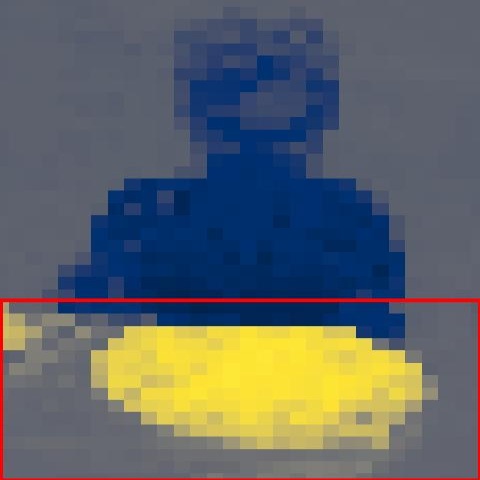} & \includegraphics[valign=B,width=0.1432\linewidth]{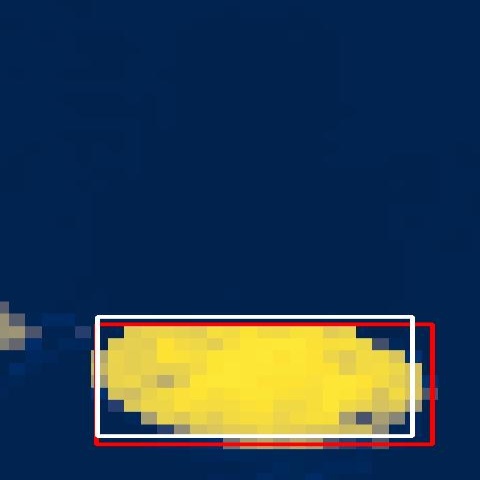} \\
\end{tabular}
   \caption{Single object discovery results on MS-COCO14\cite{cocoeccv2014} dataset. (a) the input image (b) the eigenvector attention of the TokenCut~\cite{wang2022tokencut}; the red bounding box is directly from TokenCut (the CAD model was not released and is not shown) (c) same with our refined model $f^h$ and TokenCut, the white bounding box is the prediction of CAD trained on top of $f^h$ (d) same as a (e) same as b (f) same as c}
   \label{fig:appendix_uod_tokencut_coco}
\end{figure*}

\begin{figure*}[h!]
    \setlength{\tabcolsep}{3pt}
    \renewcommand{\arraystretch}{1}
     \centering
\begin{tabular}{cccccc}
\includegraphics[valign=B,width=0.1432\linewidth]{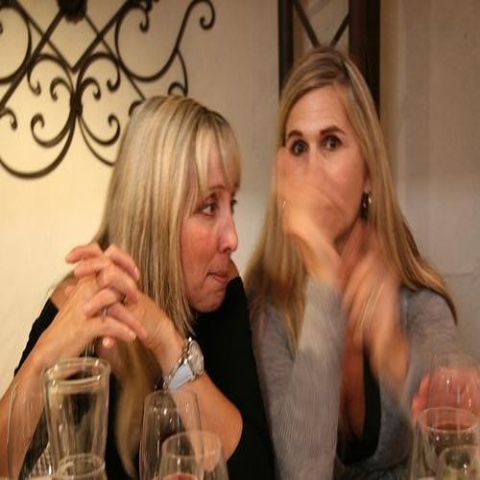} & \includegraphics[valign=B,width=0.1432\linewidth]{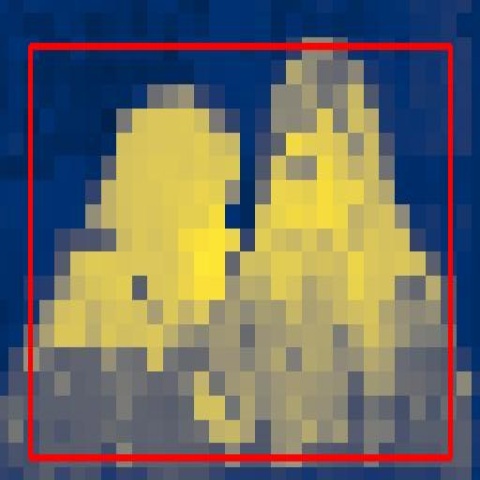} & \includegraphics[valign=B,width=0.1432\linewidth]{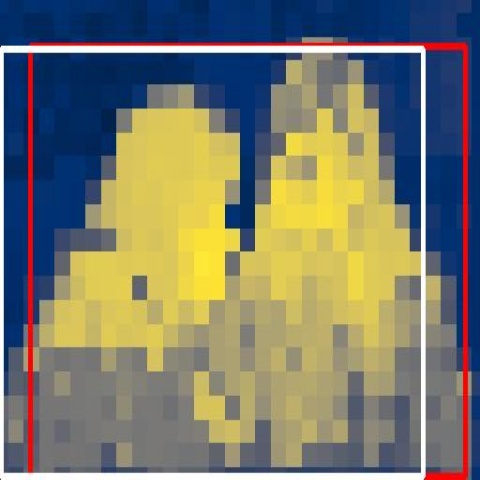} &
\includegraphics[valign=B,width=0.1432\linewidth]{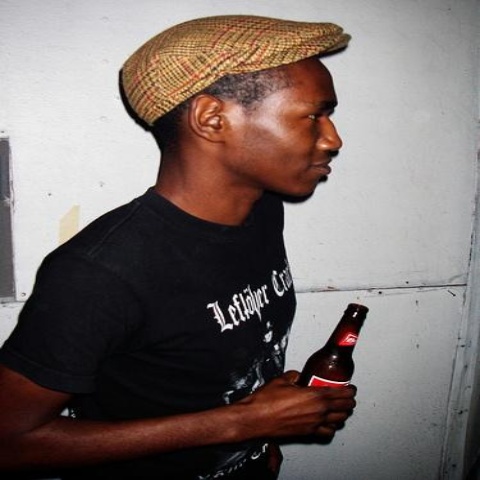} & \includegraphics[valign=B,width=0.1432\linewidth]{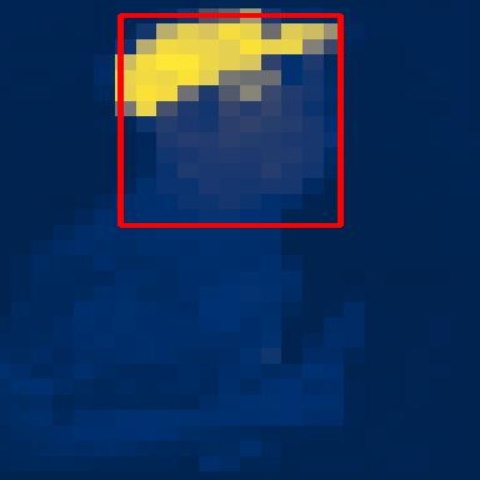} & \includegraphics[valign=B,width=0.1432\linewidth]{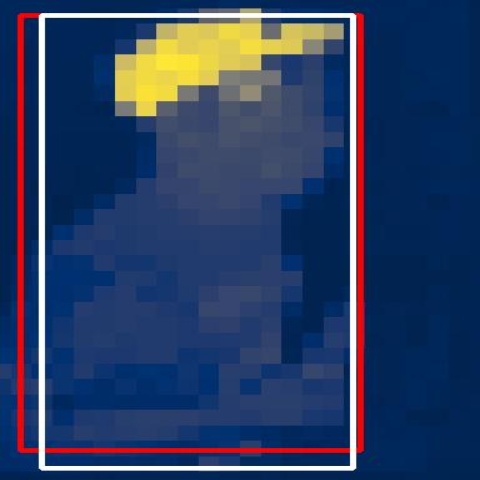} \\
\includegraphics[valign=B,width=0.1432\linewidth]{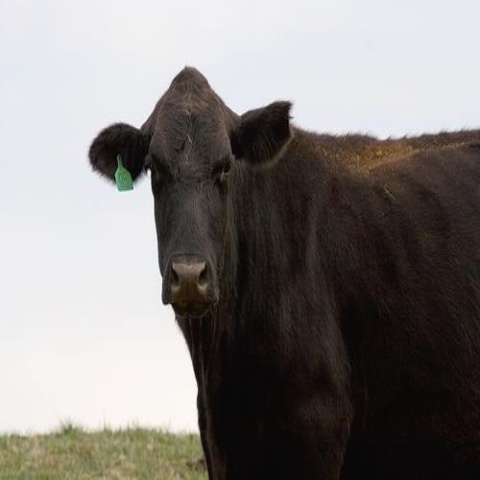} & \includegraphics[valign=B,width=0.1432\linewidth]{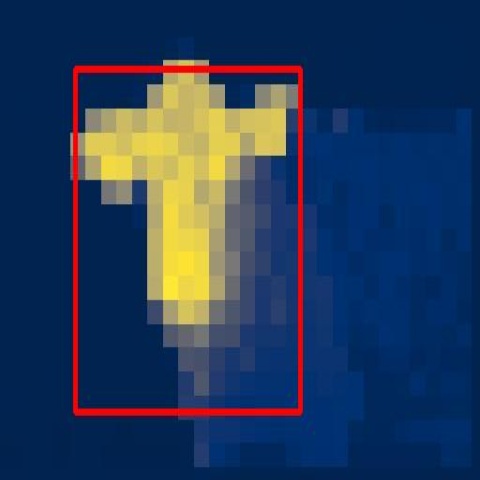} & \includegraphics[valign=B,width=0.1432\linewidth]{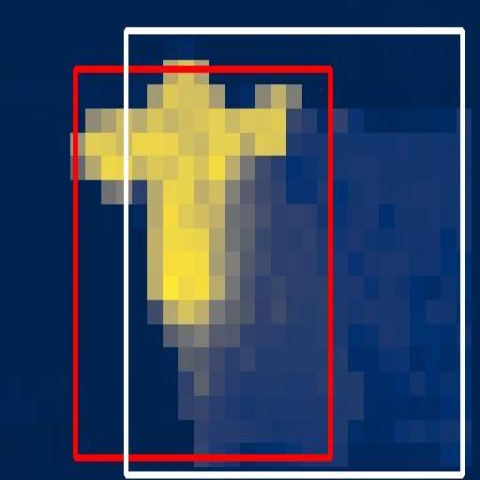} &
\includegraphics[valign=B,width=0.1432\linewidth]{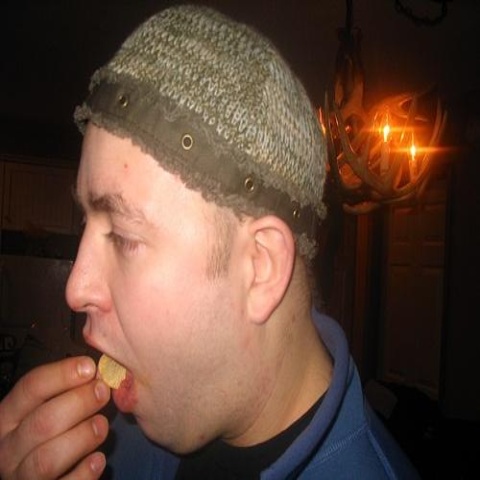} & \includegraphics[valign=B,width=0.1432\linewidth]{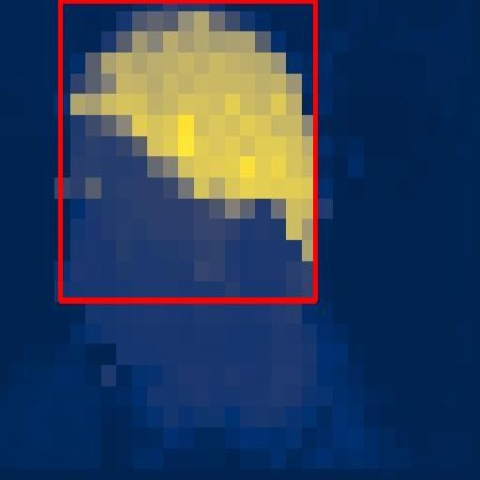} & \includegraphics[valign=B,width=0.1432\linewidth]{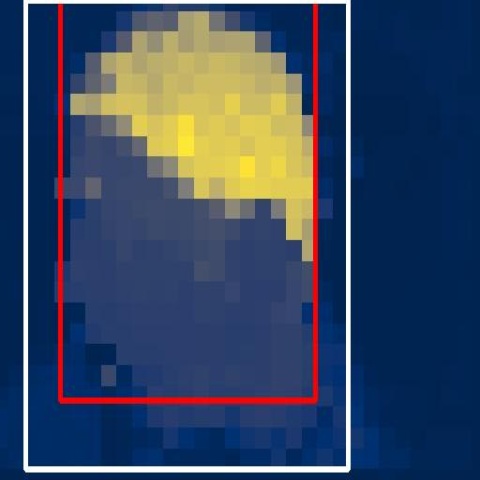} \\
\includegraphics[valign=B,width=0.1432\linewidth]{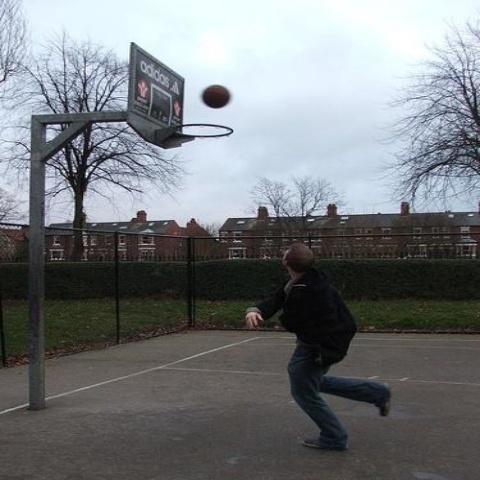}
& \includegraphics[valign=B,width=0.1432\linewidth]{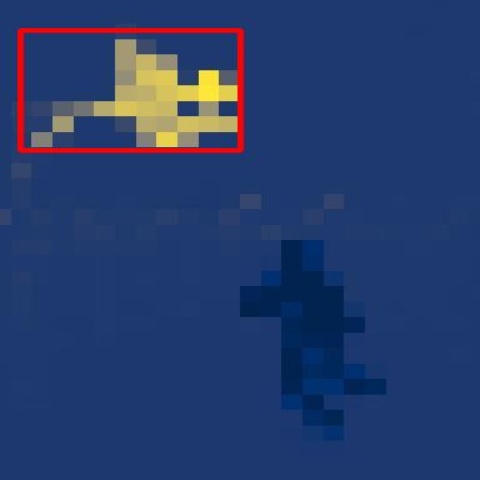} & \includegraphics[valign=B,width=0.1432\linewidth]{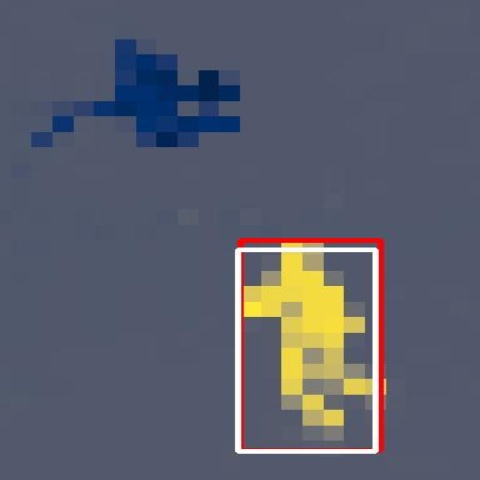} &
\includegraphics[valign=B,width=0.1432\linewidth]{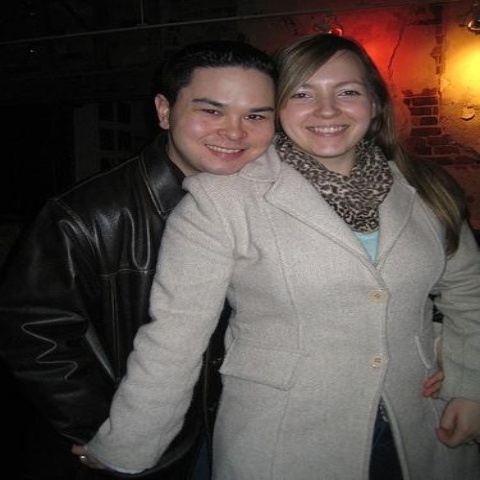} & \includegraphics[valign=B,width=0.1432\linewidth]{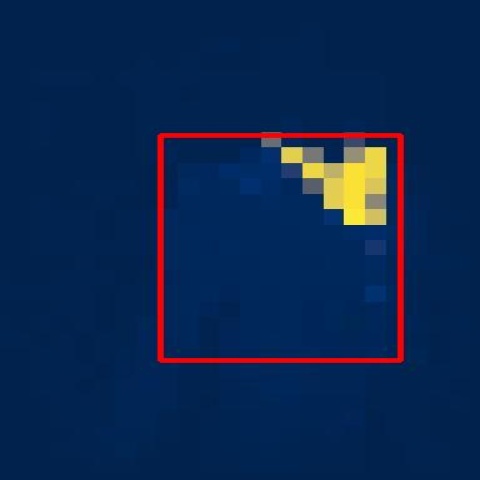} & \includegraphics[valign=B,width=0.1432\linewidth]{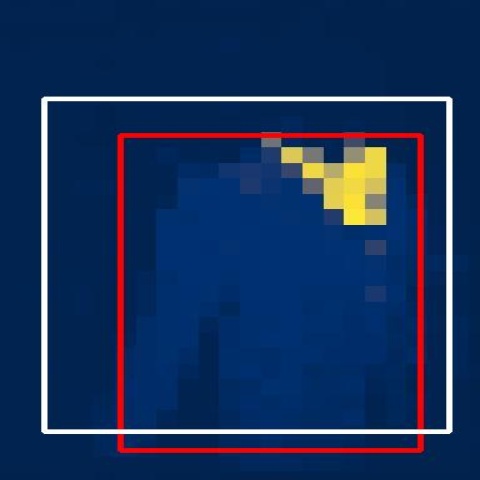} \\
        (a)  & (b) & (c) & (d) & (e) & (f) \\
   \end{tabular}
   \caption{Single object discovery results on PASCAL-VOC07\cite{pascalvoc2007} dataset. (a) the input image (b) the eigenvector attention of the TokenCut~\cite{wang2022tokencut}; the red bounding box is directly from TokenCut (the CAD model was not released and is not shown) (c) same with our refined model $f^h$ and TokenCut, the white bounding box is the prediction of CAD trained on top of $f^h$ (d) same as a (e) same as b (f) same as c}
   \label{fig:appendix_uod_tokencut_voc07}
\end{figure*}

\begin{figure*}[h!]
    \setlength{\tabcolsep}{3pt}
    \renewcommand{\arraystretch}{1}
     \centering
\begin{tabular}{cccccc}
\includegraphics[valign=B,width=0.1432\linewidth]{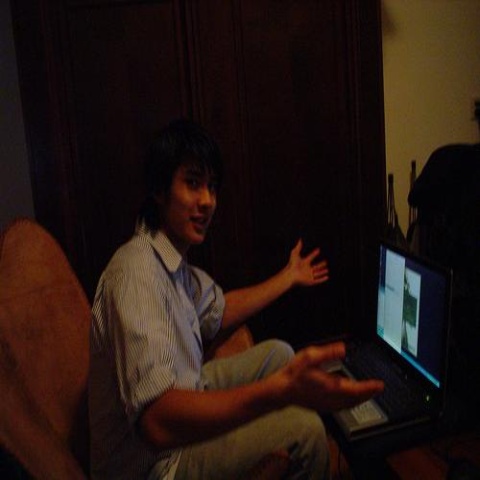} & \includegraphics[valign=B,width=0.1432\linewidth]{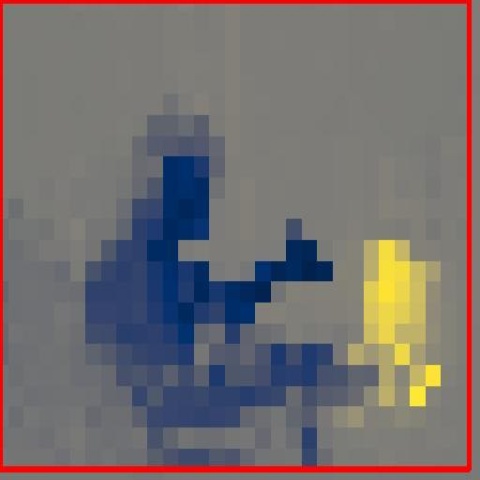} & \includegraphics[valign=B,width=0.1432\linewidth]{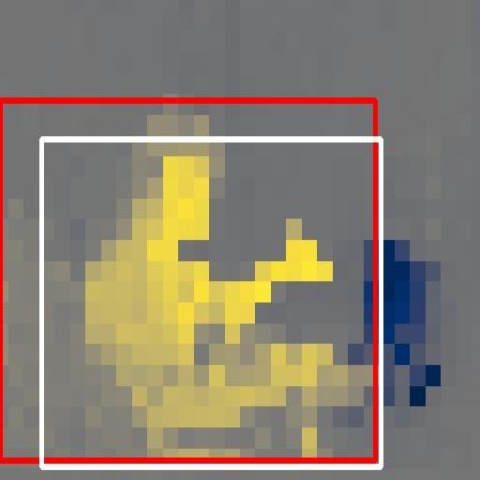} &
\includegraphics[valign=B,width=0.1432\linewidth]{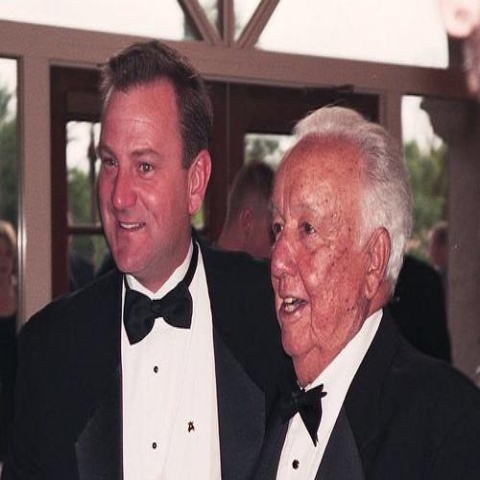} & \includegraphics[valign=B,width=0.1432\linewidth]{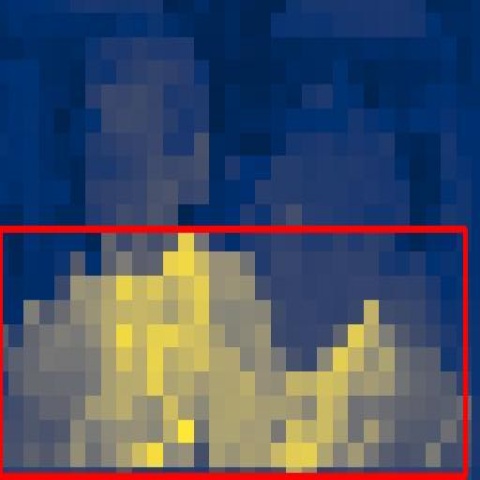} & \includegraphics[valign=B,width=0.1432\linewidth]{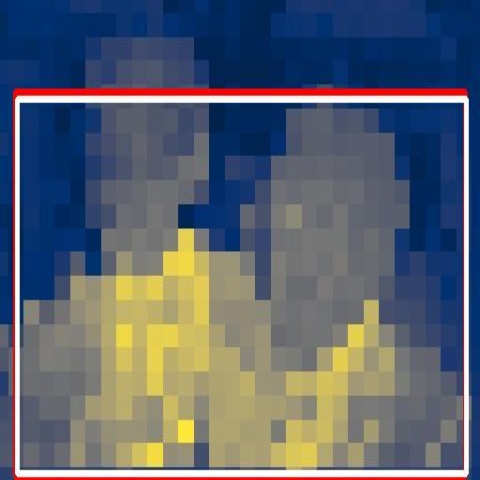} \\
\includegraphics[valign=B,width=0.1432\linewidth]{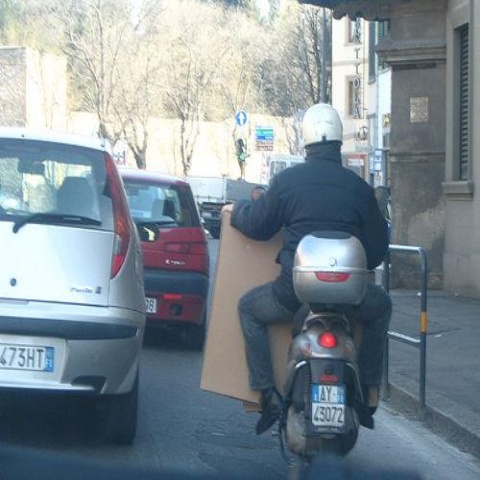} & \includegraphics[valign=B,width=0.1432\linewidth]{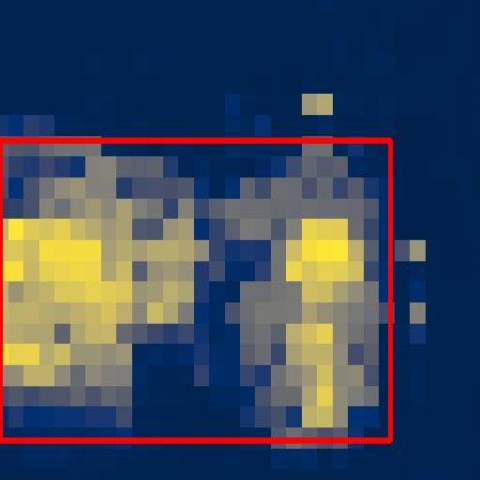} & \includegraphics[valign=B,width=0.1432\linewidth]{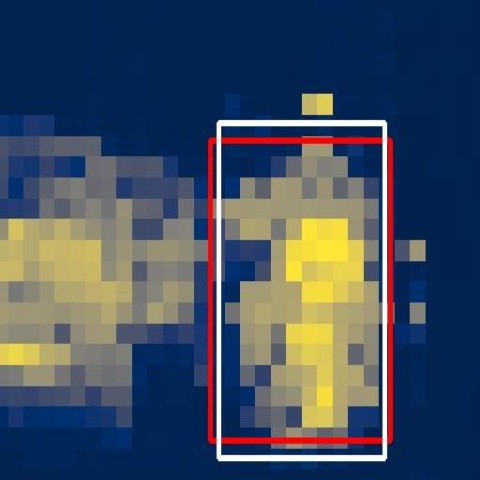} &
\includegraphics[valign=B,width=0.1432\linewidth]{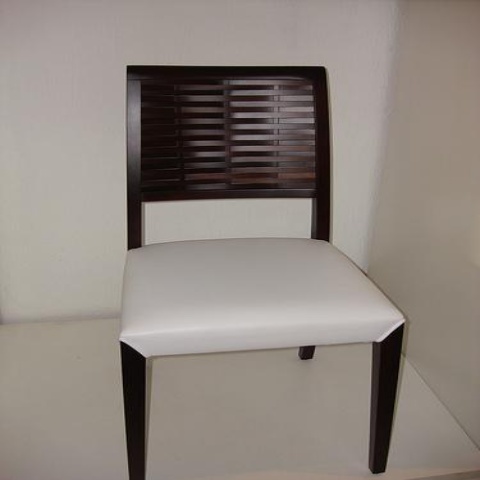} & \includegraphics[valign=B,width=0.1432\linewidth]{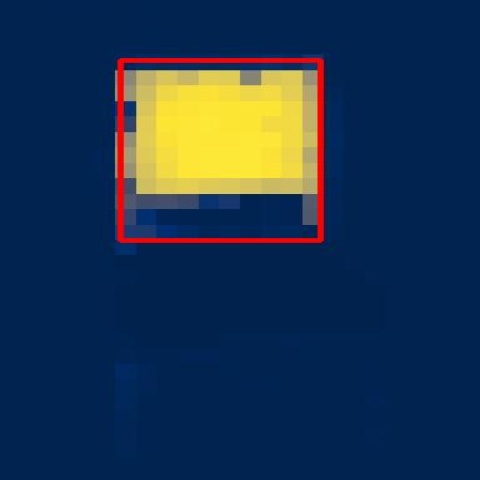} & \includegraphics[valign=B,width=0.1432\linewidth]{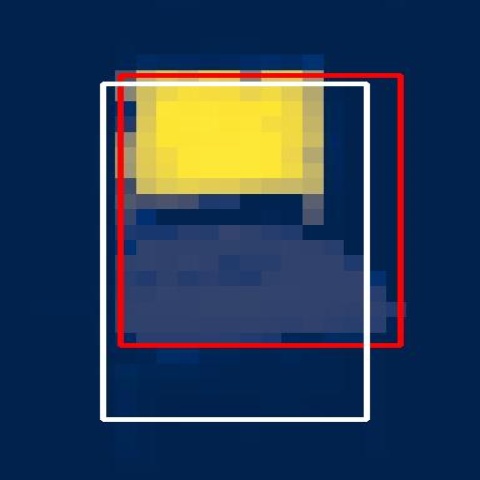} \\
\includegraphics[valign=B,width=0.1432\linewidth]{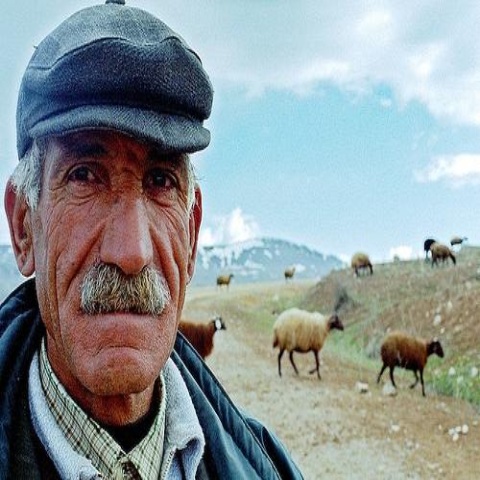} & \includegraphics[valign=B,width=0.1432\linewidth]{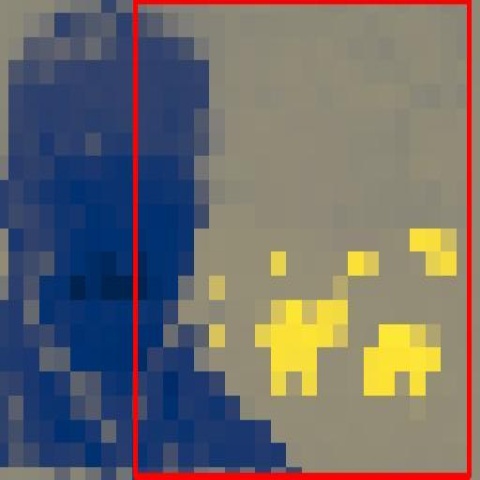} & \includegraphics[valign=B,width=0.1432\linewidth]{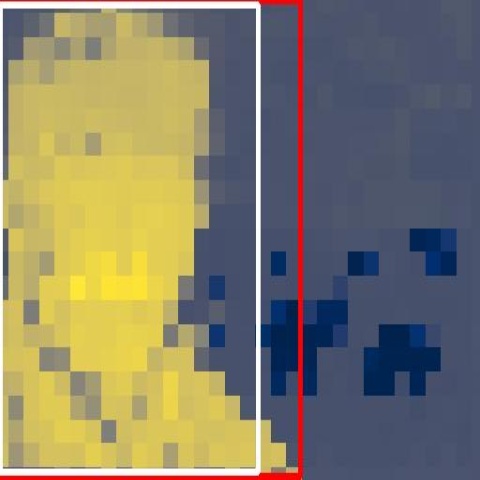} &
\includegraphics[valign=B,width=0.1432\linewidth]{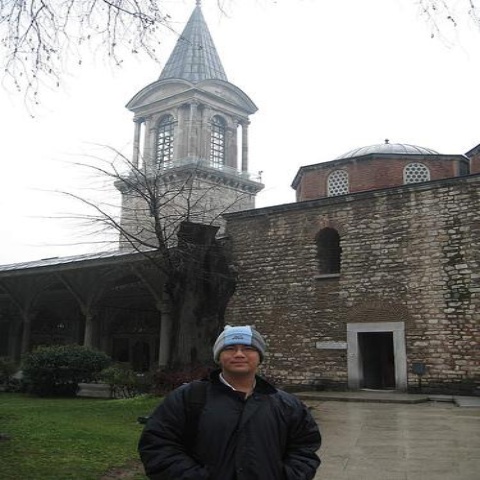} & \includegraphics[valign=B,width=0.1432\linewidth]{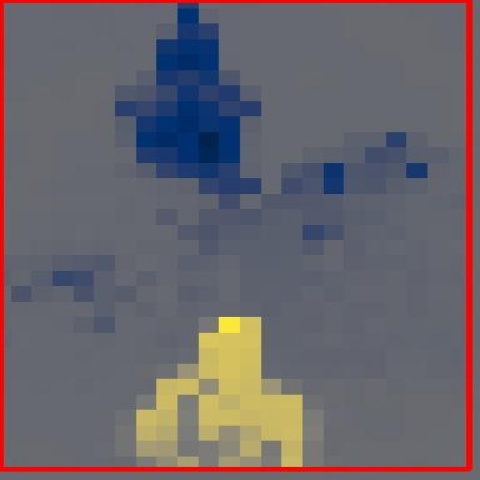} & \includegraphics[valign=B,width=0.1432\linewidth]{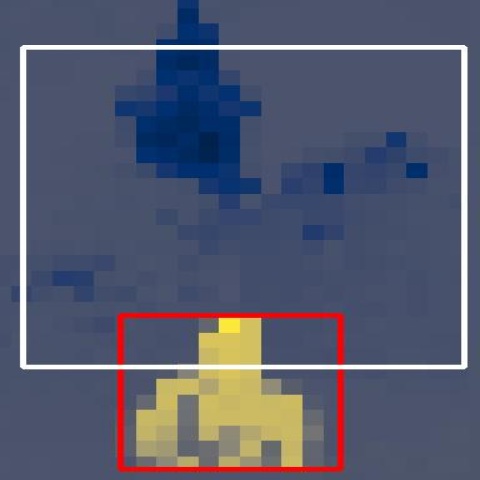} \\
\includegraphics[valign=B,width=0.1432\linewidth]{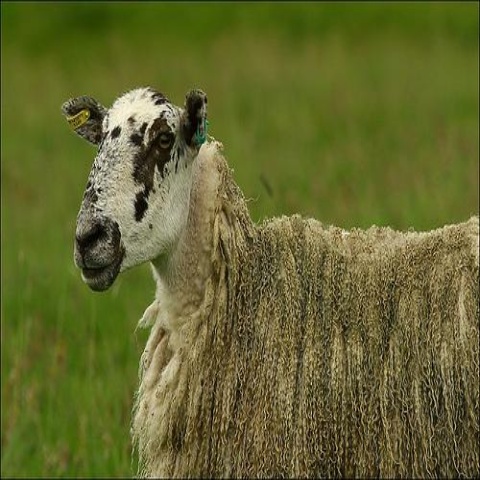} & \includegraphics[valign=B,width=0.1432\linewidth]{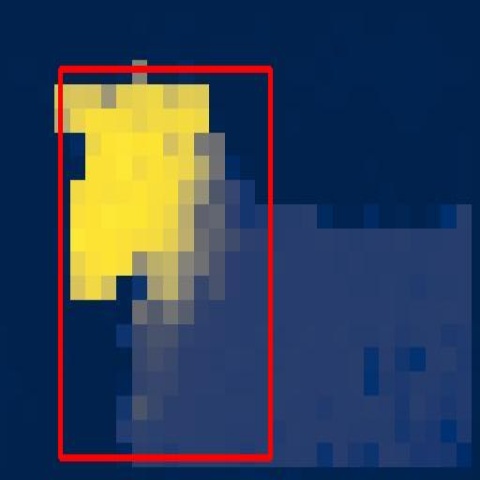} & \includegraphics[valign=B,width=0.1432\linewidth]{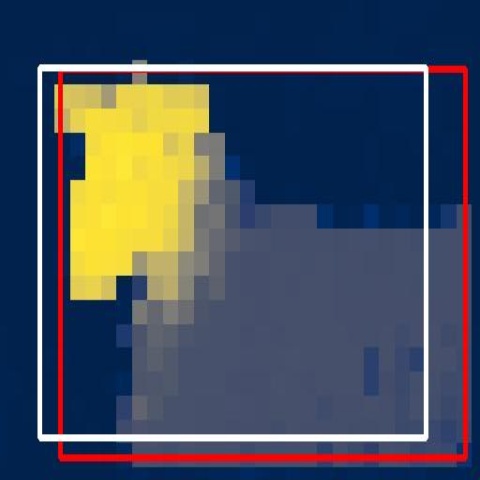} &
\includegraphics[valign=B,width=0.1432\linewidth]{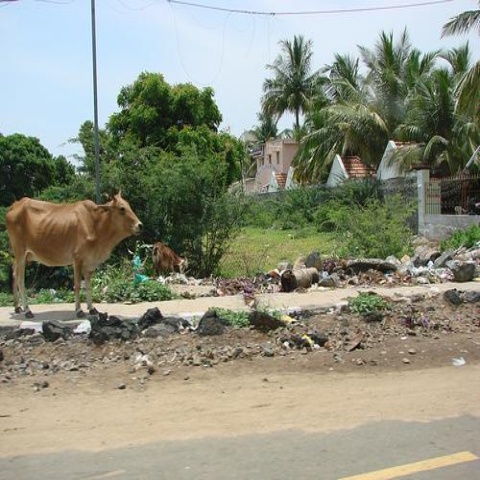} & \includegraphics[valign=B,width=0.1432\linewidth]{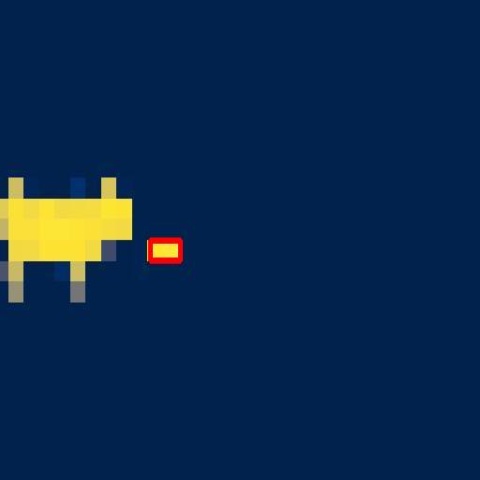} & \includegraphics[valign=B,width=0.1432\linewidth]{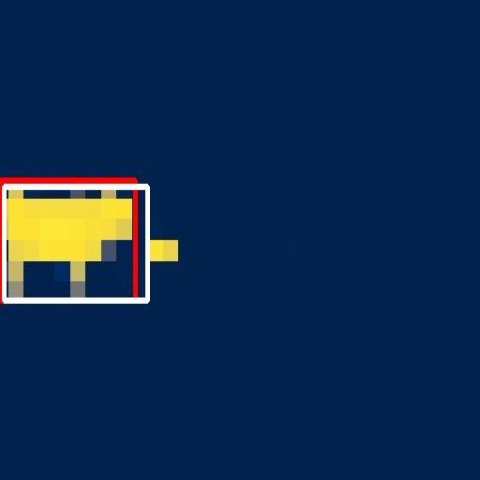} \\

        (a)  & (b) & (c) & (d) & (e) & (f) \\
   \end{tabular}
   \caption{Single object discovery results on PASCAL-VOC12\cite{pascalvoc2012} dataset. (a) the input image (b) the eigenvector attention of the TokenCut~\cite{wang2022tokencut}; the red bounding box is directly from TokenCut (the CAD model was not released and is not shown) (c) same with our refined model $f^h$ and TokenCut, the white bounding box is the prediction of CAD trained on top of $f^h$ (d) same as a (e) same as b (f) same as c}
   \label{fig:appendix_uod_tokencut_voc12}
\end{figure*}

In the context of the unsupervised single object discovery task, we display visualizations of our model $f^h$, which is based on the DINO\cite{caron2021emerging} model $f$. We compare our findings with those of LOST\cite{simeoni2021lost} and TokenCut\cite{wang2022tokencut}. For each comparison, we showcase the output attention map and the output bounding box. Additionally, we display CAD-based bounding boxes, derived from both our refined model $f^h$ and the original model $f$, if available. For each method, we exhibit results on the PASCAL-VOC07~\cite{pascalvoc2007}, PASCAL-VOC12~\cite{pascalvoc2012}, and MS-COCO20K\cite{cocoeccv2014} datasets. The outcomes for the LOST model can be found in Figures~\ref{fig:appendix_uod_lost_coco},\ref{fig:appendix_uod_lost_voc07},\ref{fig:appendix_uod_lost_voc12}, while the TokenCut model results are illustrated in Figures~\ref{fig:appendix_uod_tokencut_coco}, \ref{fig:appendix_uod_tokencut_voc07}, \ref{fig:appendix_uod_tokencut_voc12}.

\end{document}